\documentclass[11pt]{article}

\usepackage[final]{acl}

\usepackage{times}
\usepackage{latexsym}

\usepackage[T1]{fontenc}

\usepackage[utf8]{inputenc}

\usepackage{microtype}

\usepackage{inconsolata}

\usepackage{graphicx}

\usepackage{hyperref}
\usepackage{array}
\usepackage{multirow}
\usepackage{color, colortbl}
\usepackage{subcaption}
\usepackage{pifont}
\usepackage{amsmath}
\usepackage{longtable}                                                                            
\usepackage{xltabular}  
\usepackage{tabularx}
\usepackage{booktabs}  
\usepackage{cuted}
\usepackage{amssymb}

\makeatletter
\newcommand{\dashrulew}[1]{%
  \noalign{\vskip1.1pt}%
  \noalign{\hbox to #1{\color{black!45}%
    \xleaders\hbox{\rule[0.5ex]{2.2pt}{0.4pt}\hskip1.8pt}\hfill}}%
  \noalign{\vskip1.1pt}}
\newcommand{\hdashline}{\dashrulew{\linewidth}}
\makeatother

\usepackage[table]{xcolor}   
\usepackage{fancyvrb}
\usepackage{xspace}
\usepackage[most]{tcolorbox}   
\usepackage{listings}

\usepackage{dsfont}
\usepackage{threeparttable}

\usepackage{pgfplots}
\usepgfplotslibrary{groupplots}
\usetikzlibrary{positioning, patterns, patterns.meta, calc}
\usepackage[most]{tcolorbox}
\usepackage{pifont}
\usepackage{xcolor}
\usepackage{textcomp}

\usepackage{placeins}                       

\newcommand{\rparagraph}[1]{\vspace{1.2mm}\noindent\textbf{#1.}}
\newcommand{\iparagraph}[1]{\vspace{1.2mm}\noindent\textit{#1.}}

\usepackage{todonotes}

\definecolor{Gray}{gray}{0.9}
\definecolor{ashgrey}{rgb}{0.7, 0.75, 0.71}
\definecolor{darkgreen}{rgb}{0,0.39,0}

\newcolumntype{g}{>{\columncolor{Gray}}c}
\newcolumntype{a}{>{\columncolor{ashgrey}}c}

\newcommand{\qa}{QA\xspace}
\newcommand{\kg}{KG\xspace}
\newcommand{\kgs}{KGs\xspace}
\newcommand{\llm}{LLM\xspace}
\newcommand{\llms}{LLMs\xspace}
\newcommand{\mt}{MT\xspace}

\newcommand{\multiglobe}{MultiGlobeQA\xspace}
\newcommand{\sfid}{SFID\xspace}
\newcommand{\sfids}{SFIDs\xspace}

\newcommand{\gemini}{Gemini-3-Flash\xspace}
\newcommand{\claude}{Claude-Opus-4.7\xspace}
\newcommand{\qwenmoe}{Qwen3.5-35B\xspace}
\newcommand{\qwen}{Qwen3.5-27B\xspace}
\newcommand{\gemma}{Gemma-3-27B-Instruct\xspace}
\newcommand{\deepseek}{DeepSeek-v4-Flash\xspace}
\newcommand{\qwennew}{Qwen3.8-27B\xspace}
\newcommand{\gemmanew}{Gemma-4-31B\xspace}
\newcommand{\glm}{GLM-4.7-Flash\xspace}
\newcommand{\gptossl}{gpt-oss-120B\xspace}
\newcommand{\gptosss}{gpt-oss-20B\xspace}
\newcommand{\fable}{Claude-Fable-5.1\xspace}

\newcommand{\osm}{OSM\xspace}
\newcommand{\kwg}{KnowWhereGraph\xspace}
\newcommand{\worldkg}{WorldKG\xspace}
\newcommand{\osmkg}{OSMH3KG\xspace}

\lstdefinestyle{promptstyle}{%
  basicstyle=\footnotesize\ttfamily\linespread{0.9}\selectfont,
  breaklines=true,
  breakatwhitespace=true,
  columns=fullflexible,
  keepspaces=true,
  showstringspaces=false,
  upquote=true,
  aboveskip=0pt,
  belowskip=0pt,          
  gobble=4,               
  extendedchars=true,
  literate=%
    {—}{{--}}1 {–}{{-}}1
    {→}{{$\rightarrow$}}1
    {“}{{``}}1 {”}{{''}}1 {‘}{{`}}1 {’}{{'}}1
    {à}{{\`a}}1 {á}{{\'a}}1 {â}{{\^a}}1 {ä}{{\"a}}1
    {è}{{\`e}}1 {é}{{\'e}}1 {ê}{{\^e}}1
    {í}{{\'i}}1 {ó}{{\'o}}1 {ö}{{\"o}}1 {ú}{{\'u}}1 {ü}{{\"u}}1
    {ç}{{\c c}}1 {ñ}{{\~n}}1 {ß}{{\ss}}1,
}

\DeclareTCBListing{promptbox}{ O{} m }{%
  listing only,
  enhanced jigsaw,
  breakable,
  colback=gray!7,
  colframe=black!60,
  colbacktitle=gray!25,
  coltitle=black,
  fonttitle=\bfseries\footnotesize,
  title={#2},
  boxrule=0.5pt,
  arc=0.5mm,
  left=2pt, right=2pt, top=2pt, bottom=2pt,
  listing style=promptstyle,
  #1,
}

\title{MultiGlobeQA: A Multilingual and Globally Diverse Benchmark for Geospatial Reasoning}

\author{Martin Böckling, Elizaveta Nosova, Heiko Paulheim \and Andreea Iana\\
       Data and Web Science Group, University of Mannheim, Germany \\ 
       \texttt{\{martin.boeckling, heiko.paulheim, andreea.iana\}@uni-mannheim.de}\\ 
       \texttt{elizaveta.nosova@students.uni-mannheim.de} \\}

\begin{document}
\maketitle
\begin{abstract}
Geospatial reasoning, i.e., computing distances, containment, and other spatial relations over real-world entities, is central to navigation and logistics, yet large language models (\llms) struggle with the required geometric and topological computation despite considerable geographic knowledge. Existing benchmarks localize these failures only partially: they are synthetic or small-scale, largely monolingual, and offer limited control over geographic coverage.
We introduce \multiglobe, a multilingual benchmark of 46,060 question-answer pairs spanning 14 spatial-function families and 15 answer formats, with execution-based ground truth over three knowledge graphs. It covers 201 countries and territories via income- and density-stratified sampling, with parallel questions in English and 16 additional high- and low-resource languages. 
Across parametric, reasoning, and agentic settings, \llms collapse on tasks requiring grid indexing and shape computation, while topological relations and directions fare best. Retrieval and tool use yield considerable gains, yet performance plateaus near two thirds even with gold facts, indicating that computation is the bottleneck. Even a frontier model leaves more than a quarter of a subset without a correct answer. With gold facts, all models also lag on low-income regions, a gap that gold facts widen.
\end{abstract}

\section{Introduction}
\label{sec:introduction}

Large language models (\llms) are increasingly deployed in navigation, logistics, and planning systems that require geospatial reasoning (inferring distances, containment, and other spatial relations over geographic entities) to answer questions about physical space, directly or through retrieval-augmented or agentic workflows \cite{mai2024opportunities,xie2024travelplanner,dihan2025mapeval,yu2026spatial,dorobantu2026geospatial}.
While \llms encode substantial geographic knowledge, such as coordinate representations \cite{gurnee2024language} and parametric facts \cite{roberts2023gpt4geo,bhandari2023large}, geospatial reasoning entails computation over geometric and topological relationships (distances, directions, coordinate transformations), a capability with which they struggle \cite{dihan2025mapeval,li2025stbench,truong2026gpsbench}.
Fig.~\ref{fig:teaser} illustrates this gap. A model underestimates the distance between two towns by an order of magnitude and still fails to compute a geohash even when the relevant facts are provided (top). Across question categories (bottom), explicit reasoning rarely improves on direct answering, whereas geographic facts paired with computation tools bring substantial but uneven gains: metric and topological questions reach 61\%, direction, shape, and uncertainty 60\%, and centrality, grid indexing and transformation 48\%. Geographic information alone is thus insufficient for reliable geospatial reasoning.

Yet existing benchmarks capture these failures only partially. 
Early spatial-reasoning benchmarks are largely synthetic, testing qualitative relations over abstract objects \citep{mirzaee2021spartqa,shi2022stepgame}, while those grounded in real geography are small-scale \citep{kefalidis2023benchmarking,li2025mapqa,saeedan2026gs} or monolingual, with limited control over coverage \citep{dihan2025mapeval,li2025stbench,bao2026urbangeoeval}.
Many also sample directly from OpenStreetMap (\osm) and Wikidata \cite{vrandevcic2014wikidata}, inheriting their uneven coverage \cite{herfort2023spatio} and socio-economic
biases \cite{das2025social}.
Thus, no benchmark localizes failures jointly across spatial operations, answer formats, regions, and languages.
\providecolor{okgreen}{HTML}{2E8B57}
\providecolor{badred}{HTML}{C92A2A}
\providecolor{cardline}{HTML}{9AA0A6}
\providecolor{cT1}{HTML}{9E9E9E}   
\providecolor{cT2}{HTML}{E8A33D}   
\providecolor{cT3a}{HTML}{1565C0}  
\providecolor{cT3b}{HTML}{7FB2E5}  
\providecommand{\cmk}{\textcolor{okgreen}{\ding{52}}}
\providecommand{\xmk}{\textcolor{badred}{\ding{56}}}

\def\MA{28.5}\def\MB{54.4}\def\MC{22.1}\def\MD{60.6}\def\ME{39.5}\def\MF{27.9} 
\def\DA{27.9}\def\DB{51.9}\def\DC{24.1}\def\DD{60.4}\def\DE{51.2}\def\DF{31.7} 
\def\GA{3.3} \def\GB{33.7}\def\GC{3.8} \def\GD{48.2}\def\GE{34.2}\def\GF{16.3} 

\begin{figure*}[t]
\centering
\begin{tikzpicture}[font=\small]
\def\CW{5.21} \def\GAP{0.18}
\foreach \i/\nm/\q/\sq/\ra/\rb in {%
 0/{Metric \& topological}/%
   {What is the distance between Irena and Bovalpun?}/%
   {ST\_DistanceSphere(a,b) = \textbf{19.05 km}}/%
   {\xmk~Gemini 3 Flash: 1.4 km}/{\cmk~+ gold triples: 18.99 km},
 1/{Direction, shape \& uncertainty}/%
   {In which direction must one go to get from Mutshatsha Airport to Lufupa?}/%
   {bearing(a,b) $\rightarrow$ compass bin = \textbf{East}}/%
   {\cmk~Gemma 3 27B: east}/{\xmk~Qwen3.8 27B: south},
 2/{Centrality, grid \& transformation}/%
   {What is the geohash of Punakha at 6-character precision?}/%
   {ST\_GeoHash(a,6) = \textbf{tuzuuv}}/%
   {\xmk~Gemini 3 Flash: tw3p6y}/{\xmk~+ gold triples: tst66v}}{%
  \pgfmathsetmacro{\X}{\i*(\CW+\GAP)}
  \node[anchor=north west,text width=\CW cm,align=center,inner sep=0pt,font=\small\bfseries] at (\X,0.52) {\nm};
  \node[anchor=north west,draw=cardline,rounded corners=2pt,line width=0.4pt,
        inner sep=5pt] at (\X,0.16) {%
    \begin{minipage}[t][2.2cm][t]{\dimexpr\CW cm-14pt\relax}\raggedright
    \q\\[2pt]
    {\scriptsize\textcolor{black!62}{gold computed by:}}\\[-1pt]
    {\scriptsize\texttt{\textcolor{black!78}{\sq}}}\\[2pt]
    \ra\\[2.5pt]
    \rb
    \end{minipage}};
}
\foreach \i/\a/\b/\c/\d/\e/\f in {0/\MA/\MB/\MC/\MD/\ME/\MF, 1/\DA/\DB/\DC/\DD/\DE/\DF,
                                  2/\GA/\GB/\GC/\GD/\GE/\GF}{%
  \pgfmathsetmacro{\X}{\i*(\CW+\GAP)}
  \node[anchor=north west,inner sep=0pt] at (\X,-2.6) {%
  \begin{tikzpicture}
  \begin{axis}[width=\dimexpr\CW cm-4pt+0.78cm\relax,height=3.95cm,
    ybar,bar width=12pt,bar shift=0pt,
    ymin=0,ymax=70,ytick={0,10,20,30,40,50,60,70},
    xtick={1,2,3,4,5.6,6.8},xticklabels={T1,T2,T3a,T3b,T1-o,T3-o},
    xmin=0.40,xmax=7.45,
    tick label style={font=\scriptsize},ylabel={\footnotesize EM (\%)},
    ylabel style={yshift=-17pt,xshift=-6.9mm},
    nodes near coords,
    nodes near coords style={font=\scriptsize,/pgf/number format/precision=1,/pgf/number format/fixed,/pgf/number format/fixed zerofill,
                             inner sep=1pt},
    axis y line*=left,axis x line*=bottom,ymajorgrids,
    grid style={black!12,line width=0.3pt},clip=false]
  \addplot[fill=cT1,draw=black!45]  coordinates {(1,\a)};
  \addplot[fill=cT2,draw=black!45]  coordinates {(2,\c)};
  \addplot[fill=cT3a,draw=black!45] coordinates {(3,\e)};
  \addplot[fill=cT3b,draw=black!45] coordinates {(4,\f)};
  \draw[black!40,line width=0.4pt] (axis cs:4.80,0) -- (axis cs:4.80,70);
  \addplot[fill=cT1,draw=black!45,postaction={pattern={Lines[angle=45,distance=3.2pt,line width=0.8pt]},pattern color=white}]
        coordinates {(5.6,\b)};
  \addplot[fill=cT3a,draw=black!45,postaction={pattern={Lines[angle=45,distance=3.2pt,line width=0.8pt]},pattern color=white}]
        coordinates {(6.8,\d)};
  \end{axis}
  \end{tikzpicture}};
}
\node[anchor=north,text width=15.9cm,align=center,inner sep=0pt,font=\footnotesize] at (8.0,-5.72) {%
  \tikz\node[fill=cT1,draw=black!45,minimum size=6.5pt,inner sep=0pt]{};~T1 parametric\enspace
  \tikz\node[fill=cT2,draw=black!45,minimum size=6.5pt,inner sep=0pt]{};~T2 reasoning\enspace
  \tikz\node[fill=cT3a,draw=black!45,minimum size=6.5pt,inner sep=0pt]{};~T3a agentic (\kg)\enspace
  \tikz\node[fill=cT3b,draw=black!45,minimum size=6.5pt,inner sep=0pt]{};~T3b agentic (web)\enspace
  \tikz\node[fill=cT1,draw=black!45,minimum size=6.5pt,inner sep=0pt,
             postaction={pattern={Lines[angle=45,distance=2.4pt,line width=0.6pt]},pattern color=white}]{};~%
  -o: oracle, gold triples injected};
\end{tikzpicture}
\caption{\textbf{\llms systematically fail on questions requiring geospatial reasoning and computation.} \emph{Top}: three example items (distance, direction, geohash), each with its executable gold answer and the models' predictions (\cmk~correct, \xmk~wrong). \textit{+\,gold triples} marks the oracle setting, with the gold triples injected into the prompt. \emph{Bottom}: exact match (\%) per category, averaged over the nine \llms, at the parametric (T1), reasoning (T2), and agentic tiers, which execute code and retrieve from the \kg (T3a) or the web (T3b). Hatched \emph{-o} bars are the oracle conditions.
}
\label{fig:teaser}
\end{figure*}

\rparagraph{Contributions}
We address these gaps with \multiglobe, an open, large-scale multilingual benchmark for geospatial reasoning.
\multiglobe comprises \textbf{46,060 \qa pairs} from 65 templates, including false-premise questions and a multimodal slice, with \emph{execution-based} ground truth over three geographic knowledge graphs (\kgs), so answers are verified by construction. 
Compared to existing benchmarks, it is:
\textbf{(1)} \emph{broader} -- it spans 14 spatial-function families and 15 answer formats, from Booleans and sets to grid-cell encodings and geometries; 
\textbf{(2)} \emph{geographically diverse} -- entities are sampled under income and density stratification across 201 countries and territories in four income tiers; 
\textbf{(3)} \emph{multi-parallel} --  all questions are in 17 high- and low-resource languages with human-verified translations. 
These support fine-grained analysis of failure types, geographic bias, and cross-lingual comparison.

We evaluate nine \llms across parametric, reasoning, and agentic tiers, with oracle conditions injecting gold \kg triples to approximate perfect retrieval.
With gold triples and tools, the best model reaches 65.3\%, and on average grid indexing and shape stay below 30\% while coordinate questions reach 86\%.
Retrieval and tool use yield sizable gains, whereas explicit reasoning provides little benefit. For eight of the nine models, live retrieval recovers only half to three quarters of perfect retrieval.
Finally, low-income regions lag for most models under parametric knowledge and for every model under perfect retrieval, while performance remains largely stable across languages.

\section{Related Work}
\label{sec:related_work}

\rparagraph{Spatial and Geospatial Reasoning Benchmarks}
Textual spatial-reasoning benchmarks, such as bAbI tasks 17/19 \cite{weston2015towards}, SPARTQA \cite{mirzaee2021spartqa}, SpaRTUN \cite{mirzaee2022transfer}, and StepGame \cite{shi2022stepgame}, are largely synthetic, evaluating reasoning over a fixed set of spatial relations in toy worlds and leaving open how models reason about real-world geography. 
To ground reasoning in real geographic entities, one line studies question answering (\qa) over geographic \kgs \cite{mai2021geographic}. Extending template-based \qa over linked geospatial data \cite{punjani2018template}, GeoQuestions1089 \cite{kefalidis2023benchmarking,kefalidis2024question} pairs 1,089 manually authored questions with executable SPARQL and GeoSPARQL queries over YAGO2 \cite{hoffart2013yago2} and YAGO2geo \cite{karalis2019extending}, targeting primarily factual and relational retrieval.

A second line directly evaluates foundation models: \llms encode geographic coordinates \cite{gurnee2024language} and recall facts without retrieval \cite{roberts2023gpt4geo,bhandari2023large}, yet struggle to compute over them \cite{truong2026gpsbench,li2025stbench}, particularly for distance, direction, and counting questions \cite{dihan2025mapeval}.
Retrieval-, tool-, and agent-augmented systems offload computations to external resources or GIS pipelines, improving accuracy \cite{dihan2025mapeval,yu2026spatial,krechetova2025geobenchx,hasan2026mapagent,zhang2025geoanalystbench}, without isolating the model's reasoning from the tools' computation \cite{bao2026urbangeoeval}. 
\citet{suizu2026automatic}, closest to our setting yet single-country, generate geospatial questions by composing spatial and entity constraints, likewise finding that models ground entities well but fail at precise spatial reasoning.

\rparagraph{Execution-Based Benchmark Construction}
Benchmarks increasingly obtain verifiable ground truth by executing structured queries over a knowledge source rather than matching surface forms, the standard in text-to-SQL, where correctness is measured by execution accuracy \cite{yu2018spider,li2023can}.
The same recipe underlies non-spatial \kg and graph \qa benchmarks such as CRAG \cite{yang2024crag}, STaRK \cite{wu2024stark}, and GRBench \cite{jin2024graph}, which target retrieval and general graph reasoning rather than spatial computation, as well as geospatial \qa like MapQA \cite{li2025mapqa} and GS-QA \cite{saeedan2026gs}, which compute answers via spatial SQL over \osm.

\rparagraph{Limitations of Current Benchmarks}
Existing benchmarks are limited along several dimensions (Table \ref{tab:appendix_related_benchmarks}). Most are small or manually authored, cover a narrow range of spatial operations, or are city- or region-specific, and they are nearly exclusively monolingual. They inherit the high-income, urban skew of their \osm- and Wikidata-derived data \cite{herfort2023spatio,das2025social}, a bias that propagates into model behavior across regions and languages \cite{manvi2024large,moayeri2024worldbench,faisal2023geographic}.
\multiglobe addresses these gaps jointly: it derives large-scale, execution-verified, multilingual questions over a wide range of spatial operations and three independent geographic \kgs, sampled under income and density stratification for diverse geographic coverage, extending multilingual geographic \qa \cite{roh2025xlqa,hwang2025learn} from cross-lingual knowledge to spatial reasoning.
\section{\multiglobe}
\label{sec:benchmark}

We construct \multiglobe in four stages: we define a taxonomy of typed templates (\S\ref{sec:template_taxonomy}), sample entities from three \kgs (\S\ref{sec:kgs_entity_sampling}), instantiate each template into execution-verified English \qa pairs (\S\ref{sec:ground_truth}), and translate the templates into 16 languages, re-instantiating them with the same entities, using target-language labels if available (\S\ref{sec:multilingual_extension}).
\begin{table}[t]
    \centering 
    \small
    \resizebox{\columnwidth}{!}{%
         \begin{tabular}{@{} l l @{}}
        \toprule
        \textbf{Spatial Operation (\#templates)} & \textbf{Spatial Functions (\sfids)} \\
        \midrule
        Metric \& topological (23)            & (A) distance, (B) containment, (C) topology  \\
        Network \& path (4)            & (D) network/path  \\
        Hierarchy \& comparison (9)            & (E) administrative hierarchy, (F) comparison  \\
        Centrality, grid \& transformation (10)            & (G) centrality, (H) grid, (M) transformation, (N) coord.  \\
        Direction, shape \& uncertainty (14)            & (J) direction, (K) shape, (L) uncertainty  \\
        Spatio-temporal events (5)            & (I) spatio-temporal events  \\
        \bottomrule
        \end{tabular}%
        }

    \caption{\textbf{\multiglobe's template taxonomy}.}
    \label{tab:taxonomy_overview}
\end{table}

\subsection{Template Taxonomy}
\label{sec:template_taxonomy}

We construct \multiglobe around a taxonomy of 65 top-level templates, organized into 14 spatial-function families (spatial-function IDs, \sfids). Table \ref{tab:taxonomy_overview} summarizes the spatial-function coverage across six clusters of common geospatial reasoning operations \cite{egenhofer1991point,randell1992spatial,cohn2001qualitative,worboys2004gis}. 
We author each template, which expands into one or more parameterized sub-templates (129 in total), covering variations such as cardinal directions or aggregation modes.
To increase linguistic diversity, we generate 1-5 \llm-drafted, human-verified paraphrases per sub-template (315 English variants in total).
Templates use typed placeholders for entities, values, and parameters, and support 15 answer formats (e.g., Boolean, numeric, geometric, categorical, and temporal).
Table \ref{tab:appendix_templates} lists all templates with examples.

\subsection{Stratified Entity Sampling}
\label{sec:kgs_entity_sampling}

\rparagraph{Data Sources}
We sample entities from three complementary \kgs (Table \ref{tab:appendix_kgs}): (i) \textit{\worldkg} \cite{dsouza2021worldkg}, built from tagged \osm nodes, contributing named places and amenities as points with the broadest country coverage; (ii) \textit{\kwg} \cite{janowicz2022know}, a cross-domain \kg integrating hazard, health, and administrative hierarchies from heterogeneous, largely US-based sources; and (iii) \textit{\osmkg} \cite{bockling2024planet}, an H3-indexed \kg spanning the full range of \osm geometries, from POIs to road networks and administrative polygons. 
We convert each \kg into a common entity table of WKT geometries, canonical names, semantic categories mapped from \texttt{rdf:type}, and H3 indices \citep{brodsky2018h3}. We resolve ambiguous names (e.g., chain store names), by appending contextual properties such as neighborhood or street name, and coordinates if needed.

\rparagraph{Strata Grid}
To mitigate the over-representation of high-income and urban regions in \osm- and Wikidata-derived \kgs \cite{herfort2023spatio,das2025social}, we assign each entity two stratification tags (Appendix \ref{sec:appendix_benchmark_construction}): (i) an income tier (low-, lower-middle-, upper-middle-, or high-income countries; \textit{LIC}/\textit{LMC}/\textit{UMC}/\textit{HIC}) obtained from the World Bank FY2026 classification, and (ii) a density tier (\textit{low}/\textit{medium}/\textit{high}) obtained from the WorldPop R2025A 1km population raster \cite{tatem2017worldpop}, binning values at the global 33rd and 67th percentiles.
Combining four income and three density tiers yields a $4\times3$ grid of 12 strata. We sample entities and generate questions independently within each cell, drawing up to 30 entity tuples per (template, \kg, stratum) pair so that every populated cell contributes whenever the template is satisfiable.
%
\multiglobe spans 201 countries and territories across all income and density tiers (Table \ref{tab:strata_counts}, Fig.~\ref{fig:appendix_entity_map}).

\begin{table}[t]
  \centering
  \footnotesize
  \setlength{\tabcolsep}{6pt}
  \resizebox{\columnwidth}{!}{%
  \begin{tabular}{@{}l rrr rr@{}}
      \toprule
      \textbf{Income} & \multicolumn{3}{c}{\textbf{Density tier}} & & \\
      \cmidrule(lr){2-4}
        
       \textbf{tier} & \emph{low} & \emph{medium} & \emph{high} & \textbf{\#Questions} & \textbf{\#Countries}\\
      \midrule

      \textsc{HIC}    & 5{,}521 & 6{,}009 & 6{,}713 & 18{,}243 (40\%)& 76\\
      \textsc{UMC}    & 1{,}156 & 4{,}012 & 4{,}194 & 9{,}362 (20\%)& 56\\
      \textsc{LMC}    & 2{,}446 & 4{,}095 & 3{,}699 & 10{,}240 (22\%)& 44\\
      \textsc{LIC}    & 502 & 3{,}858 & 3{,}855 & 8{,}215 (18\%)& 25\\
      \midrule
      
      \shortstack[l]{\textbf{Total}\\\phantom{(00\%)}} & \shortstack[r]{\textbf{9{,}625}\\(21\%)} & \shortstack[r]{\textbf{17{,}974}\\(39\%)} & \shortstack[r]{\textbf{18{,}461}\\(40\%)} & \shortstack[r]{\textbf{46{,}060}\\\phantom{(00\%)}} & \shortstack[r]{\textbf{201}\\\phantom{(00\%)}}\\

      \bottomrule
  \end{tabular}%
  }
  \caption{\textbf{Question counts per strata cell and country in \multiglobe}. Rows: World Bank income tiers; columns: H3-resolution-3 density terciles.}
  \label{tab:strata_counts}
\end{table}

\subsection{Ground-truth Computation}
\label{sec:ground_truth}
We compute execution-based ground truth by pairing each question template with a Spark SQL query template executed over the \kg, instead of generating questions and verifying answers post-hoc.\footnote{We use Spark SQL with Apache Sedona over the entity tables of \S\ref{sec:kgs_entity_sampling}, as the three \kgs share no SPARQL endpoint and GeoSPARQL does not cover all 14 families.}
For each (template, \kg, strata cell) combination, the executor filters candidate entities by role-specific type (e.g., \texttt{cafe}), evaluates the query, and samples from the results. 
We validate each result against type-specific constraints (e.g., distances bounded by Earth's circumference, polygons closed and non-self-intersecting) and discard degenerate executions (i.e., divisions by zero, disconnected graphs). 
We de-duplicate on $(\textrm{question}, \textrm{ground truth})$ and instantiate each tuple with a uniformly sampled natural-language variant. The English benchmark contains \textbf{46,060 verified \qa pairs} spanning all \textbf{65 templates} and \textbf{315 variants}, with approximately 6.9k, 24k, and 15k pairs grounded in \kwg, \osmkg, and \worldkg, respectively (Table \ref{tab:appendix_per_kg_sfid_counts}).
Appendix \ref{sec:appendix_data_sources} reports per-\sfid and answer-format distributions.

\subsection{False-Premise and Multimodal Questions}
\label{sec:multimodal_false_premise}
We include \textbf{3,589 false-premise questions} (7.8\% of the English benchmark), for which the ground-truth query finds that the asserted relation does not hold for the sampled entities, e.g., when a region contains no entity of the requested type. We set no target ratio, so their share reflects how often each relation fails over real geometries. These questions probe whether models detect the false premise, by declining to answer or returning the empty result, instead of hallucinating a plausible answer.

For a subset of templates constrained to visually salient entities (e.g., landmarks, monuments), we release a separate multimodal slice over \osmkg and \worldkg (\textbf{946 questions}, 2\% of the English benchmark), replacing the entity name with a representative image from Wikidata or Wikimedia Commons via the \kg's image triples.\footnote{We exclude \kwg because its hazard- and event-centric entities lack canonical image bindings.} 

\subsection{Multilingual Extension}
\label{sec:multilingual_extension}

We translate the English benchmark into 16 languages, chosen to span six language families, three resource tiers, and seven scripts (Table \ref{tab:languages}). 
\begin{table}[t]
    \centering
    \footnotesize
  \setlength{\tabcolsep}{2.7pt}   
  \begin{tabular}{@{}lllll@{}}                          
  \toprule                                                                                                                                                                                  
  \textbf{Code} & \textbf{Language} & \textbf{Family / Subgroup} & \textbf{Script} & \textbf{Tier}\\                                                                                                                                     
    \midrule
    fra & French     & IE / Romance         & Lat & H \\
    ita & Italian    & IE / Romance         & Lat & H \\
    por & Portuguese & IE / Romance         & Lat & H \\
    ron & Romanian   & IE / Romance         & Lat & M \\
    spa & Spanish    & IE / Romance         & Lat & H \\
    
    bul & Bulgarian  & IE / Balto-Slavic    & Cyr & M \\
    rus & Russian    & IE / Balto-Slavic    & Cyr & H \\
    
    deu & German     & IE / Germanic        & Lat & H \\
    eng & English    & IE / Germanic        & Lat & H \\
    
    urd & Urdu       & IE / Indo-Aryan      & PA  & L \\
    
    sqi & Albanian   & IE / Albanian        & Lat & L \\
    
    ell & Greek      & IE / Graeco-Phrygian & Grk & M \\

    jpn & Japanese   & Japonic        / Japanesic    & JK  & M \\
    kat & Georgian   & Kartvelian     / Georgian-Zan & Geo & M \\
    tur & Turkish    & Turkic         / Oghuz        & Lat & M \\
    vie & Vietnamese & Austro-Asiatic / Vietic       & Lat & M \\
    zho & Chinese    & Sino-Tibetan   / Sinitic      & Han & M \\   
    \bottomrule                                                                                                                                                                               
  \end{tabular}
  
    \caption{\textbf{The 17 languages in \multiglobe.} Codes are ISO 639-3;
    \textit{Family}/\textit{Subgrouping} follow Glottolog \cite{hammarstrom2026glottolog} (IE: Indo-European);
    \textit{Tier} (H/M/L = high/mid/low resource) is adapted 
    from FLORES \cite{goyal-etal-2022-flores,costa2022no}. Scripts: Lat 
    (Latin), Cyr (Cyrillic), Grk (Greek), PA (Perso-Arabic), 
    Han, JK (Japanese Kana/Kanji), Geo (Georgian).}
    \label{tab:languages}
    \vspace{-0.5em}
  \end{table}

\rparagraph{Template Translation}
We translate the 315 template variants into each target language with Google Cloud Translation Advanced (v3), an adaptive \llm-based machine translation (\mt) system.\footnote{To preserve template slots, we mark placeholders as non-translatable. A post-processing pass verifies that every source placeholder appears exactly once in the output.}
Two proficient speakers of both English and the target language then post-edit every translated template to ensure semantic precision (e.g., the difference between ``within'' and ``inside'') and fluency, assigning one of three ordinal labels (\textit{Correct} $<$ \textit{Correct – Phrasing Improvement Needed} $<$ \textit{Incorrect}) and supply a corrected string when the label is not \textit{Correct} (Appendix~\ref{sec:appendix_human_annotation_protocol}).

\rparagraph{Disagreement Resolution}
The two annotators per language reach an agreement of 69.7\%. For the remaining cases, we consolidate the annotators' judgments into a single translation per language in two stages: a rule-based pass which handles deterministic cases, and an \llm-based judge (Claude Opus 4.7) for the non-deterministic ones (see Appendix~\ref{sec:disagreement_resolution_pipeline}). 
Rule-based resolution settles 88.6\% of items across 16 target languages, and the judge arbitrates the rest (Table \ref{tab:appendix_annotation_stats}, Fig. \ref{fig:resolution_pipeline_results}).
On a sample of four languages, a third annotator agreed with the judge on at least 88.9\% of arbitrated questions.

\subsection{Post-editing Instantiated Questions}
\label{sec:postediting_questions}
Instantiating templates with concrete entities introduces language-specific grammatical artifacts (e.g., wrong articles, agreement mismatches) that are tedious to fix manually at scale. 
We correct them with an ensemble of three \llm post-editors (\gemini, \deepseek, and \qwen), each restricted to the grammar surrounding substituted entities and returning typed edits from a 13-category taxonomy. We apply edits proposed by at least two models, falling back to \gemini's output otherwise. Across all 17 languages, the ensemble edits 16.2\% of questions, with at least two models agreeing on 90.1\% (see Appendix \ref{sec:appendix_postediting_ensemble}).

\subsection{Released Splits}
\label{sec:released_variants}
We provide \multiglobe in two splits, each in all 17 languages.
The \textit{large} split is the full benchmark of 46,060 \qa pairs per language.
The \textit{small} split is a canonical subset of \textit{large}, comprising 5,916 questions (5,636 text, 280 image) sampled at up to 200 per (\kg, \sfid) cell, preserving per-cell coverage while reducing evaluation cost roughly eightfold.
We release the benchmark\footnote{\href{https://huggingface.co/datasets/aiana94/MultiGlobeQA}{https://huggingface.co/datasets/aiana94/MultiGlobeQA}}, code\footnote{\href{https://github.com/andreeaiana/MultiGlobeQA}{https://github.com/andreeaiana/MultiGlobeQA}}, and \kg snapshots \cite{bockling_2026_21786184}.
\section{Experimental Setup}
\label{sec:experimental_setup}

\rparagraph{Models}
We evaluate eight open-weight models from four families -- \qwen and \qwenmoe \cite{qwen3.5}, \qwennew \cite{qwen38}, \gemma \cite{gemmateam2025gemma3technicalreport}, \gemmanew \cite{team2026gemma}, \glm \cite{zeng2025glm}, \gptosss and \gptossl \cite{openai2025gptoss120bgptoss20bmodel} -- and one closed-source, \gemini \cite{googledeepmind2025gemini3flash}. In the reasoning tier, \qwen, \qwenmoe and \gemma use prompted chain-of-thought, while the others use native reasoning modes.
We also report a \textbf{majority-class baseline} that predicts each sub-template's most frequent gold answer, scored by exact equality.
Appendices \ref{sec:appendix_model_details} and \ref{sec:appendix_inference_setup} report model configurations and inference setup.

\rparagraph{Evaluation Settings}
We evaluate models under three tiers that isolate parametric knowledge, reasoning, and tool use.
\textbf{Tier 1 (T1, parametric)} directly presents the question. 
\textbf{Tier 2 (T2, reasoning)} adds explicit reasoning over the same prompt.
\textbf{Tier 3 (T3, agentic)} lets the model write and execute Python in a restricted \textsc{CodeAgent} interpreter \cite{smolagents}, iteratively calling tools to gather evidence. Its variants differ in the retrieval source: spatial \kgs (\textbf{T3a}), web search (\textbf{T3b}), and both (\textbf{T3c}). The \kg tools support entity lookup but no query execution, so the model must compose the spatial computation itself.
We additionally evaluate four \emph{oracle} conditions that inject the gold \kg triples supporting each answer, approximating perfect retrieval. Three differ only in surface form -- structured JSON, raw N-Triples, or verbalized prose (\textbf{T1-o}, structured/raw/verbalized) -- to test representation sensitivity. The fourth (\textbf{T3-o}) supplies the same structured triples to the T3 agent, retaining the Python interpreter but disabling retrieval tools. 
These comparisons separate missing knowledge from missing computation (T1-o vs. T1) and perfect from live retrieval (T3-o vs. T3).
Appendix \ref{sec:appendix_model_prompts} lists all prompts.
\begin{table*}[t]
\centering
\small
\setlength{\tabcolsep}{1pt}
\newcommand{\sdv}[1]{\textsubscript{\tiny #1}}

    \resizebox{\textwidth}{!}{%
    \begin{tabular}{l@{\hspace{5pt}}cgcgcgcgcgcgcgcgcg}
    \toprule
     & \multicolumn{2}{c}{\textbf{Qwen3.5}} & \multicolumn{2}{c}{\textbf{Qwen3.5}} & \multicolumn{2}{c}{\textbf{Qwen3.8}} & \multicolumn{2}{c}{\textbf{Gemma-3}} & \multicolumn{2}{c}{\textbf{Gemma-4}} & \multicolumn{2}{c}{\textbf{gpt-oss}} & \multicolumn{2}{c}{\textbf{gpt-oss}} & \multicolumn{2}{c}{\textbf{GLM-4.7}} & \multicolumn{2}{c}{\textbf{Gemini-3}} \\
     & \multicolumn{2}{c}{\textbf{27B}} & \multicolumn{2}{c}{\textbf{35B}} & \multicolumn{2}{c}{\textbf{27B}} & \multicolumn{2}{c}{\textbf{27B}} & \multicolumn{2}{c}{\textbf{31B}} & \multicolumn{2}{c}{\textbf{120B}} & \multicolumn{2}{c}{\textbf{20B}} & \multicolumn{2}{c}{\textbf{Flash}} & \multicolumn{2}{c}{\textbf{Flash}} \\
    \cmidrule(lr){2-3} \cmidrule(lr){4-5} \cmidrule(lr){6-7} \cmidrule(lr){8-9} \cmidrule(lr){10-11} \cmidrule(lr){12-13} \cmidrule(lr){14-15} \cmidrule(lr){16-17} \cmidrule(lr){18-19}
    \textbf{Tier} & EM & EM\textsubscript{cov} & EM & EM\textsubscript{cov} & EM & EM\textsubscript{cov} & EM & EM\textsubscript{cov} & EM & EM\textsubscript{cov} & EM & EM\textsubscript{cov} & EM & EM\textsubscript{cov} & EM & EM\textsubscript{cov} & EM & EM\textsubscript{cov} \\
    \midrule
    \multicolumn{19}{@{}l}{\emph{no external context}} \\
    
    T1 & 22.7\sdv{0.1} & 42.8\sdv{0.2} & 17.3\sdv{0.0} & 57.4\sdv{0.2} & 18.5\sdv{0.1} & 44.3\sdv{0.4} & 16.7\sdv{0.1} & 17.3\sdv{0.1} & 14.7\sdv{0.0} & 28.3\sdv{0.0} & 12.4\sdv{0.2} & 51.6\sdv{0.6} & 16.5\sdv{0.1} & 44.8\sdv{0.2} & 15.1\sdv{0.1} & 29.9\sdv{0.1} & 22.7\sdv{0.1}\phantom{$^{\dagger}$} & 24.3\sdv{0.1}\phantom{$^{\dagger}$} \\
    T2 & 21.7\sdv{0.1} & 37.0\sdv{0.2} & 10.3\sdv{0.1} & 67.3\sdv{0.2} & 13.7\sdv{0.3} & 25.2\sdv{0.4} & 16.1\sdv{0.0} & 19.2\sdv{0.0} & 12.6\sdv{0.2} & 36.8\sdv{0.5} & 17.6\sdv{0.0} & 29.8\sdv{0.1} & \phantom{0}4.5\sdv{0.2} & 64.1\sdv{1.7} & 13.2\sdv{0.4} & 32.8\sdv{1.2} & 24.6\sdv{0.1}\phantom{$^{\dagger}$} & 26.9\sdv{0.1}\phantom{$^{\dagger}$} \\
    
    \midrule
    \multicolumn{19}{@{}l}{\emph{agentic retrieval from \kg (T3a), web (T3b), or both (T3c)}} \\
    T3a & \textbf{44.3}\sdv{0.4} & 72.5\sdv{0.4} & \textbf{37.0}\sdv{0.1} & 75.0\sdv{0.0} & \textbf{49.0}\sdv{0.0} & 72.0\sdv{0.1} & \textbf{30.7}\sdv{0.0} & 51.8\sdv{0.0} & \textbf{47.0}\sdv{0.0} & 57.9\sdv{0.0} & \textbf{40.7}\sdv{0.2} & 52.6\sdv{0.2} & \textbf{29.5}\sdv{0.5} & 51.8\sdv{0.6} & \textbf{31.5}\sdv{0.4} & 61.9\sdv{0.1} & \textbf{29.5}\sdv{3.5}\phantom{$^{\dagger}$} & 40.1\sdv{2.2}\phantom{$^{\dagger}$} \\
    T3b & 22.6\sdv{0.7} & 64.6\sdv{0.9} & 22.8\sdv{0.4} & 62.3\sdv{0.2} & 23.2\sdv{0.9} & 66.2\sdv{0.6} & 21.7\sdv{0.4} & 37.8\sdv{0.3} & \textbf{31.8}\sdv{0.7} & 48.5\sdv{0.4} & 26.9\sdv{0.4} & 45.4\sdv{0.5} & 14.2\sdv{0.1} & 41.7\sdv{0.4} & 16.5\sdv{0.5} & 56.1\sdv{1.3} & 27.2$^{\dagger}$\phantom{\sdv{0.0}} & 36.7$^{\dagger}$\phantom{\sdv{0.0}} \\
    T3c & \textbf{44.1}\sdv{0.1} & 73.7\sdv{0.3} & \textbf{37.2}\sdv{0.2} & 76.4\sdv{0.3} & \textbf{48.5}\sdv{0.2} & 75.1\sdv{0.2} & \textbf{31.5}\sdv{0.5} & 49.2\sdv{0.3} & \textbf{49.8}\sdv{0.1} & 57.2\sdv{0.2} & \textbf{40.8}\sdv{0.3} & 52.6\sdv{0.5} & \textbf{30.1}\sdv{0.1} & 51.0\sdv{0.2} & \textbf{31.5}\sdv{0.2} & 65.1\sdv{1.0} & 28.0$^{\dagger}$\sdv{0.0} & 37.1$^{\dagger}$\sdv{0.0} \\
    
    \midrule
    \multicolumn{19}{@{}l}{\emph{oracle conditions (gold triples injected)}} \\
    T1-o & \textbf{50.3}\sdv{0.1} & 65.2\sdv{0.1} & \textbf{29.6}\sdv{0.1} & 69.5\sdv{0.4} & \textbf{49.1}\sdv{0.1} & 63.8\sdv{0.1} & \textbf{38.2}\sdv{0.0} & 44.5\sdv{0.0} & \textbf{50.3}\sdv{0.0} & 66.1\sdv{0.0} & \textbf{51.0}\sdv{0.2} & 79.8\sdv{0.3} & \textbf{48.9}\sdv{0.3} & 76.5\sdv{0.4} & \textbf{34.1}\sdv{0.2} & 49.1\sdv{0.2} & \textbf{57.0}$^{\dagger}$\sdv{0.0} & 63.9$^{\dagger}$\sdv{0.0} \\
    T3-o & \textbf{61.6}\sdv{0.2} & 77.1\sdv{0.4} & \textbf{55.0}\sdv{0.2} & 79.1\sdv{0.2} & \textbf{65.3}\sdv{0.1} & 77.0\sdv{0.1} & \textbf{45.3}\sdv{0.2} & 73.2\sdv{0.4} & \textbf{63.3}\sdv{0.2} & 78.4\sdv{0.1} & \textbf{61.4}\sdv{0.1} & 76.3\sdv{0.3} & \textbf{57.4}\sdv{0.2} & 78.4\sdv{0.1} & 13.5\sdv{0.3} & 55.3\sdv{1.0} & \textbf{60.2}$^{\dagger}$\sdv{3.1} & 81.7$^{\dagger}$\sdv{2.0} \\
    \midrule
    
    \emph{Majority baseline} & 28.8\phantom{\sdv{0.0}} & 28.8\phantom{\sdv{0.0}} & 28.8\phantom{\sdv{0.0}} & 28.8\phantom{\sdv{0.0}} & 28.8\phantom{\sdv{0.0}} & 28.8\phantom{\sdv{0.0}} & 28.8\phantom{\sdv{0.0}} & 28.8\phantom{\sdv{0.0}} & 28.8\phantom{\sdv{0.0}} & 28.8\phantom{\sdv{0.0}} & 28.8\phantom{\sdv{0.0}} & 28.8\phantom{\sdv{0.0}} & 28.8\phantom{\sdv{0.0}} & 28.8\phantom{\sdv{0.0}} & 28.8\phantom{\sdv{0.0}} & 28.8\phantom{\sdv{0.0}} & 28.8\phantom{$^{\dagger}$}\phantom{\sdv{0.0}} & 28.8\phantom{\sdv{0.0}}\\
    \bottomrule
    \end{tabular}}

    \caption{\textbf{Results across evaluation tiers} (\emph{small} split, English, text modality; 4{,}979 \emph{true-premise} questions; $\mathrm{EM}\uparrow$, $\mathrm{EM}_{\mathrm{cov}}\uparrow$). $\mathrm{EM}$ counts refusals and non-answers as incorrect; $\mathrm{EM}_{\mathrm{cov}}$ is computed over answered questions only. Oracle tiers inject gold triples; T1-o uses \emph{verbalized} triples (Appendix \ref{sec:oracle_context_ablation}). $\mathrm{EM}$ above baseline is in bold. Subscripts give standard deviation over three seeds ($^{\dagger}$ marks fewer seeds; single-run cells show no standard deviation).}
    \label{tab:main_results}
    \vspace{-0.5em}
\end{table*}

\rparagraph{Evaluation Metrics}
We report two core metrics: \emph{exact match} ($\mathrm{EM}$) and \emph{normalized error} ($\mathrm{NE}$), both computed by deterministic, answer-type-dependent functions.
For $\mathrm{EM}$, we define a correctness function $E_{t_i}(\hat a_i, a_i^{*})\in\{0,1\}$ for each answer type $t_i$, which returns 1 iff the prediction $\hat{a}_i$ matches the gold $a_i^*$ under type-specific criteria: exact equality for categorical and grid-cell answers, geodesic or relative tolerance $\tau_i$ for continuous answers (distances, areas, ratios, coordinates, geometries, dates), exact set equality for enumerations, and $\mathrm{IoU}\geq 0.5$ for polygons. $\mathrm{NE}$ instead measures error magnitude through $e_t(\hat a_i, a_i^{*})\in[0,1]$: $\min(1, d_i/\tau_i)$ for continuous types, $1-\mathrm{Jaccard}$ for sets, and $1-\mathrm{IoU}$ for polygons, with $e_{t_i}=1$ for refusals, non-answers (budget exhaustion and harness errors), and format errors.
$\mathrm{EM}$ and $\mathrm{NE}$ are the means of $E_{t_i}$ and $e_{t_i}$ over the $N$ questions; $\mathrm{NE}=0$ is a perfect prediction and $\mathrm{NE}=1$ an error beyond tolerance (Table \ref{tab:appendix_eval_metrics}).
%
Unless stated otherwise, we compute $\mathrm{EM}$ and $\mathrm{NE}$ over \emph{true-premise} questions, counting refusals and non-answers as incorrect.

We additionally report coverage ($\mathrm{Cov}$), the share of true-premise questions receiving a parseable answer, and $\mathrm{EM}_\mathrm{cov}$ over those, so that $\mathrm{EM} = \mathrm{Cov}\times\mathrm{EM}_{\mathrm{cov}}$ separates willingness to answer from accuracy when answering. Coverage is reduced by abstention, format errors, harness errors ($\mathrm{Err}$; e.g., context-window overflow), and budget exhaustion ($\mathrm{Exh}$), i.e., agent runs that reach the step cap or end without a final answer and single-shot completions cut off at the output-token limit. 
The false-refusal rate ($\mathrm{FRR}$) is the share of true-premise questions with an explicit abstention, and the negative rejection rate ($\mathrm{NRR}$) the share of correctly rejected false-premise questions, separating premise detection from indiscriminate abstention. $\mathrm{Exh}$ and $\mathrm{Err}$ are reported separately and are not refusals.
For differences between tiers or models, we report 95\% intervals from a paired bootstrap over templates (Appendix \ref{sec:appendix_eval_metrics}).
Unless otherwise specified, we report means and standard deviations over three runs. 

\section{Results and Discussion}
\label{sec:results_discussion}

\begin{figure*}[t]
    \centering
    \includegraphics[width=\textwidth]{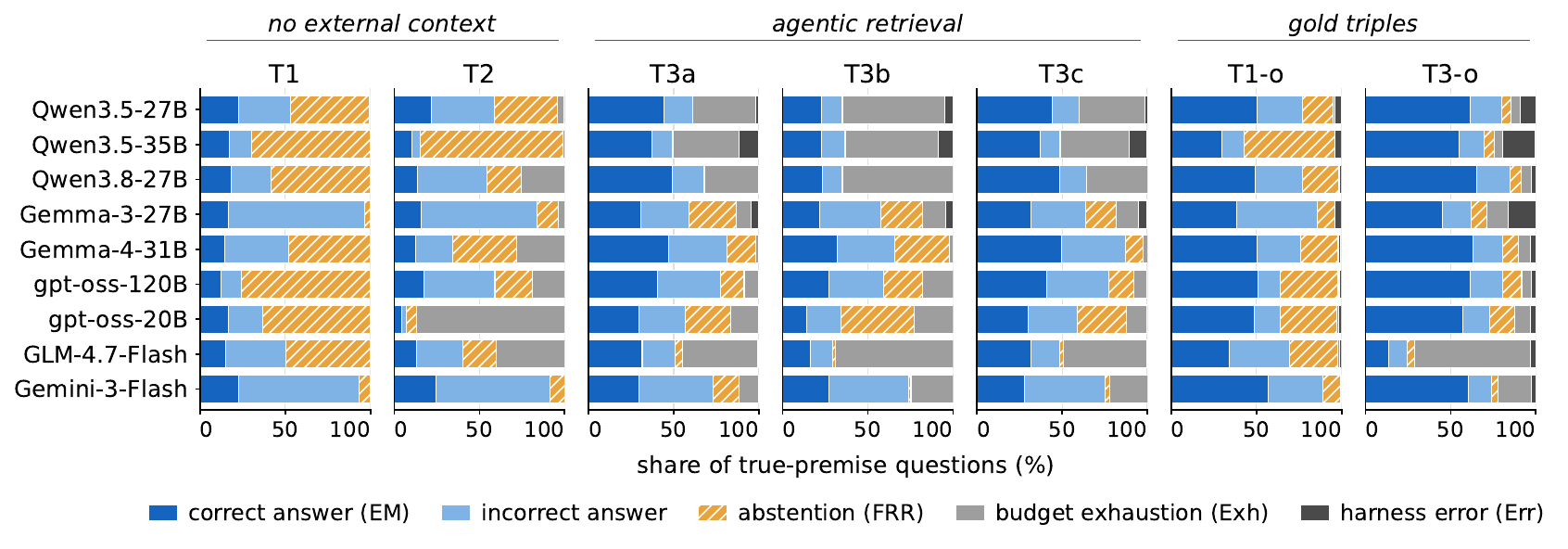}
    \caption{\textbf{Results by model and tier} (\emph{small} split, English, text modality; 4,979 \emph{true-premise} questions; means over runs). Each bar divides the questions into correct answers ($\mathrm{EM}$), incorrect answers, explicit abstentions ($\mathrm{FRR}$), budget exhaustion ($\mathrm{Exh}$) and harness errors ($\mathrm{Err}$); the first two segments together give the coverage ($\mathrm{Cov}$).}
    \label{fig:outcome_shares}
    \vspace{-0.5em}
\end{figure*}

We evaluate all models across evaluation tiers and along the dimensions \multiglobe stratifies. Unless stated otherwise, we report performance on the English text portion of the \textit{small} split, which is representative of the \textit{large} one after re-weighting to its spatial-function distribution (Appendix \ref{sec:appendix_representativeness}). 
We also evaluate six models on the 269 \emph{true-premise} image questions, where none reaches the majority baseline of 69.1 at any tier (Appendix \ref{sec:appendix_multimodal}).

\subsection{Performance Across Evaluation Tiers}
\label{sec:main_results}
No model exceeds the 28.8 $\mathrm{EM}$ majority baseline without retrieval or injected gold triples (Table \ref{tab:main_results}): T1 scores range from 12.4 (\gptossl) to 22.7 (\qwen, \gemini).
Boolean questions account for 63.1\% to 90.8\% of each model's correct answers at T1, where only \qwen answers them more accurately than a constant ``false'' and $\mathrm{EM}$ on the other questions is at most 5.9 for the open-weight models and 11.6 for \gemini (Table \ref{tab:appendix_boolean_split}).
Reasoning before answering (T2) does not close that gap. It lowers $\mathrm{EM}$ for seven of the nine models (Table \ref{tab:appendix_tier_gaps}), and the largest loss (\gptosss, -12.1) coincides with 86.9\% of completions cut off at the output-token limit (Fig. \ref{fig:outcome_shares}). Gold answers are computed from geometry and are not stated in text, so reasoning alone has no memorized answer to retrieve.
Retrieval reverses this. \kg retrieval raises $\mathrm{EM}$ by 12.9 (\gptosss) to 32.3 (\gemmanew) points for the open-weight models and lifts all nine models above the baseline. It outperforms web search for the open-weight models by 9.0 (\gemma) to 25.8 (\qwennew) points, as with web search these models answer fewer questions, and answer them less accurately (coverage up to 33.0 and $\mathrm{EM}_{\mathrm{cov}}$ up to 14.0 points lower; Fig.~\ref{fig:outcome_shares}, Table \ref{tab:main_results}). Combining the two changes their $\mathrm{EM}$ by at most 2.8 points (\gemmanew), indicating that web search is largely redundant once the \kg supplies the facts.

\rparagraph{Abstention}
Models differ more in whether they answer than in whether they are right when they do, at every tier except the two with \kg retrieval (T3a and T3c), where both vary about equally. At T1, coverage ranges from 24.1\% (\gptossl) to 96.5\% (\gemma) and $\mathrm{EM}_{\mathrm{cov}}$ from 17.3 (\gemma) to 57.4 (\qwenmoe; Table \ref{tab:main_results}, Fig. \ref{fig:outcome_shares}). \gemini hardly abstains but is frequently wrong, whereas \qwenmoe prefers abstaining.
The models abstain more often on the spatial-function families that are harder under gold triples. The rank correlation between  $\mathrm{FRR}$ and family accuracy under gold triples is negative for all nine at T1 and T2, and under \kg retrieval for the five open-weight models that abstain on more than 1\% of the questions, whereas \gemini shows no relation (Appendix \ref{sec:appendix_abstention}).

\begin{figure*}[t]
    \centering
    \begin{subfigure}[t]{0.48\textwidth}
        \centering
        \includegraphics[width=\linewidth]{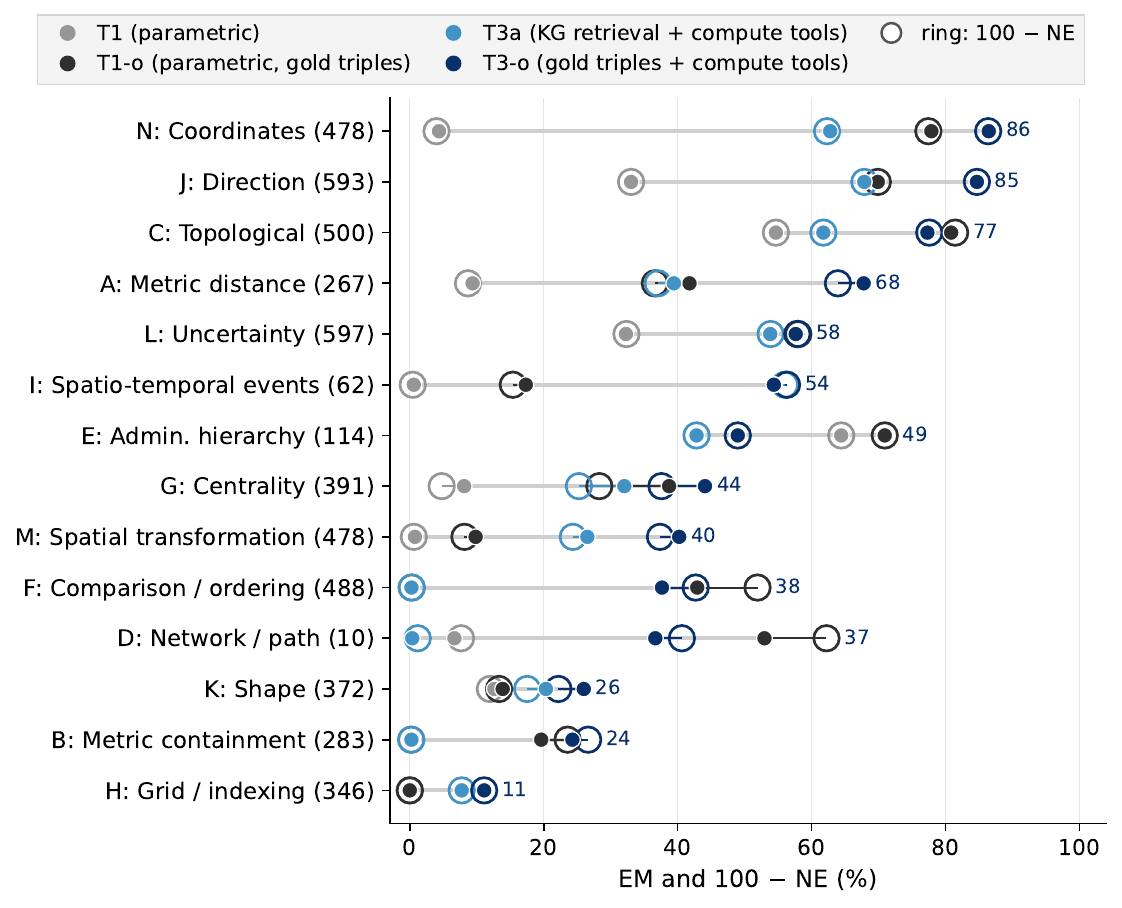}
        \caption{$\mathrm{EM}$ and $\mathrm{NE}$ by spatial function.}
        \label{fig:sfid_performance_em_overall}
    \end{subfigure}
     ~
    \begin{subfigure}[t]{0.48\textwidth}
        \centering
        \includegraphics[width=\linewidth]{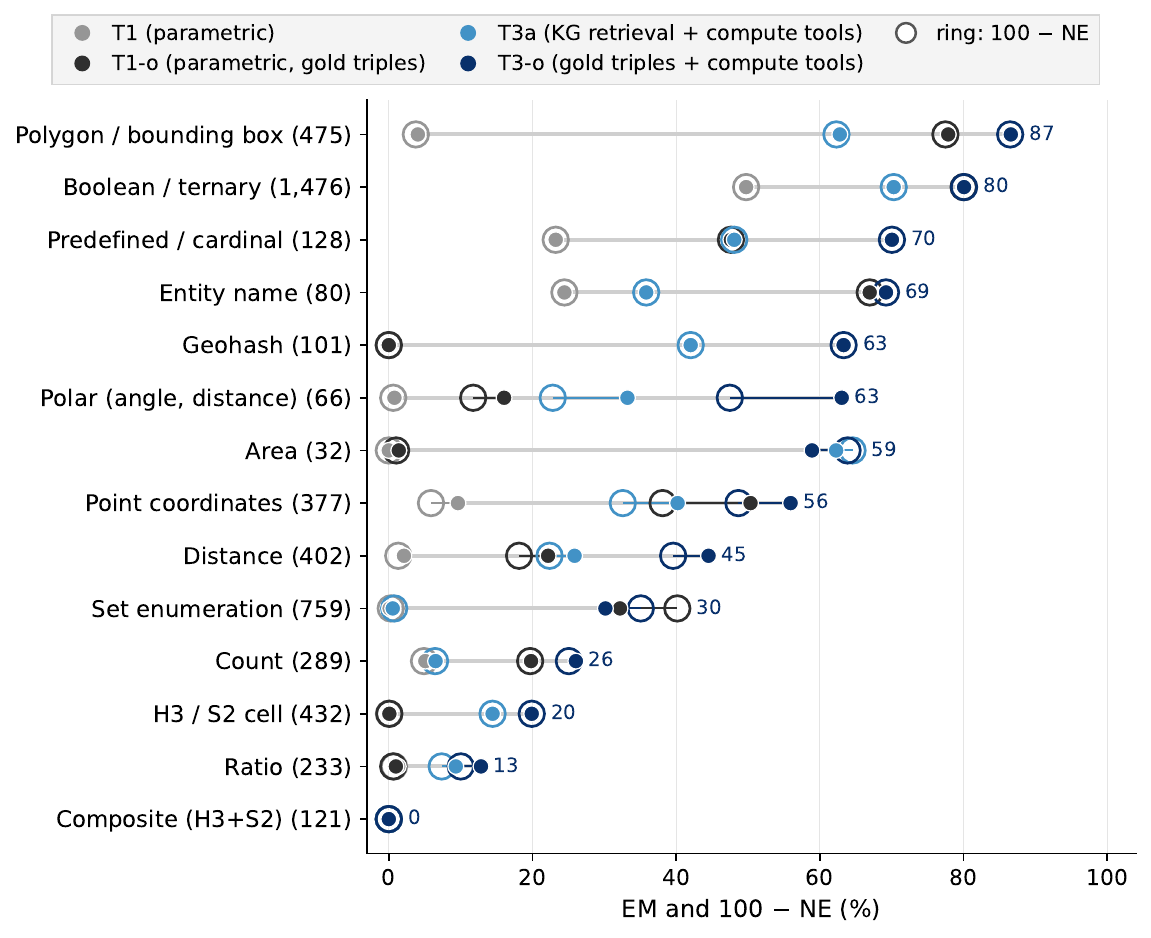}
        \caption{$\mathrm{EM}$ and $\mathrm{NE}$ by answer format.}
        \label{fig:answer_format_em_overall}
    \end{subfigure}%
    \caption{\textbf{Accuracy ($\mathrm{EM}$) and normalized error ($\mathrm{NE}$) by (a) spatial function and (b) answer format, across evaluation conditions} (\emph{small} split, English, text modality; averaged over the nine models). Dots mark $\mathrm{EM}$ and rings $100{-}\mathrm{NE}$ of the same condition, so a ring apart from its dot shows where the two metrics diverge. Parentheses give the number of \emph{true-premise} questions in each row; the T3-o $\mathrm{EM}$ is printed at the right of each row. We omit \emph{Date} as it covers only 8 questions (b).}

    \label{fig:rem_sfid_answer}
    \vspace{-0.5em}
\end{figure*}

\rparagraph{Oracle Conditions}
Both oracle conditions supply the same gold triples, verbalized in the prompt at T1-o and structured for the tool-using agent at T3-o. Even with tools, accuracy caps at 65.3 (\qwennew, statistically tied with \gemmanew and \gptossl; Table \ref{tab:appendix_model_differences}), leaving 34.7\% to 54.7\% of questions without a correct answer under perfect retrieval for eight models, while \glm reaches the step cap on 66.0\% of its T3-o questions and scores 13.5. Native thinking with a larger per-step output budget lifts \qwennew and \gemmanew only to 67.6 and 67.8, respectively, so no model exceeds 68 $\mathrm{EM}$ even when all required facts are supplied (Table \ref{tab:appendix_thinking}).

Injecting the triples lifts parametric accuracy 1.7$\times$ (\qwenmoe) to 4.1$\times$ (\gptossl), and \kg retrieval recovers only 51.4\% (\gptosss) to 75.0\% (\qwennew) of the accuracy at T3-o for the open-weight models other than \glm. We attribute the remainder largely to locating the entities and their facts in a large \kg within the step budget, independently of the spatial reasoning targeted.
Adding tools to the same gold triples (T3-o vs. T1-o) raises accuracy by 7.1 (\gemma) to 25.5 (\qwenmoe) points for seven open-weight models (\glm: -20.6) but only 3.2 for \gemini, which computes in context where the others depend on execution, and its lead of 6.0 points over the best open-weight model at T1-o disappears at T3-o.

The T3-o contexts exceed the context windows we serve for \qwen, \qwenmoe and \gemma on 8.9\% to 19.6\% of questions, against at most 3.0\% for the other models. On each model's questions without a harness error, \qwenmoe reaches 68.4, the highest of the nine, so the oracle ranking partly reflects context capacity. On the 3{,}619 questions without a harness error for any model, the best models, \gemmanew (75.2) and \qwennew (74.9), still leave a quarter of them without a correct answer, so accuracy is limited by computation over the evidence.

\rparagraph{Frontier Probe}
We evaluate a frontier model, \fable, using its own agent harness, on 109 questions with gold triples, stratified by spatial-function family and difficulty. \fable reaches 73.4 $\mathrm{EM}$, whereas the best evaluated model on these questions, \qwennew, reaches 52.9 at T3-o and 64.2 with thinking, and both gaps are statistically significant (Appendix \ref{sec:appendix_frontier_probe}). 

\rparagraph{Compute Cost}
\kg retrieval multiplies median tokens per question by 39 (\gptossl) to 436 (\glm). Comparing T3a with T3-o isolates retrieval costs: all models but \glm reach higher accuracy at T3-o with 1.3 to 9.4 times fewer tokens, and the Qwen models exhaust their budget on 31.9\% to 38.4\% of retrieval questions against 4.7\% to 6.2\% there (Appendix \ref{sec:appendix_cost}).

\subsection{Error Analysis}
\label{sec:error_analysis}
Under gold context, accuracy spreads by 75 points across the 14 spatial-function families (Fig. \ref{fig:sfid_performance_em_overall}; per model in Appendix \ref{sec:appendix_performance_sfid_answer}): coordinates reach 86, direction 85, shape 26, metric containment 24 and grid indexing 11. More than half of the metric-containment questions ask for the population within a radius, which the models almost never answer correctly even with gold triples. At T1, the four highest-scoring families (administrative hierarchy 64.4, topological relations 54.7, direction 33.0, uncertainty 32.3) are those with 72.2\% to 79.8\% Boolean or ternary questions, so the T1 ranking largely reflects the answer format.
Discrete spatial encodings do not fail uniformly (Fig. \ref{fig:answer_format_em_overall}): geohash recovers to 63.3 under gold context, H3 and S2 cell indices only 19.9, and their composition is never scored correct in any condition. Geohash interleaves latitude and longitude bits, whereas H3 and S2 require projection onto an icosahedral or spherical-cube grid and hierarchical cell arithmetic. Grid indexing ranks last, as all of its questions require an H3/S2 or composite answer.

Across all models and conditions, $\mathrm{NE}$ is 0 for 24.5\% and 1 for 66.3\%, leaving 9.2\% with partial credit: when a model is wrong, it is typically wrong by a wide margin. $\mathrm{NE}$ therefore tracks $100{-}\mathrm{EM}$ closely (rings in Fig. \ref{fig:rem_sfid_answer}; Table \ref{tab:appendix_further_metrics}), lying within one point of $100{-}\mathrm{EM}$ for eight of the 14 answer formats with at least 10 questions and departing by more than three only for five of the eight graded ones. Among these, set enumerations are near misses, while point coordinates that pass the tolerance retain a large residual error.

\begin{figure*}[t]
    \centering
    \includegraphics[width=\textwidth]{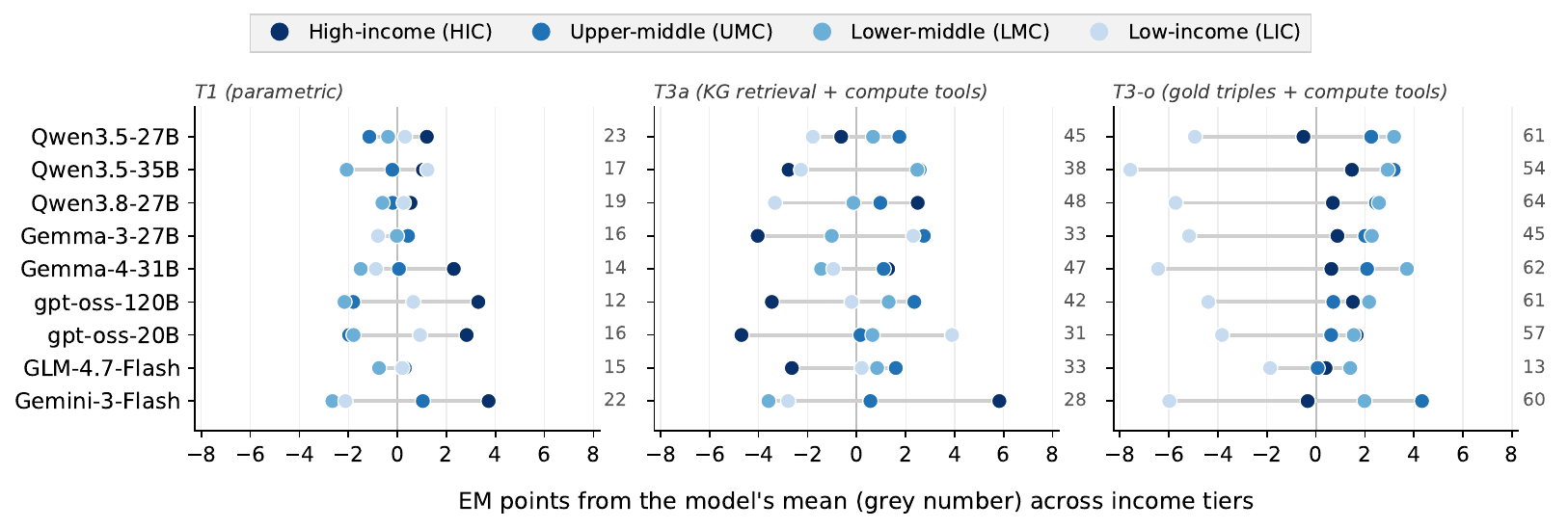}
    \caption{\textbf{Standardized $\mathrm{EM}$ by income tier} (\emph{small} split, English, text modality; 4{,}979 \emph{true-premise} questions).}
    \label{fig:income_equity_small}
    \vspace{-1em}
\end{figure*}

\rparagraph{Failure Analysis}
Under \kg retrieval, the models answer on average 26.5\% of the questions wrongly, abstain on 11.6\% and exhaust the budget on 22.1\%, so unanswered questions outnumber wrong answers (Fig. \ref{fig:outcome_shares}, Table \ref{tab:appendix_failure_analysis}). Exhaustion concentrates in the three Qwen models and \glm (31.9\% to 44.3\%) and abstention in the four other open-weight models (14.0\% to 27.5\%). For the eight models whose traces show tool replies, a \kg lookup returns no result on at least 17.1\% to 43.5\% of the T3a questions, and a code step fails on 22.3\% to 59.1\%. With gold triples, harness errors rise from 2.1\% to 6.8\% on average, mostly through context overflow, and on 61.1\% of the grid-indexing questions the agent calls a nonexistent, mostly outdated \texttt{h3} function.

\rparagraph{Question Complexity}
$\mathrm{EM}$ does not decrease steadily with the number of gold facts and peaks at 9 to 16 facts for all models at T3a and T3-o (Table \ref{tab:appendix_complexity}). Questions with 17 or more facts (11.1\%), mostly set enumerations, are where retrieval fails most often, as the models average 2.2 at T3a against 35.9 at T3-o. Polygon questions score on average 19.8 points below point questions at T3-o and 11.6 below at T3a, so the difficulty of polygon geometries remains when the facts are supplied.

\subsection{Stratified Analysis}
\label{sec:stratified_analysis}

\rparagraph{Income Tiers}
We next analyze accuracy by income tier, standardizing each tier to the split-wide spatial-function distribution while holding its own per-family accuracy fixed.\footnote{Raw $\mathrm{EM}$ is not comparable across tiers, because the high-income tier contains a larger share of questions from the hardest spatial-function families.}
At T1, eight of the nine models answer questions about high-income regions more accurately than those about low-income regions ($-$0.2 \qwenmoe to +5.8 \gemini; Fig.~\ref{fig:income_equity_small}). Under \kg retrieval the gap has no consistent direction, positive for four models and negative for five ($-$8.6 \gptosss to +8.6 \gemini). With gold triples it is positive for all nine (+2.3 \glm to +9.0 \qwenmoe) and wider than at T1 for every open-weight model. The low-income tier is then the lowest for every model, while a middle-income tier is the highest for eight of the nine, so accuracy does not rise steadily with income. The gold contexts are themselves uneven, at the median 1.4 times as long for high-income as for low-income questions, so evidence volume and regional familiarity cannot be separated.\footnote{Fig.~\ref{fig:appendix_density_equity_small} shows the same analysis by population-density tier, and Fig.~\ref{fig:appendix_equity_large} shows both at T1 on the \textit{large} split.}

\rparagraph{False-Premise Questions}
On the 13 sub-templates with false premises (Fig. \ref{fig:appendix_false_premise}), the models do not distinguish false-premise questions from answerable ones without retrieval, as they refuse answerable questions more often than they reject false premises in 14 of 18 model-tier cells. Under \kg retrieval, \gemma, \gemmanew, and the two gpt-oss models reject 61.7\% to 83.6\% of the false premises and refuse only 11.0\% to 26.5\% of the answerable questions. The Qwen models and \glm exhaust their budget on 75.3\% to 97.5\% of the false-premise questions, which keeps both rates at or below 21.0\%.

\rparagraph{Languages}
As \multiglobe's 17 languages are translations of one fixed question set, a per-language comparison isolates language from question difficulty. Accuracy is stable: at T1 the best-to-worst spread is at most 5.7 points for the five models evaluated in all languages, and 3.9 to 6.1 for \qwen at T1-o, T3a and T3-o (Fig. \ref{fig:appendix_multilingual}).
$\mathrm{FRR}$ varies more at T1, by up to 11.2 points across languages.
These comparisons vary only the question text, as the instructions, answer format, and tool catalog remain English, and thus isolate sensitivity to question surface form rather than end-to-end multilingual competence. The largest deviation from English is Georgian's, 5.8 points below at T1-o and 3.8 at T3a for \qwen.

\section{Conclusion}
\label{sec:conclusion}

We introduce \multiglobe, a benchmark of 46,060 geospatial questions spanning 14 spatial-function families and 15 answer formats, stratified by income and population density over 201 countries and territories, and released parallel in 17 languages with false-premise and multimodal slices. We evaluate nine models across parametric, reasoning, and agentic tiers. Retrieval and tools account for nearly all the gains, but with gold triples and compute tools no model exceeds 68 $\mathrm{EM}$ and grid indexing averages 11.1, so computation is the bottleneck. Under gold triples, every model also scores lower on low-income than on high-income regions. Progress in geospatial reasoning therefore depends on models that perform spatial computation reliably, in context or through tools, more than on additional geographic knowledge.

\section*{Limitations}
\label{sec:limitations}

Template coverage is bounded by query authoring. Each template requires a hand-written SQL shape and a matching gold triple query, so a new spatial function or \kg requires a new query, while further questions follow from sampling entities that satisfy an existing query. In return, correctness is decided by executing a query and requires no \llm judge.

Ground truth is computed against a fixed snapshot of each \kg and inherits its gaps and errors. Missing entities affect questions whose answer ranges over a set of entities, such as counts, set enumerations, and nearest-entity questions, since the query still executes but returns an incomplete answer, and inaccurate geometries or mislabeled feature types propagate to the gold answers. We release the gold triples for each question so that individual answers remain auditable. Snapshots also become outdated, so a model retrieving current information through web search may be scored against a gold answer that no longer holds. Re-executing the queries against a more recent snapshot would quantify this drift.

Frontier closed models have been reported to reason more reliably and use tools and search more effectively than open-weight ones \citep{chen2025browsecomp,golikov2026robust,fan2026agentprocessbench}. Our academic budget limits the full protocol to eight open-weight models from four families, served locally, and one closed model from the small, fast tier of its family. This closed model is not uniformly stronger, as \gemini has the highest $\mathrm{EM}$ at T2 and T1-o, while an open-weight model leads at every agentic tier (Table \ref{tab:main_results}).
\citet{krechetova2025geobenchx} likewise find that the newest of eight commercial models is not the most accurate on multistep geospatial agent tasks. We evaluate a frontier model, \fable, only as a probe on 109 questions with gold triples, in a single run and in its own agent harness (Table \ref{tab:appendix_frontier_probe}). It outperforms the best evaluated model on these questions, which are not a random sample of the benchmark, and we therefore cannot rule out that larger or newer models would close the gaps we report.

We fix the answer format to the gold answer type, through a JSON schema at T1, T2, and T1-o, and validation of the final answer at the agentic tiers. This isolates spatial reasoning from format compliance and reduces format errors, but it also reveals the shape of the answer. \multiglobe therefore does not test whether a model can produce a spatial answer in free-form text or select the appropriate representation, which we leave to future work.

\section*{Ethical Considerations}
\label{sec:ethical_considerations}

\multiglobe is derived from openly licensed sources (Appendix \ref{sec:appendix_benchmark_construction}), and images in the multimodal slice come from Wikidata and Wikimedia Commons through the entities' own image links. Entities represent public geographic features and points of interest, so neither the questions nor the released triples contain personal data. Annotators for the 16 target languages were recruited through the authors' academic networks and among professional translators, gave informed consent, and were appropriately compensated (Appendix \ref{sec:appendix_human_annotation_protocol}).

The coverage disparities we analyze are properties of the underlying \kgs and are inherited by any benchmark built on them. Stratified sampling spreads questions across all income and density strata, but the entity pool remains denser in high-income regions, and models tuned to \multiglobe may inherit this uneven coverage. Aggregate accuracy therefore does not establish uniform performance across regions. By reporting accuracy per income tier, population density, and language, \multiglobe makes such disparities measurable.
Spatial errors affect the applications that motivate this work, such as navigation, logistics, and disaster response. Current models fail on computations that these applications require, so a high score on \multiglobe is not sufficient evidence of reliability in them. Agentic evaluation multiplies the median tokens per question by at least 39 relative to direct answering (Appendix \ref{sec:appendix_cost}), and the \textit{small} split lowers this cost.

\section*{Acknowledgment of AI Use}
\label{sec:ai_use}

In this work, we used generative AI tools for the following purposes: (i) data processing and filtering, (ii) implementing the annotation interface, the evaluation harness, and standard methods, (iii) supporting data analysis, (iv) visualizing results for submission, (v) drafting and editing parts of the paper (e.g., grammar, spelling, word choice, concision), and (vi) preparing code and data releases. The \llms that are part of the benchmark construction, namely the machine translation system, the judge for annotator disagreements and the post-editing ensemble, are described in \S\ref{sec:multilingual_extension} and \S\ref{sec:postediting_questions}, and the evaluated models in \S\ref{sec:experimental_setup}. The judge only chose between the versions of the two annotators, so no \llm annotation replaced the human annotation. We designed the benchmark and experiments, ran all evaluations, reviewed, edited, and verified all AI-assisted output, and retain full responsibility for the final content of the manuscript.
\section*{Acknowledgments}
\label{sec:acknowledgments}

This work was supported by the SpatialBenchRAG project grant of the Open Science Office of the University of Mannheim, by the Google Cloud Research Credits program with the award EDU460771163, and by the state of Baden-Württemberg through bwHPC.
\section*{Author Contributions}
\label{sec:author_contributions}

Andreea Iana conceived the research contributions, obtained project funding, designed and coordinated the human annotation study, ran the experiments and analyses, and wrote the final version of the paper. Martin Böckling contributed to the initial research idea and an early draft of the manuscript, and built and ran the data generation pipeline. Andreea Iana, Martin Böckling, and Elizaveta Nosova designed the templates and curated the data. Elizaveta Nosova refined the templates and their constraints, and helped recruit the annotators. Heiko Paulheim provided feedback and institutional support throughout the project.

\bibliography{custom}

\clearpage
\appendix
\makeatletter
\setlength\@fptop{0pt}\setlength\@fpsep{12pt plus 2pt minus 2pt}\setlength\@fpbot{0pt plus 1fil}
\setlength\@dblfptop{0pt}\setlength\@dblfpsep{12pt plus 2pt minus 2pt}\setlength\@dblfpbot{0pt plus 1fil}
\makeatother
\onecolumn
\section{Benchmark Construction}
\label{sec:appendix_benchmark_construction}

\subsection{Full Template Listing}
\label{sec:appendix_benchmark_templates}

Table \ref{tab:appendix_templates} lists all 65 templates with an example and answer format. 
\begingroup
\scriptsize
\setlength{\tabcolsep}{3pt}
\setlength{\LTleft}{0pt}
\setlength{\LTright}{0pt}
\setlength{\LTcapwidth}{\textwidth}
\renewcommand{\arraystretch}{1.15}
\newcommand{\templategroup}[1]{%
  \rowcolor{gray!15}\multicolumn{4}{@{}>{\raggedright\arraybackslash}p{\dimexpr\textwidth-2\tabcolsep\relax}@{}}{\textbf{#1}}\\}

\begin{longtable}{@{}>{\raggedright\arraybackslash}p{1.85cm}
 >{\raggedright\arraybackslash}p{\dimexpr0.5\textwidth-2.1cm\relax}
 >{\raggedright\arraybackslash}p{\dimexpr0.5\textwidth-2.1cm\relax}
 >{\raggedright\arraybackslash}p{1.65cm}@{}}

\toprule
\textbf{Sub-function} & \textbf{Template} & \textbf{Example} & \textbf{Answer type}\\
\midrule
\endfirsthead

\multicolumn{4}{@{}>{\raggedright\arraybackslash}p{\dimexpr\textwidth-2\tabcolsep\relax}@{}}{\footnotesize\itshape Table~\ref{tab:appendix_templates} continued from previous page.}\\
\toprule
\textbf{Sub-function} & \textbf{Template} & \textbf{Example} & \textbf{Answer type}\\
\midrule
\endhead

\midrule
\multicolumn{4}{@{}>{\raggedleft\arraybackslash}p{\dimexpr\textwidth-2\tabcolsep\relax}@{}}{\footnotesize\itshape Continued on next page.}\\
\endfoot

\bottomrule

\addlinespace[10pt]
\caption{\textbf{The 65 top-level templates in \multiglobe}, grouped by Spatial Function ID (\sfid{}). Per template we report the sub-function category, canonical English template with placeholders, one example instantiation, and answer type. Each template expands into one or more parameter-bound sub-templates.}
\label{tab:appendix_templates}
\endlastfoot

\templategroup{SFID A --- Metric Distance \& Proximity Computation / Reasoning}
Direct distance & What is the distance between \{entity\_A\} and \{entity\_B\}? & How far is Mbélé from Bilsem? & Distance\\
 & How far \{cardinal\_direction\} is \{entity\_A\} from \{entity\_B\}? & How far west is Hov from Olsby? & Distance\\
\hdashline

Event-to-location & How far away was the \{event\_A\} from \{entity\_B\} of \{entity\_C\} in \{time\_period\}? & How far away was the "Earthquake from USGS Earthquake Catalog with ID pr2020052043" from Farmacia Asturias LLC of Puerto Rico from 2020-01-01 to 2023-12-31? & Distance\\
\hdashline

Comparative & Is \{entity\_A\} or \{entity\_B\} \{rank\_type\} to \{entity\_C\}? & Is Behruz or Payawak closer to Sa tapa? & Entity name (single)\\
 & Which is closer to the country that borders the country containing \{entity\_A\}: \{entity\_B\} or \{entity\_C\}? & Which of Uma Boco or Dilor is nearer to a neighboring country of Jadin’s country? & Entity name (single)\\
\hdashline

Aggregation & What is the \{aggregation\_type\} distance between \{entity\_A\} in \{entity\_B\}? & What is the average distance between hotels in Pedasí? & Distance\\
\hdashline

Optimization & Which \{entity\_A\} in \{entity\_B\} minimizes total distance to all \{entity\_C\}? & Which Pharmacy in Lucas County, Ohio is closest overall to all BPHC\_Sites? & Entity name (single)\\
\hdashline

Distance-based selection & What is the distance between the \{rank\_position\} nearest \{entity\_A\} to the \{entity\_B\} and the \{entity\_C\}? & How far is the second closest Market to the 7-11 from 24 Bar? & Distance\\
\hdashline

Boundary distance & What is the minimum distance from \{entity\_A\} to the boundary of \{entity\_B\}? & What is the minimum distance from Koibi to the boundary of Metekel? & Distance\\

\templategroup{SFID B --- Metric Containment}
Counting & How many \{entity\_A\} are within \{value\} km of \{entity\_B\}? & How many pharmacies are within 5 km of Siren? & Count\\
\hdashline

Enumeration & Which \{entity\_A\} are within \{value\} km of \{entity\_B\}? & Which pharmacies are within 5 km of KFC? & Set enumeration\\
\hdashline

Spatial join & What is the total population of the area within \{value\} km of \{entity\_A\}? & How many people live within 5 km of Choco? & Count\\

\templategroup{SFID C --- Topological Relationships}
Intersection (boolean) & Does \{entity\_A\} intersect \{entity\_B\}? & Do Ocelot and Karon intersect spatially? & Boolean \\
 & Do the geometries between \{entity\_A\} intersect with the geometry of \{entity\_B\}? & Do the geometries of Basco intersect with the geometry of Nusa? & Boolean\\
\hdashline

Intersection (count) & In how many points do \{entity\_A\} and \{entity\_B\} intersect? & In how many points do Lystvej and Rudbølvej intersect? & Count\\
 & How many \{entity\_A\} intersect \{entity\_B\}? & How many highways intersect Jacros? & Count \\
\hdashline

Intersection (enumeration) & Which \{entity\_A\} intersect \{entity\_B\}? & Which highway intersects Calle 1? & Set enumeration\\
 & Which \{entity\_A\} intersect \{entity\_B\} but not \{entity\_C\}? & Which highway intersects Itbayat but not Raele? & Set enumeration\\
\hdashline

Overlap & Do \{entity\_A\} and \{entity\_B\} overlap spatially? & Do Dano and Fafan overlap spatially? & Boolean\\
 & Does \{entity\_A\} touch the boundary of \{entity\_B\}? & Does Kumejima touch the boundary of Suzu? & Boolean\\
\hdashline

Adjacency & Is \{entity\_A\} adjacent to \{entity\_B\}? & Is Utiroa adjacent to Roreti? & Boolean\\
\hdashline

Disjointness & Is \{entity\_A\} physically disconnected from \{entity\_B\}? & Are Klepalo and Grad spatially disjoint? & Boolean\\
\hdashline

Enclosure & Does \{entity\_A\} completely surround \{entity\_B\}? & Does Basey completely surround Maragat? & Boolean\\

\templategroup{SFID D --- Network \& Path-Based Spatial Reasoning}
Traversal & Which \{entity\_A\} does the \{entity\_B\} traverse? & Which administrative regions does the Carretera 10 traverse? & Set enumeration\\
 & How many \{entity\_A\} does the \{entity\_B\} traverse? & How many administrative regions does the Promachonas - Agistro pass through? & Count\\
\hdashline

Path proximity & Which \{entity\_A\} does the \{entity\_B\} pass by? & Which cafe does the Pont d'Andotsy pass by? & Set enumeration\\
 & How many \{entity\_A\} does the \{entity\_B\} pass by? & How many cafes does the General Espejo pass by? & Count\\

\templategroup{SFID E --- Containment \& Administrative Hierarchy}
Membership & Is \{entity\_A\} administratively a part of \{entity\_B\}? & Is Itbayat administratively a part of Basco? & Boolean\\
\hdashline

Counting & How many \{entity\_A\} make up \{entity\_B\}? & How many components make up Jerma? & Count\\
 & How many \{collection\_of\_entity\} are in \{entity\_A\}? & How many bars, cafes and restaurants are in Dano? & Count\\
\hdashline

Full containment & Does the administrative boundary of \{entity\_A\} fully contain the area of \{entity\_B\}? & Does the administrative boundary of Siaton fully contain the area of Salag? & Boolean\\
 & Does the administrative border of \{entity\_A\} fully contain the \{rank\_type\} \{entity\_B\} of \{entity\_C\}? & Does the administrative border of Arege fully contain the longest Yonga of Teungo? & Boolean\\

\templategroup{SFID F --- Spatial Comparison \& Ordering}
Extremes & Which \{entity\_A\} lies furthest \{cardinal\_direction\} within \{entity\_B\}? & Which Pharmacy lies furthest west within Thomaston, GA? & Entity name (single)\\
\hdashline

Ranking & What is the \{rank\_type\} \{entity\_A\} in \{entity\_B\} by \{attribute\_A\}? & What is the biggest ZipCodeArea in Big Rapids, MI by elongation? & Entity name (single)\\
 & What is the \{rank\_position\} \{attribute\} \{entity\_A\} in \{entity\_B\}? & What is the longest RoadSegment in Alexandria, MN? & Entity name (single)\\
\hdashline

Compound comparison & Which \{entity\_A\} are within \{value\} km of \{entity\_B\} and \{cardinal\_direction\} of \{entity\_C\}? & Which cafes are within 10 km of OMV and south of Viva? & Set enumeration\\

\templategroup{SFID G --- Centrality \& Medial Geometry}
Centroid / medoid & What is the most central point of \{entity\_A\}? & What is the most central point of Ivana? & Point coordinates\\
\hdashline

Entity centrality & What is the most central point of \{entity\_A\} where \{entity\_B\} is located? & What is the most central point of Walmara where Kolobo is located? & Point coordinates\\
\hdashline

Scale sensitivity & How does the centroid of \{entity\_A\} change when computed at spatial resolutions \{value\_1\} km and \{value\_2\} km? & How does the centroid of Øyer change when computed at spatial resolutions 10 km and 100 km? & Distance\\

\templategroup{SFID H --- Grid \& Spatial Indexing}
Spatial unit containment & Which grid cell contains both \{entity\_A\} and \{entity\_B\}? & Which grid cell contains both DHL and Auto Will? & H3 / S2 cell index\\
 & Which grid cell at resolution \{value\} contains both \{entity\_A\} and \{entity\_B\}? & Which grid cell at resolution 7 contains both Dy Cafe and KFC? & H3 / S2 cell index\\
\hdashline

Multi-indexing & Which grid cell contains \{entity\_A\} both in h3 and s2 formats? & Which grid cell contains Grad both in h3 and s2 formats? & Composite (H3$+$S2)\\

\templategroup{SFID I --- Spatio-Temporal Reasoning: Spatio-Temporal Events}
Event distance & How far away was the \{event\_A\} from \{entity\_B\} in \{time\_period\}? & During 2020-01-01 to 2023-12-31, how far was "Earthquake from USGS Earthquake Catalog with ID ci40019119" from AIPHARM LLC? & Distance\\
\hdashline

Occurence count & Which \{entity\_A\} were affected by \{event\_A\}? & Which regions were affected by Earthquake from USGS Earthquake Catalog with ID nn00793499? & Set enumeration\\
\hdashline

Temporal extent & When did \{event\_A\} hit \{entity\_A\}? & When did "Earthquake from USGS Earthquake Catalog with ID hv357401" hit Hilo, HI? & Date\\
\hdashline

Spatial impact & How large is the area affected by \{event\_A\}? & How large is the area affected by "Smoke plume snapshot with index 8 on date 2018-12-22"? & Area \\
 & How large is the area affected by \{event\_A\} in \{entity\_A\}? & How large is the area affected by "SELLEM fire (from the MTBS dataset) that occurred in 1999-08-28 with ID NV3829211543619990828" in Pahrump, NV? & Area\\

\templategroup{SFID J --- Directional \& Orientation Reasoning}
Cardinal & Is \{entity\_A\} \{cardinal\_direction\} of \{entity\_B\}? & Is BCA east of Duquinha? & Boolean\\
\hdashline

Relative to path / location & Is \{entity\_A\} to the left or right of the route from \{entity\_B\} to \{entity\_C\}? & Is Fried Chicken to the left or right of the route from PNB to Burger King? & Predefined\\
 & Does \{entity\_A\} lie upstream or downstream of \{entity\_B\}? & Does Gooddays lie upstream or downstream of 7-Eleven? & Predefined\\
\hdashline

Angular orientation & What direction would you travel from \{entity\_A\} to reach \{entity\_B\}? & What direction would you travel from Lokenie to reach Kiyungi? & Cardinal direction\\

\templategroup{SFID K --- Shape \& Geometry Properties}
Area comparison & Is \{entity\_A\} larger in area than \{entity\_B\}? & Is Lurøy larger in area than Røros? & Boolean\\
 & Is \{entity\_A\} more compact than \{entity\_B\}? & Is Afder more compact than Fafan? & Boolean\\
 & Is \{entity\_A\} more elongated than \{entity\_B\}? & Is Cagayan more elongated than Basco? & Boolean\\
\hdashline

Perimeter & What is the perimeter of \{entity\_A\}? & What is the perimeter of Dishu? & Distance\\
\hdashline

Reasoning about shape & How irregular is the boundary of \{entity\_A\}, measured by \{irregularity\_metric\} at \{scale\_value\} km \{scale\_unit\}? & How irregular is the boundary of Korahe, measured by shape index at 1.0 km simplification? & Ratio (dimensionless)\\

\templategroup{SFID L --- Uncertainty \& Fuzzy Spatial Reasoning}
Approximate distance & Is \{entity\_A\} approximately near \{entity\_B\}? & Is Terrasse approximately near Oasis? & Boolean\\
 & Is \{entity\_A\} far from \{entity\_B\}? & Is La ceiba far from Tienda? & Boolean\\
 & Is \{entity\_A\} more likely near \{entity\_B\} than \{entity\_C\}? & Is La plage more likely near Pukka than Lio Villas Resort? & Boolean\\
\hdashline

Vagueness & Which \{entity\_A\} are roughly closest to \{entity\_B\}? & Which Bank is roughly closest to Ecobank? & Set enumeration\\
\hdashline

Probabilistic containment & Is \{entity\_A\} plausibly within \{entity\_B\}? & Is Ugat plausibly within El Nido? & Ternary\\

\templategroup{SFID M --- Spatial Transformation \& Reference Change}
Relative position & What is the relative position of \{entity\_A\} in the coordinate system of \{entity\_B\}? & What is the relative position of Pukka in the coordinate system of Caltex? & Polar (angle, distance)\\
\hdashline

Conversion & Convert the location of \{entity\_A\} to \{grid\_type\} at \{resolution\_type\} \{value\} \{metric\}. & Convert the location of Ons to S2 at level 15. & Grid cell / coords\\

\templategroup{SFID N --- Spatial Reference \& Geometric Representation}
Absolute coordinate retrieval & What are the latitude and longitude of the centroid of \{entity\_A\}? & What are the latitude and longitude of the centroid of Ruse? & Point coordinates\\
\hdashline

Spatial extent representation & What is the bounding box of \{entity\_A\}? & What is the bounding box of Josue? & Polygon / bounding box\\
\end{longtable}

\endgroup

\begin{table*}[t]
\centering
    \small
    \begin{threeparttable}
    \begin{tabular*}{\textwidth}{@{\extracolsep{\fill}}l r r r@{}}
    \toprule
     & \textbf{\worldkg} & \textbf{\kwg} & \textbf{\osmkg} \\
    \midrule
    Source                 & \osm{} PBF\tnote{a} & GraphDB endpoint\tnote{b} & \osm{} PBF\tnote{a} \\
    License                & ODbL & CC-BY 4.0 & ODbL \\
    \hdashline
    \# Node types          & 137 & 17 & 175 \\
    \# Nodes               & 745,194,840 & 308,071,129 & 280,078,239 \\
    \# Geographic nodes    & 14,304,791 & 5,026,581 & 9,505,778 \\
    \# Leaves              & 560,306,023 & 264,469,472 & 187,064,086 \\
    \# Predicates          & 1,645 & 358 & 34,038 \\
    \# Edges               & 1,308,464,629 & 860,069,872 & 838,673,446 \\
    Mean / max out-degree  & 3.72 / 42 & 5.59 / 1,163,503 & 4.69 / 8,968,560 \\
    Mean / max in-degree   & 2.30 / 176{,}096{,}861 & 2.82 / 67,810,074 & 3.02 / 41,102,070 \\
    \bottomrule
    \end{tabular*}
    \begin{tablenotes}[flushleft]\footnotesize
      \item[a] Constructed in-house from a single \osm{} planet PBF snapshot (2025-12-08).
      \item[b] Crawled from \url{https://stko-kwg.geog.ucsb.edu/graphdb/repositories/KWG}; no longer publicly accessible (upstream GraphDB license lapsed). Data remains CC-BY~4.0; we redistribute the cached snapshot used for generation.
    \end{tablenotes}
    \end{threeparttable}
    
    \captionof{table}{\textbf{The three \kgs used to construct \multiglobe}. \textit{ODbL}: Open Data Commons Open Database License; \textit{CC-BY~4.0}: Creative Commons Attribution 4.0.}
    \label{tab:appendix_kgs}
\end{table*}
\begin{figure*}[t]
    \centering
    \includegraphics[width=\textwidth]{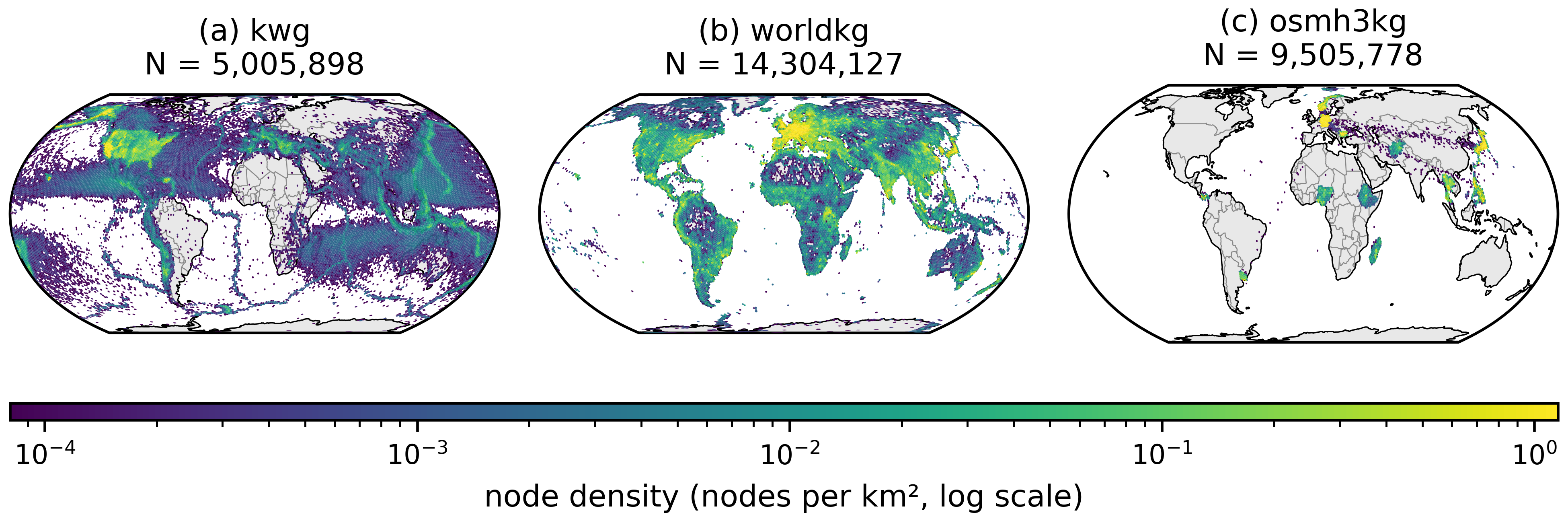}
    \caption{Geographic density of geometries  over (a) \kwg (KWG), (b) \worldkg, and (c) \osmkg.}
    \label{fig:kgs_geo_density}
\end{figure*}
\begin{figure*}[t]
    \centering
    \includegraphics[width=0.8\textwidth]{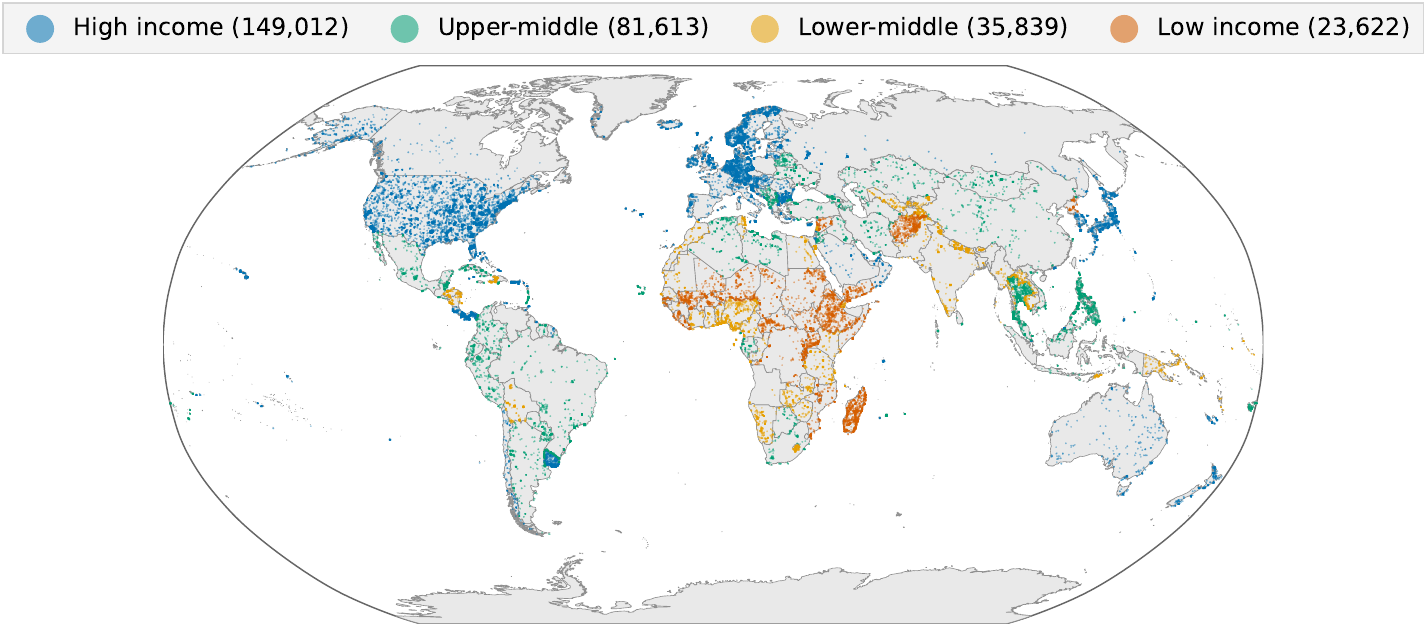}
    \caption{\textbf{Global distribution of \multiglobe entities.} 290,086 distinct entity sites across 201 countries and territories, colored by World Bank income tier.}
    \label{fig:appendix_entity_map}
\end{figure*}
\twocolumn

\subsection{Data Sources}
\label{sec:appendix_data_sources}
\begin{figure}[t]
    \centering
    \includegraphics[width=\columnwidth]{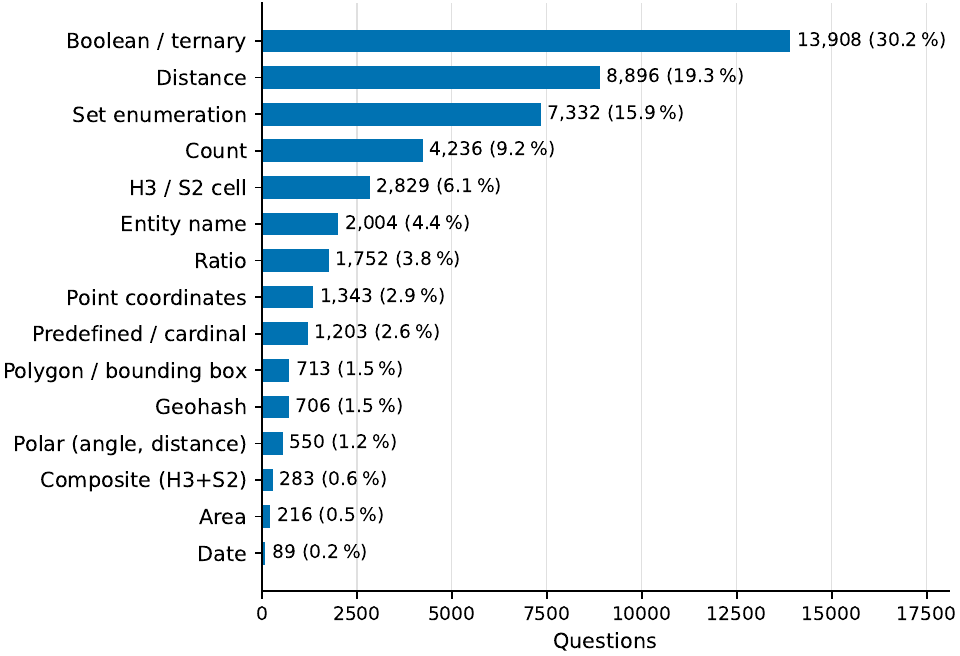}
    \caption{\textbf{Distribution of \multiglobe questions over the 15 answer formats.}}
    \label{fig:appendix_answer_format_dist}
\end{figure}

Table \ref{tab:appendix_kgs} reports provenance and statistics of each source \kg, Table~\ref{tab:appendix_per_kg_sfid_counts} the \sfid distribution per \kg, and Fig. \ref{fig:appendix_answer_format_dist} the answer-format distribution.
Fig. \ref{fig:kgs_geo_density} shows geometry density across the \kgs. 

\rparagraph{Entity-name Canonicalization}
On the English benchmark, we use each entity's English label where available, falling back to \texttt{rdfs:label} otherwise. We also preserve aliases from language-specific labels on the entity row and surface them in the alias store used for evaluation.

\rparagraph{Geographic Strata and Coverage}
Income tiers are assigned by spatially joining each entity's geometry with Natural Earth 10m country polygons\footnote{\href{https://www.naturalearthdata.com/downloads/10m-cultural-vectors/}{www.naturalearthdata.com/10m-cultural-vectors/}} and mapping the resulting ISO-3 to the World Bank FY2026 classification\footnote{\href{https://api.worldbank.org/v2/country}{https://api.worldbank.org/country}}. 
Density tiers aggregate the WorldPop R2025A 1km population raster \cite{tatem2017worldpop} to H3 resolution 3, and are binned at the global 33rd and 67th percentiles. 
Fig. \ref{fig:appendix_entity_map} shows the global distribution of distinct entity sites.
\begin{table}[t]
\centering
\footnotesize
\resizebox{\columnwidth}{!}{%
    \begin{tabular}{@{}lrrr@{}}
    \toprule
    \textbf{(\sfid) Spatial Function} & \textbf{\worldkg} & \textbf{\kwg} & \textbf{\osmkg} \\
    \midrule
        (A) Metric Distance & 4{,}537 & 1{,}225 & 4{,}046 \\
        (B) Metric Containment & 329 & 365 & 1{,}193 \\
        (C) Topological & 1{,}108 & 925 & 3{,}632 \\
        (D) Network / Path & -- & -- & 645 \\
        (E) Admin Hierarchy & -- & 348 & 3{,}686 \\
        (F) Comparison / Ordering & 2{,}295 & 954 & 2{,}178 \\
        (G) Centrality & -- & 214 & 616 \\
        (H) Grid / Indexing & 718 & 210 & 745 \\
        (I) Spatio-Temp.\ Events & -- & 477 & -- \\
        (J) Direction & 2{,}782 & 550 & 2{,}303 \\
        (K) Shape & -- & 680 & 2{,}433 \\
        (L) Uncertainty & 1{,}071 & 378 & 1{,}251 \\
        (M) Spatial Transformation & 1{,}669 & 479 & 1{,}302 \\
        (N) Coordinates & 331 & 94 & 291 \\
    \midrule
    \textbf{Total} & \textbf{14{,}840} & \textbf{6{,}899} & \textbf{24{,}321} \\
    \bottomrule
    \end{tabular}%
    }
    \caption{\textbf{Row counts per (\kg, \sfid{}) cell}, including the multimodal slice for \osmkg{} and \worldkg{}. Dashes (\textit{--}) mark \sfids with no data in that \kg.}
    \label{tab:appendix_per_kg_sfid_counts}
\end{table}

\subsection{Comparison with Related Datasets}

Table \ref{tab:appendix_related_benchmarks} compares \multiglobe with 10 other geospatial \qa and agent benchmarks.
\begin{table*}[t]
\centering
\scriptsize
\setlength{\tabcolsep}{2.5pt}
\newcolumntype{L}[1]{>{\raggedright\arraybackslash\hspace{0pt}}p{#1}}
\newcolumntype{Y}{>{\raggedright\arraybackslash\hspace{0pt}}X}
    
    \begin{tabularx}{\textwidth}{@{}Y Y Y Y Y Y Y Y@{}}
    \toprule
    
    \textbf{Benchmark} & \textbf{Size} & \textbf{Tasks \& Templates} & \mbox{\textbf{Languages}} & \textbf{Answer Provenance} & \textbf{Knowledge Source} & \textbf{Geographic coverage} & \textbf{Modalities} \\
    \midrule
    
    GeoQuestions1089 \citep{kefalidis2023benchmarking,kefalidis2024question} & 1{,}089 questions & 9 question categories, 3 spatial relation types & English & executed SPARQL and GeoSPARQL queries & YAGO2, YAGO2geo & USA, Greece, UK, Ireland & text \\   
    \addlinespace[3pt]
    
    MapEval \citep{dihan2025mapeval} & 700 questions & 6 question types over 3 tasks & English & derived from Google Maps & Google Maps & 180 cities in 54 countries & text, map images \\ 
    \addlinespace[3pt]

    STBench \citep{li2025stbench} & 70,000+ \qa pairs & 15 tasks over 4 dimensions & English & computed from the source data & 4 datasets & -- & text \\
    \addlinespace[3pt]

    MapQA \citep{li2025mapqa} & 3{,}154 \qa pairs & 9 templates & English & executed SQL queries & \osm & Southern California and Illinois (USA) & text, entity geometries \\
    \addlinespace[3pt]

    GS-QA \citep{saeedan2026gs} & 2{,}800 \qa pairs & 28 templates, 5 spatial predicates & English & executed SQL queries & \osm, Wikipedia & USA & not stated \\
    \addlinespace[3pt]

    GPSBench \citep{truong2026gpsbench} & 57{,}800 samples & 17 tasks in 2 tracks & English & geodetic formulae, GeoNames look-ups & GeoNames \cite{geonames} & six continents, sampled by continent quota & text \\
    \addlinespace[3pt]

    GeoBenchX \citep{krechetova2025geobenchx} & 200+ tasks & 4 task groups & English & \llm-drafted and manually rewritten reference solutions & 18 statistical, 21 vector, 11 raster datasets & -- & -- \\
    \addlinespace[3pt]

    UrbanGeoEval \citep{bao2026urbangeoeval} & $\sim$54{,}000 knowledge and 3{,}148 reasoning questions & 20 templates over 6 knowledge dimensions, 7 reasoning tasks & English & \osm-derived knowledge base; human annotation & \osm, public web sources & 18 cities on 6 continents & text \\
    \addlinespace[3pt]

    GeoAnalystBench \citep{zhang2025geoanalystbench} & 50 tasks & 6 spatial-analysis categories & -- & human-designed reference workflows and code & task-specific GIS datasets & -- & -- \\
    \addlinespace[3pt]

    \citet{suizu2026automatic} & 5{,}309 questions & 14 templates & Japanese & deterministic filtering of the source data & \osm, Wikidata & Japan & text \\    
    \midrule
    
    \textbf{\multiglobe} (ours) & 46{,}060 \qa pairs per language & 65 templates, 14 spatial-function families & 17 & executed SQL queries & \worldkg, \kwg, \osmkg & 201 countries and territories & text, entity images  \\
    \bottomrule
    \end{tabularx}

\caption{\textbf{Comparison of \multiglobe against related geospatial \qa and agent benchmarks.} Dashes (--) mark information not reported by the corresponding paper.}
\label{tab:appendix_related_benchmarks}
\end{table*}

\section{Multilingual Extension}
\label{sec:appendix_multilingual_extension}

\subsection{Human Annotation Protocol}
\label{sec:appendix_human_annotation_protocol}

\rparagraph{Recruitment and Compensation}
For each of the 16 target languages, we recruited two annotators through the authors' academic networks, predominantly Master's and PhD students in data science, computer science, or related fields, plus a few professional translators. All had C2-level proficiency in both the target language and English. We compensate externally recruited annotators (those not affiliated with the authors' labs) with \texteuro{}14 per hour.

\rparagraph{Annotation Workflow}
We use the Potato annotation tool \citep{pei2022potato}. The annotation task comprises six phases.

\iparagraph{(1) Welcome and Consent}
The welcome page summarizes the project, task, and expected workload.
Annotators must agree to three consent items before proceeding: that they are at least 18, that their annotations will be used solely for anonymized academic research with no personally identifiable information published, and that they understand these terms. We collect no personally identifiable information apart from a self-chosen annotator ID. 

\iparagraph{(2) Annotation Guidelines}
Each annotator must read the full guidelines page before any item is shown; the \textsc{Guidelines} link remains accessible throughout.
We frame the task as translation quality assessment: judging whether a \mt of an English template faithfully preserves the original meaning, structure, and spatial intent. The labels are: 
\begin{itemize}
    \setlength{\itemsep}{1pt}
    \item \textbf{Correct:} the spatial relationship, interrogative structure, entity placeholders, and meaning are all preserved, and the phrasing reads naturally to a native speaker.
    \item \textbf{Correct -- Phrasing Improvement Needed:} the spatial meaning is preserved and the phrasing is grammatical but unnatural, register-mismatched, or non-idiomatic. 
    \item \textbf{Incorrect:} the translation alters, omits or uses the wrong spatial relation; merges, renames, reorders, or omits entity placeholders; adds or removes critical information; or changes the question type (e.g., turning a yes/no question into a \textit{what} or \textit{where} question).
\end{itemize}
\begin{table*}[t]
\centering\
\footnotesize

\begin{tabularx}{\textwidth}{@{} l X @{}}
    \toprule
    \textbf{Reason} & \textbf{Description} \\
    \midrule
    
     \rowcolor{gray!15}\multicolumn{2}{l}{\textit{Correct -- Phrasing Improvement Needed}} \\
     \addlinespace[2pt]
    Word choice / vocabulary &  The translation uses a word or phrase that is technically correct but not the most natural or commonly used in the target language (e.g., a less idiomatic synonym). \\
    Grammar / syntax         & The sentence structure or grammar is acceptable but awkward, unusual, or not preferred in standard usage. \\
    Register / formality     & The translation is too formal, too informal, or does not match the neutral register expected for a factual spatial question. \\
    Word order               & The word order, while grammatically correct, is unusual or less natural. Reordering improves readability. \\
    
    \rowcolor{gray!15}\multicolumn{2}{l}{\textit{Incorrect}} \\
    \addlinespace[2pt]
    Meaning changed or lost  & The translation alters, omits, or distorts the original meaning of the question. \\
    Information added or removed & The translation includes details not present in the original or leaves out important parts. \\
    Wrong spatial relationship & The spatial relation in the translation differs from the original (e.g., "between" → "near", "north of" → "above"). \\
    Entity placeholders altered & One or more \texttt{\{entity\_X\}} placeholders are missing, renamed, merged, split, or otherwise modified. \\
    \bottomrule
    \end{tabularx}
    
    \caption{Correction-reason taxonomy non-\textit{Correct} items. Annotators select all that apply.}
    \label{tab:appendix_correction_reason_taxonomy}
\end{table*}
\begin{figure*}[t]
       \centering
       \begin{subfigure}[t]{0.49\textwidth}
           \centering
           
            \includegraphics[width=\textwidth]{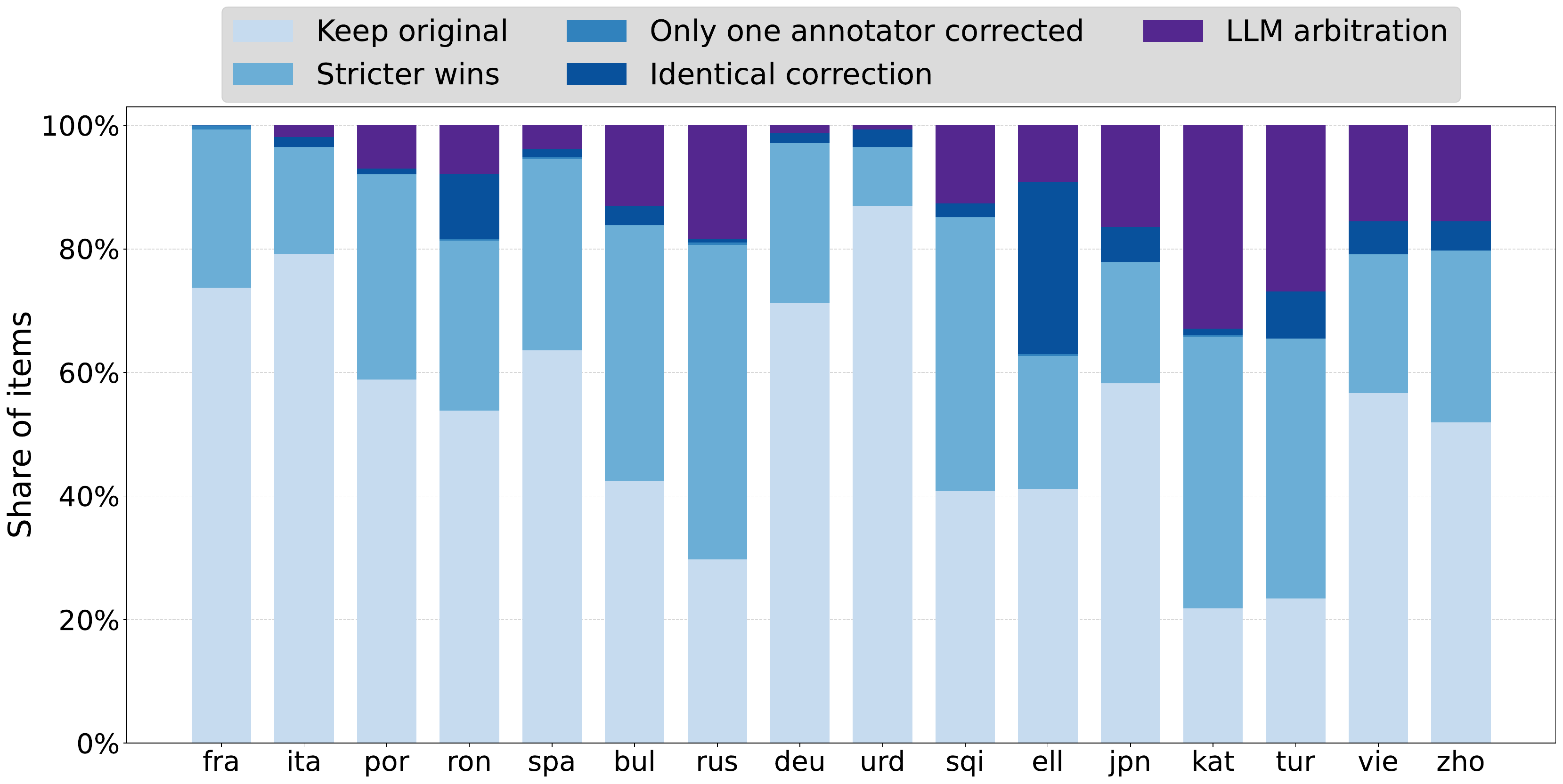}
            
            \caption{\textbf{Resolution source by language.} Share of items resolved by each deterministic rule (keep original \mt, stricter judgment, one-sided resolution, identical corrections) or escalated to \llm arbitration.}
            \label{fig:appendix_resolution_pipeline_outcomes}
       \end{subfigure}
       \hfill
       \begin{subfigure}[t]{0.49\textwidth}
           \centering
           \includegraphics[width=\textwidth]{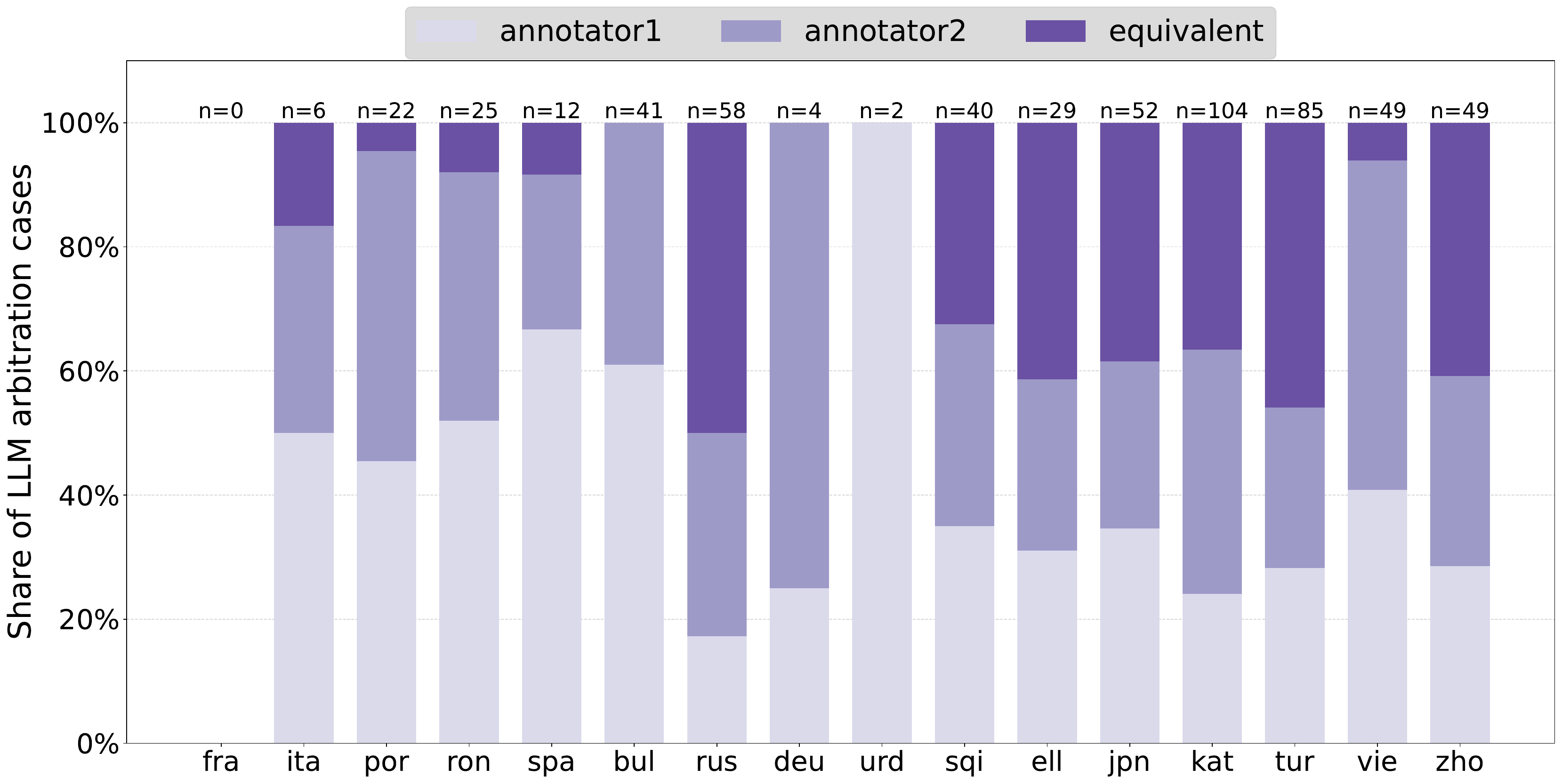}
           \caption{\textbf{\llm arbitration outcomes by language.} For items escalated to the blind \llm judge (Claude Opus 4.7), distribution of final selections across annotator 1, annotator 2, and equivalent.}
           \label{fig:appendix_resolution_llm_outcomes}
       \end{subfigure}
       \caption{\textbf{Resolution pipeline outcomes across languages.}}
       \label{fig:resolution_pipeline_results}
   \end{figure*}
For any non-\textit{Correct} label, the annotator must provide a corrected translation and select at least one reason from the fixed taxonomy in Table \ref{tab:appendix_correction_reason_taxonomy}. 
Two additional rules prevent over-correction. 
First, entity placeholders (e.g.\ \texttt{\{entity\_A\}}) must appear verbatim; annotators are told \emph{not} to translate, capitalize, or otherwise modify them, though they may reorder placeholders to match natural word order. 
Second, annotators must \emph{not} correct for entity-dependent grammatical variation, e.g., a Romanian masculine participle that would shift to feminine for a specific placeholder is marked \textit{Correct}, as such morphology is handled by the post-editing \llm ensemble (Appendix \ref{sec:appendix_postediting_ensemble}). 
The instructions also include language-specific rules (diacritics, non-Latin scripts, RTL handling, agglutinative-suffix attachment).

\iparagraph{(3) Annotation Items}
Each item appears on its own page showing the English template, the machine translation, an entity-constraint description, and one or two English example instantiations. 

\iparagraph{(4) Attention Checks}
We interleave attention checks with real items throughout each session to detect inattentive responding. Each check is an item whose text contains an explicit instruction with the expected label, the required correction reasons (when applicable), and the corrected string. 

\iparagraph{(5) Review \& Feedback}
Before submission, annotators may revise any completed item. They also answer three mandatory feedback questions on task difficulty, guideline clarity, and translation quality, plus an optional free-text field.

\iparagraph{(6) Completion}
The final page confirms completion and provides a completion code.

\subsection{Disagreement Resolution Pipeline}
\label{sec:disagreement_resolution_pipeline}

The first stage is a rule-based pass over deterministic cases: if both annotators label the \mt \textit{Correct}, we retain it; if their corrections agree, we take either; if the labels differ, we defer to the stricter annotator, since a stricter label is less likely to be a false-positive \textit{Correct}. We validated this rule against a third annotator on a sample of six languages, observing near-perfect agreement. All text comparisons use NFC Unicode normalization followed by whitespace collapse, so typographically-equivalent corrections resolve via rules.

\begin{table*}[t]
    \centering 
    \footnotesize
    \setlength{\tabcolsep}{3.5pt}\renewcommand{\arraystretch}{1.2}
    \begin{threeparttable}
    \begin{tabular}{@{} l r g r g r g r g r g r g r g r g @{}}
    \toprule
    \textbf{Metric} & \textbf{fra} & \textbf{ita} & \textbf{por} & \textbf{ron} & \textbf{spa} & \textbf{bul} & \textbf{rus} & \textbf{deu} & \textbf{urd} & \textbf{sqi} & \textbf{ell} & \textbf{jpn} & \textbf{kat} & \textbf{tur} & \textbf{vie} & \textbf{zho} \\
    \midrule
    \% Correct    & 86.9 & 87.9 & 74.4 & 66.5 & 79.0 & 60.4 & 50.9 & 83.4 & 91.6 & 58.8 & 50.7 & 66.6 & 38.6 & 43.4 & 66.7 & 64.8 \\
    \% Phrasing   & 3.5 & 12.1 & 17.5 & 28.4 & 19.2 & 34.9 & 40.9 & 13.7 & 5.0 & 35.6 & 45.2 & 8.7 & 45.6 & 27.6 & 28.5 & 12.9 \\
    Corr.\ rate   & 25.6 & 20.8 & 41.0 & 46.1 & 36.6 & 57.7 & 70.0 & 28.7 & 12.9 & 59.3 & 59.0 & 42.0 & 77.9 & 76.3 & 43.2 & 47.9 \\
    Agreement     & 74.3 & 82.7 & 66.9 & 72.2 & 68.8 & 58.7 & 49.2 & 74.1 & 90.5 & 55.2 & 78.5 & 80.1 & 56.1 & 58.0 & 77.6 & 72.2 \\
    $\alpha$      & -0.10 & 0.19 & 0.16 & 0.37 & 0.09 & 0.23 & 0.16 & 0.15 & 0.48 & 0.20 & 0.59 & 0.69 & 0.37 & 0.52 & 0.57 & 0.60 \\
    $\kappa$      & 0.01 & 0.19 & 0.19 & 0.38 & 0.09 & 0.28 & 0.23 & 0.20 & 0.48 & 0.21 & 0.59 & 0.69 & 0.39 & 0.52 & 0.57 & 0.61 \\
    chrF          & 92.9 & 95.8 & 91.7 & 94.2 & 95.0 & 88.2 & 83.4 & 94.4 & 97.0 & 89.7 & 93.6 & 92.8 & 85.5 & 85.8 & 92.1 & 90.5 \\
    \% \llm\ arb. & 0.0 & 1.9 & 6.9 & 7.9 & 3.8 & 13.2 & 18.3 & 1.3 & 0.6 & 12.6 & 9.2 & 16.4 & 32.8 & 26.8 & 15.5 & 15.5 \\
    \bottomrule
    \end{tabular}
    
    \begin{tablenotes}[flushleft]\scriptsize
      \item The negative $\alpha$ for French reflects an annotator-calibration mismatch (one annotator never used \textit{Incorrect}); the stricter-wins rule routes these items to the conservative correction with no \llm{} escalation, so the released data stays coherent.
    \end{tablenotes}
    \end{threeparttable}
    
    \caption{\textbf{Per-language annotation statistics} ($N=315$ items). \textit{\% Correct}: share of annotator-item decisions labeled \textit{Correct} (\% \textit{Incorrect} $=100-$\% Correct $-$\% Phrasing). \textit{Corr.\ rate (\%)}: share of items where at least one annotator corrected original \mt. \textbf{Agreement (\%)}: exact agreement on the three-level label scale; $\mathbf{\alpha}$: Krippendorff's ordinal $\alpha$; $\mathbf{\kappa}$: Cohen's linear-weighted $\kappa$ on the same scale. \textit{chrF} is computed between the original \mt and the final resolved translation. \textit{\% \llm\ arb.}: share sent to \llm{} arbitration.}
    \label{tab:appendix_annotation_stats}
\end{table*}

\begin{figure*}[t]
    \centering
    \includegraphics[width=\textwidth]{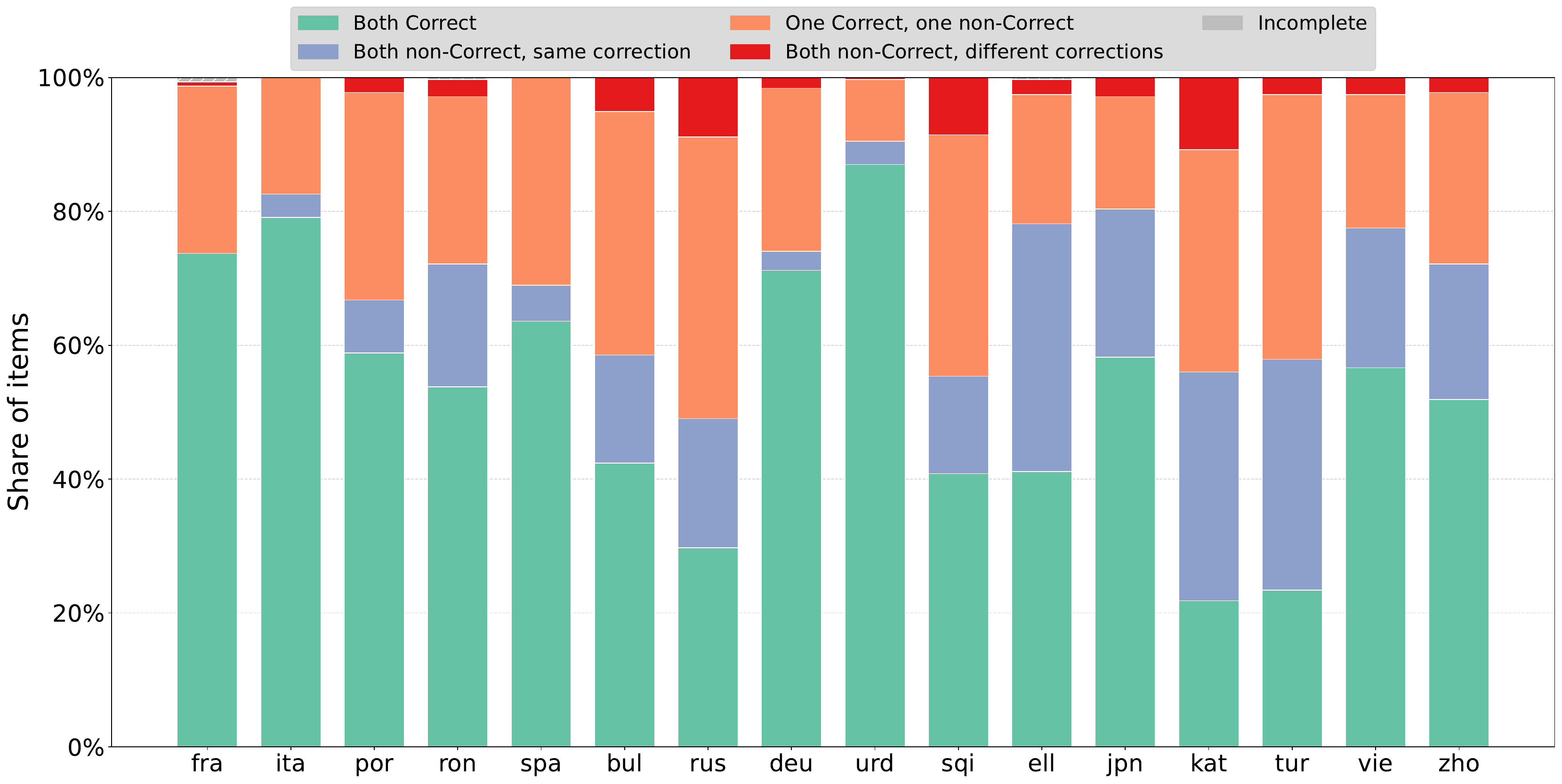}
    \caption{\textbf{Pairwise human-annotator agreement on translation post-editing, per target language}. Two annotators independently labeled every translated question as \emph{Correct} (no edit needed) or \emph{non-Correct}, supplying an edited string in the latter case. }
    \label{fig:annotation_label_agreement}
\end{figure*}
In the second stage, the \llm judge (Claude Opus 4.7) receives the English template, the original \mt, and the two anonymized candidate corrections in randomized order to mitigate positional bias.\footnote{We choose a frontier closed-source model to reduce training-data overlap with the models under evaluation.}
For each item it returns one of the three choices (\textit{annotator 1}, \textit{annotator 2}, or \textit{equivalent}), with a one-sentence rationale naming the decisive criterion. The output is strict JSON, and the pipeline rejects any response failing schema validation. The final correction is retrieved via the chosen annotator label, so the judge never produces translation content.
Fig. \ref{fig:resolution_pipeline_results} shows the distribution of items across deterministic rules per language, and the judge's selections for escalated questions.

\rparagraph{Decision Criteria}
The prompt instructs the judge to choose between candidate corrections by the following priority criteria, consulting the next only when the current one does not decide:

\begin{enumerate}
    \item \textit{Placeholder integrity:} every placeholder must appear verbatim, matching the count and identity in the English template. A correction that renames, merges, splits, or loses a placeholder is worse than one that keeps them intact.
    \item \textit{Semantic fidelity:} the spatial relation, quantifier, and question type (yes/no vs. what/where) must match the English; the correction preserving meaning more faithfully is preferred.
    \item \textit{Grammatical correctness:} agreement, case, verb form, and word order must be valid in the target language; the correction with fewer grammar errors is preferred.
    \item \textit{Technical terminology:} for domain terms (e.g., centroid, variance, fractal dimension), the correction using established vocabulary over approximations is preferred.
    \item \textit{Natural phrasing:} among otherwise-equal options, the more idiomatic phrasing is preferred.
\end{enumerate}
To prevent spurious choices, the prompt disallows four dimensions as decision drivers: (i) gender and case agreement hard-coded on a placeholder, (ii) whitespace and trivial-punctuation differences, (iii)  correction length, and (iv) register choice (formal/informal), provided it is internally consistent.

\rparagraph{Language-specific Instructions}
We expand the system prompt with short per-language instructions injected into the  \texttt{\{language\_specific\_notes\}} slot.
Each instruction comprises 4-5 points capturing three categories of target-language knowledge not covered by the universal criteria: (i) orthographic conventions and variations that should not influence the choice (e.g., the legacy-cedilla vs. comma-below distinction in Romanian), (ii) terminology preferences for domain-specific vocabulary, and (iii) typological anti-patterns, in particular reminders not to penalize corrections for failing to hard-code gender or case agreement on a placeholder.
All are available in the project's repository.

\rparagraph{System Prompt Template}
Fig. \ref{fig:appendix_llm_arbitrator_prompt} shows the system prompt template, with two expandable slots: \texttt{\{target\_language\}} (the human-readable language name), and \texttt{\{language\_specific\_notes\}} (the per-language instructions). 

\subsection{Per-language Annotation Statistics}
\label{sec:appendix_lang_annotation_stats}
Table \ref{tab:appendix_annotation_stats} summarizes the annotation effort and reliability signals per target language. 

\subsection{Pairwise Annotator Agreement}
\label{sec:appendix_pairwise_annotator_agreement}
Fig. \ref{fig:annotation_label_agreement} shows pairwise agreement on the translation post-editing task. Each item falls into one of five mutually exclusive categories combining the label decision (\textit{Correct} vs. non-\textit{Correct}) with the change type for items corrected by at least one annotator.

\begin{figure}[t]
    \centering
    \includegraphics[width=\columnwidth]{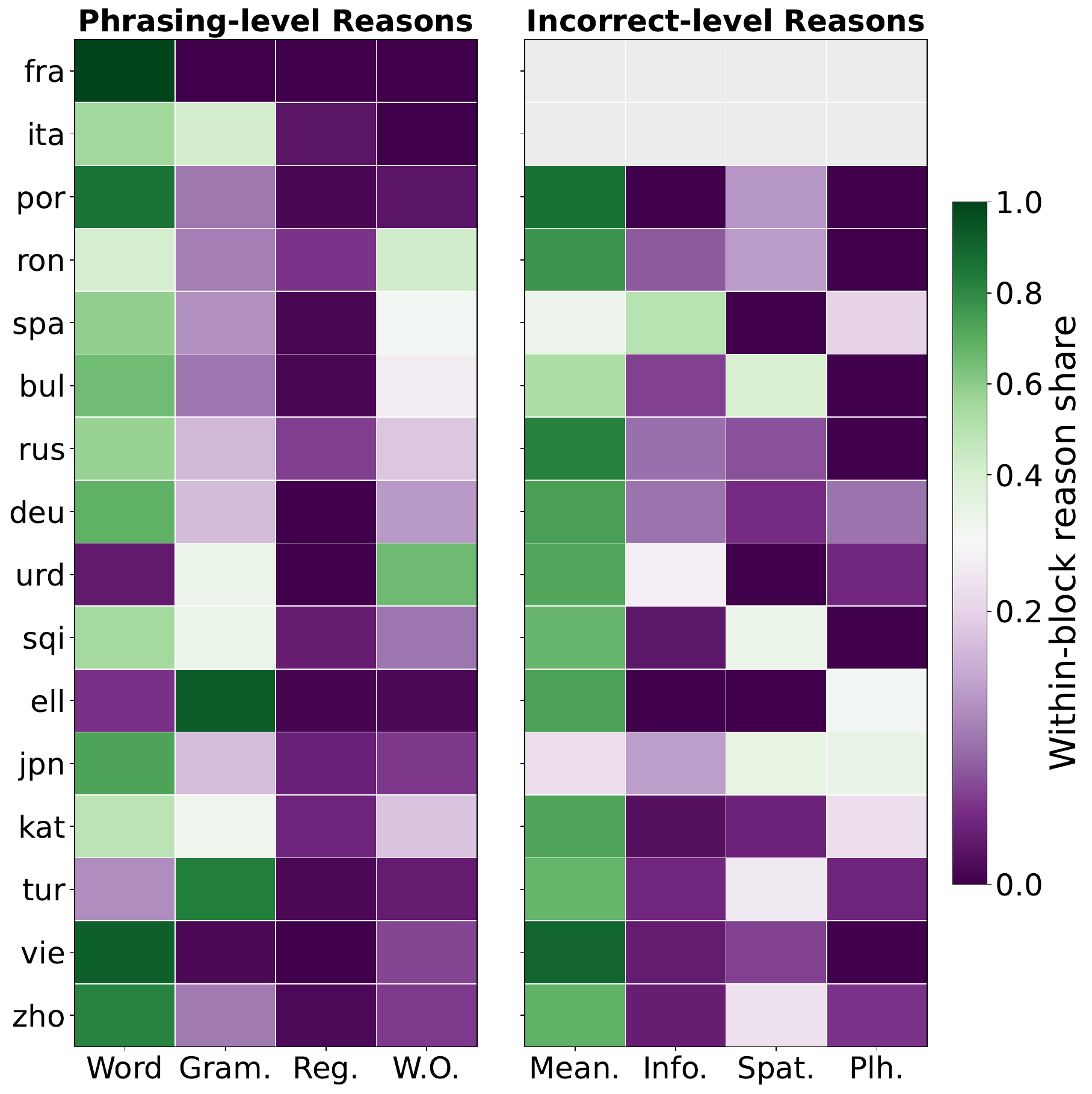}
    \caption{\textbf{Normalized multi-select correction-reason distributions per language.} Phrasing-level reasons (left block): \textit{Word}=word choice/vocabulary; \textit{Gram.}=grammar/syntax; \textit{Reg.}=register/formality; \textit{W.O.}=word order/sentence structure. Incorrect-level reasons (right block): \textit{Mean.}=meaning changed or lost; \textit{Info.}=information added or removed; \textit{Spat.}=wrong spatial relationship; \textit{Plh.}=entity placeholders altered.}
    \label{fig:appendix_annotation_reason_heatmap}
\end{figure}

\subsection{Correction Reason Distribution}
\label{sec:appendix_corr_reason_distribution}

Fig. \ref{fig:appendix_annotation_reason_heatmap} reports the normalized correction-reason distribution per language and label. 
\section{Post-editing Ensemble}
\label{sec:appendix_postediting_ensemble}
\begin{table*}[!ht]

\centering
\footnotesize

\begin{tabularx}{\textwidth}{@{} l X X l @{}}
    \toprule
    
    \textbf{Operation} & \textbf{Description} & \textbf{Example (before $\rightarrow$ after)} & \textbf{Typological trigger} \\
    \midrule
    
    \texttt{NO\_CHANGE} & The instantiated question is already correct; no edit needed. & EN: What is the distance between London and Paris? (unchanged). & --- \\ \hdashline
    \addlinespace[2pt]
    
    \texttt{ART\_ADD}   & Add an article that is missing because the template did not anticipate the entity's type. & EN: ...distance between \textit{Thames} and \textit{Seine}... $\rightarrow$ ...distance between \textit{the Thames} and \textit{the Seine}... & Article system \\
    \addlinespace[2pt]
    
    \texttt{ART\_DEL}   & Delete an article that becomes redundant with the instantiated entity. & DEU: in \textit{the Tokyo} $\rightarrow$ \textit{in Tokyo} & Article system \\
    \addlinespace[2pt]
    
    \texttt{ART\_CHG}   & Change one article for another, typically to match the entity's gender or number. & FRA: \textit{le} Tour Eiffel $\rightarrow$ \textit{la} Tour Eiffel & Article system $+$ gender \\ \hdashline
    \addlinespace[2pt]

    \texttt{ADP\_ADD}  & Add an adposition (e.g., preposition, postposition, or case / topic particle) required when the entity is substituted in. & ENG: the capital \textit{France} $\rightarrow$ the capital \textit{of France} & Adposition system \\
    \addlinespace[2pt]
    
    \texttt{ADP\_DEL}  & Delete an adposition (e.g., preposition, postposition, or case/topic particle) that is now redundant. & ENG: within \textit{of 5 km} of Paris $\rightarrow$ within \textit{5 km} of Paris & Adposition system \\
    \addlinespace[2pt]
    
    \texttt{ADP\_CHG}  & Change an adposition to another, including case / topic particles. & FRA: \textit{\`a Italie} $\rightarrow$ \textit{en Italie} & Adposition system \\ \hdashline
    \addlinespace[2pt]
    
    \texttt{DEC\_CHG}   & Change the declension / case ending of a word adjacent to the entity. & RON: capitala \textit{țara} Franța $\rightarrow$ capitala \textit{țării} Franța & Case system \\
    \addlinespace[2pt]
    
    \texttt{AGR\_CHG}   & Change gender, number, or person agreement on a modifier or verb. & ITA: Roma è \textit{stato fondato} $\rightarrow$ Roma è \textit{stata fondata} & Gender / number agreement \\
    \addlinespace[2pt]
    
    \texttt{CONTR\_CHG} & Add, remove, or change a contraction triggered by the entity. & FRA: \textit{de le Louvre} $\rightarrow$ \textit{du Louvre} & Article--preposition fusion \\ \hdashline
    \addlinespace[2pt]

    \texttt{ELISION}    & Elide a vowel before a vowel-initial entity. & FRA: \textit{de Italie} $\rightarrow$ \textit{d'Italie} & Vowel-hiatus phonotactics \\ \hdashline
    \addlinespace[2pt]

    \texttt{PUNC\_CHG}  & Fix punctuation or spacing around the entity. & FRA: la distance entre Paris et Rome\textit{?} $\rightarrow$ la distance entre Paris et Rome\textit{ ?} & Orthographic convention \\ \hdashline
    \addlinespace[2pt]
        
    \texttt{OTHER}      & Any other minor fix (e.g., diacritic correction, compound-split, stray casing) that does not fit the categories above. & Language specific & --- \\
    \bottomrule
\end{tabularx}

\caption{\textbf{The 13-category taxonomy of instantiation-artifact edits for the post-editing ensemble.}  \emph{Operation}: the acronym the models output; \emph{Description}: the explanation given to the model; \emph{Example}: a minimal before/after pair in a language where the edit is typical; \emph{Typological trigger}: the morpho-syntactic feature prompting the edit.}
\label{tab:appendix_postedit_taxonomy}

\end{table*}

\rparagraph{Edit Taxonomy}
Table \ref{tab:appendix_postedit_taxonomy} lists the full 13-category taxonomy used by the post-editing \llm ensemble.

\rparagraph{System Prompt}
Fig. \ref{fig:appendix_posedit_system_prompt} shows the system prompt for the ensemble \llms. The change operations (\verb|{_TAXONOMY_BLOCK}|) are generated from Table \ref{tab:appendix_postedit_taxonomy}. Placeholders such as \verb|{{entity_A}}| mark proper nouns to be preserved verbatim.

\rparagraph{User Prompt Template}
Fig. \ref{fig:appendix_posedit_user_prompt} shows the per-question user prompt, specifying the target-language ISO 639-3 code, the original English template, and the instantiated question.

\rparagraph{Expected Output Scheme}
Each model returns strict JSON formatted to the required number of edits (Fig. \ref{fig:appendix_posedit_output_schema}). Per-language prompts and model-specific parameters are in the code release.  

\section{Model Details}
\label{sec:appendix_model_details}
Table \ref{tab:appendix_models_used} summarizes the configurations of all models used.
\begin{table*}[t]
\centering
\footnotesize
\resizebox{\textwidth}{!}{%
    \begin{tabular}{@{}l l l c c c >{\raggedright\arraybackslash}p{4.2cm}@{}}
    \toprule
    
        \textbf{Model Name} & \textbf{Huggingface / API Identifier} & \textbf{\# Parameters} & \textbf{Provider} & \textbf{Cutoff} & \textbf{Served context} & \textbf{Serving Notes} \\
        \midrule
                
        \qwen & Qwen/Qwen3.5-27B-FP8 & 27B & Alibaba Cloud & n/d & 32K; YaRN 64K at T3a/T3c/T3-o & vLLM; FP8 \\
        \qwenmoe & Qwen/Qwen3.5-35B-A3B-FP8 & 35B (3B active) & Alibaba Cloud & n/d & 32K & vLLM; FP8; sparse MoE \\
        \qwennew & Qwen/Qwen3.8-27B-FP8 & 27B & Alibaba Cloud & n/d & 256K & vLLM; FP8; thinking (T2/T3-o+thinking) \\
        \addlinespace[2pt]

        \gemma & google/gemma-3-27b-it & 27B & Google & Aug 2024 & 32K & vLLM; \texttt{bfloat16} \\
        \gemmanew & google/gemma-4-31B-it & 31B & Google & Jan 2025 & 256K & vLLM; \texttt{bfloat16}; thinking (T2/T3-o+thinking) \\
        \addlinespace[2pt]

        \gptossl & openai/gpt-oss-120b & 117B (5.1B active) & OpenAI & June 2024 & 128K & vLLM; MXFP4; sparse MoE; reasoning effort \texttt{low} (\texttt{high} at T2) \\
        \gptosss & openai/gpt-oss-20b & 21B (3.6B active) & OpenAI & June 2024 & 128K & vLLM; MXFP4; sparse MoE; reasoning effort \texttt{low} (\texttt{high} at T2) \\
        \addlinespace[2pt]

        \glm & zai-org/GLM-4.7-Flash & 30B (3B active) & Z.ai & April 2025 & 198K & vLLM; \texttt{bfloat16}; sparse MoE; thinking (T2/T3-o+thinking) \\
        \addlinespace[2pt]
        
        \gemini & gemini-3-flash-preview & n/d & Google & Jan 2025 & 1M & API; thinking \texttt{minimal}; \texttt{high} (T2) \\
        \midrule
        
        \fable & claude-fable-5-1 & n/d & Anthropic & June 2026 & 1M & Claude Code 2.1.274; unrestricted libraries; web tools disabled \\
        \midrule
        
        \deepseek & deepseek-v4-flash & 284B (13B active) & DeepSeek AI & n/d & 1M & API; sparse MoE \\
        \claude & claude-opus-4-7 & n/d & Anthropic & Jan 2026 & 1M & -- \\
        
    \bottomrule
    \end{tabular}%
    }
    \caption{\textbf{Models used across evaluation, post-editing, and annotation.} The first block lists the nine evaluated \llms, the second the frontier model (\fable) of the 109-question probe, and the third the models used only for post-editing (\deepseek) and for arbitrating annotator disagreements (\claude). ``n/d'' = not disclosed (closed model) or not on the model card (Qwen3.5 released Feb 2026; DeepSeek~v4 released Apr 2026; Qwen3.8 released Aug 2026). \textit{Served context} is the window at which we serve the model; Qwen3.5 and Gemma-3 models support larger windows natively. \textit{Cutoff} is the training-data/knowledge cutoff per provider documentation.}

\label{tab:appendix_models_used}
\end{table*}
\section{Inference Setup}
\label{sec:appendix_inference_setup}

We serve open-weight models with vLLM \cite{kwon2023efficient} on one 141 GB H200, and query \gemini through its official API.
The frontier model, \fable, runs in its own separate agent harness (Claude Code). 
All conditions use greedy decoding (T = 0, top-p = 1).
Table \ref{tab:appendix_models_used} lists the reasoning settings and the served context window of each model.
At T2, the output budget is 2,048 tokens for prompted chain-of-thought and 32,768 for native reasoning (16,384 for \gemmanew, 8,192 for \gemini). 
For T3 runs, we cap each question at 10 agent steps with a 4,096-token output budget per step.\footnote{Steps are code generation/execution rounds, not individual tool calls. Across T3 variants (\textit{small} split, English, averaged over models), correct answers average 3.9 executed steps vs. 6.1 for incorrect ones, suggesting extra steps often signal non-convergence.} The T3-o+thinking ablation raises this budget (Table \ref{tab:appendix_thinking}).
\qwen, \qwenmoe and \gemma are served at our default window of 32K tokens, although the Qwen3.5 models support 256K natively, and \qwen's T3a, T3c and T3-o runs extend it to 64K via YaRN \cite{peng2024yarn} to accommodate cumulative tool outputs.
At T1, T2 and T1-o, we constrain model outputs to the per-question gold answer-type JSON schema, separating format compliance from spatial reasoning, and at the agentic tiers the answer passed to \texttt{final\_answer} is validated against the same answer type.

\section{Evaluation Metrics}
\label{sec:appendix_eval_metrics}
\begin{table*}[t]
\centering
\small
\begin{threeparttable}
    \begin{tabular}{@{}l l l@{}}
    \toprule
    \textbf{Answer type} & \textbf{Distance $d_i$} & \textbf{Tolerance $\tau_i$} \\
    \midrule
    \rowcolor{gray!15}\multicolumn{3}{@{}l}{\emph{Discrete}}\\
    Boolean / ternary (T/F/plausibly)  & $\mathds{1}[\hat a \neq a^{*}]$                     & exact ($\tau{=}1$) \\
    Entity name (single)               & $\mathds{1}[\mathrm{Resolve}(\hat a)\notin G^{*}]$\tnote{a} & exact ($\tau{=}1$) \\
    Predefined / cardinal direction    & $\mathds{1}[\hat a \neq a^{*}]$                     & exact ($\tau{=}1$) \\
    Set enumeration                    & $1-\mathrm{Jaccard}(\hat A, A^{*})$                 & exact set\tnote{b} \\
    \addlinespace[2pt]

    \rowcolor{gray!15}\multicolumn{3}{@{}l}{\emph{Grid-cell}}\\
    H3 / S2 cell index                 & $\mathds{1}[\hat a \neq a^{*}]$                     & exact\tnote{c} \\
    Geohash string                     & $\mathds{1}[\hat a \neq a^{*}]$                     & exact \\
    Composite (H3$+$S2)                & $\mathds{1}[\hat a \neq a^{*}]$ (component-wise)    & exact\tnote{c} \\
    \addlinespace[2pt]

    \rowcolor{gray!15}\multicolumn{3}{@{}l}{\emph{Numeric}}\\
    Distance                           & $|\hat y - y^{*}|$   & $\max(\delta,\alpha|y^{*}|)$,\; $\delta{=}0.5\,/\,5\,/\,10$~km \\
    Area                               & $|\hat y - y^{*}|$   & $\max(\delta,\alpha|y^{*}|)$,\; $\delta{=}0.25\,/\,25\,/\,100$~km$^2$ \\
    Count                              & $|\hat y - y^{*}|$   & $\max(1,\alpha|y^{*}|)$ \\
    Ratio (dimensionless)              & $|\hat y - y^{*}|$   & $\alpha|y^{*}|$ (relative, $\delta{=}0$) \\
    \addlinespace[2pt]

    \rowcolor{gray!15}\multicolumn{3}{@{}l}{\emph{Geometric / spatial}}\\
    Point coordinates                  & geodesic $d_{\mathrm{geo}}$   & $\max(\delta,\alpha\,d_{\text{scale}})$,\; $\delta{=}5$, $d_{\text{scale}}{=}200$~km \\
    Polar (angle, distance)            & angular $+$ radial error      & $5^{\circ}$ angular \textbf{and} radial within numeric $\tau$ \\
    Polygon / bounding box             & $1-\mathrm{IoU}$             & $\mathrm{IoU}\ge0.5$\tnote{b}\; (EPSG:6933) \\
    \addlinespace[2pt]

    \rowcolor{gray!15}\multicolumn{3}{@{}l}{\emph{Temporal}}\\
    Date                               & $|\hat d - d^{*}|$ days       & $1\,/\,15\,/\,365$~days (day/month/year) \\
    
    \bottomrule
    \end{tabular}
    
    \begin{tablenotes}[flushleft]\footnotesize
      \item[a] $G^{*}$ is the gold-acceptable answer set (including tied entities).
      \item[b] Exceptions to $E_{t_i}=\mathds{1}[d_i<\tau_i]$: $E_{t_i}=1$ requires exact set equality ($e_{t_i}=1-\mathrm{Jaccard}$) / $\mathrm{IoU}\ge0.5$ ($e_{t_i}=1-\mathrm{IoU}$).
      \item[c] $\mathrm{EM}$ requires equality with the gold cell index, and agreement at a coarser resolution is not credited. 

    \end{tablenotes}
    
\end{threeparttable}

    \caption{\textbf{Per-answer-type distance $d_i$ and tolerance $\tau_i$ used to compute $\mathrm{EM}$ and $\mathrm{NE}$.} $E_{t_i}=\mathds{1}[d_i<\tau_i]$ and $e_{t_i}=\min(1,d_i/\tau_i)$ (so $\mathrm{EM}=\tfrac1N\sum_i E_{t_i}$ and $\mathrm{NE}=\tfrac1N\sum_i e_{t_i}$), except where marked. Exact-match types use $d_i=\mathds{1}[\hat a\neq a^{*}]$ with $\tau_i{=}1$. For numeric answers $\tau_i=\max(\delta,\alpha|y^{*}|)$, where $\alpha=0.05$ and $\delta$ is a magnitude-dependent floor: distances use $\delta=0.5\,/\,5\,/\,10$~km for urban ($<50$~km), regional ($50$--$500$~km) and national ($\ge500$~km) scales; areas use $\delta=0.25\,/\,25\,/\,100$~km$^2$ over the analogous bands ($<2500$, $2500$--$250000$, $\ge250000$~km$^2$); counts $\delta=1$; ratios $\delta=0$. Coordinates use $\tau_i=\max(\delta,\alpha\,d_{\text{scale}})$ with $\delta=5$, $d_{\text{scale}}=200$~km; dates use precision-dependent day tolerances.}
    \label{tab:appendix_eval_metrics}
    
\end{table*}

Table \ref{tab:appendix_eval_metrics} gives the per-answer-type distance $d_i$ and tolerance $\tau_i$ used to score each of the 15 answer formats. 
Each true-premise question of a run has one of five outcomes, whose shares add up to 100\%: a parseable answer ($\mathrm{Cov}$), an explicit abstention ($\mathrm{FRR}$), budget exhaustion ($\mathrm{Exh}$), a harness error ($\mathrm{Err}$), or an answer that cannot be parsed.
An explicit abstention is the \texttt{null} answer that the prompts reserve for a question the model cannot answer (Appendix \ref{sec:appendix_model_prompts}).
Budget exhaustion also covers agent runs whose answer passed to \texttt{final\_answer} does not match the answer type, at most 2.7\% of the questions of a model and tier.

We compute every statistic per run and average it over the runs of a model and condition, with standard deviations over these runs. 
For differences between two tiers or two models, we report 95\% percentile intervals of a paired cluster bootstrap that resamples the 64 parent templates of the 4,979 true-premise questions with replacement (200,000 replicates), without adjustment for multiple comparisons. These intervals quantify the uncertainty from the choice of templates and questions and do not include the variation between runs, which the standard deviations show. The intervals of the frontier probe resample the 48 parent templates of its 109 questions (10,000 replicates; Appendix \ref{sec:appendix_frontier_probe}). 

\section{Additional Results}
\label{sec:appendix_additional_results}

\subsection{Representativeness}
\label{sec:appendix_representativeness}
\begin{table*}[t]
\centering
\small
\setlength{\tabcolsep}{1.5pt}
    \begin{tabular}{lcgcacgcacgcacgca}
    \toprule

     & \multicolumn{4}{c}{\textbf{EM}} & \multicolumn{4}{c}{\textbf{Cov.}} & \multicolumn{4}{c}{\textbf{EM\textsubscript{cov}}} & \multicolumn{4}{c}{\textbf{FRR}} \\
    \cmidrule(lr){2-5} \cmidrule(lr){6-9} \cmidrule(lr){10-13} \cmidrule(lr){14-17}

    \textbf{Model}
     & \emph{large} & \emph{small} & $\Delta$ & $\Delta^{\mathrm{rw}}$
     & \emph{large} & \emph{small} & $\Delta$ & $\Delta^{\mathrm{rw}}$
     & \emph{large} & \emph{small} & $\Delta$ & $\Delta^{\mathrm{rw}}$
     & \emph{large} & \emph{small} & $\Delta$ & $\Delta^{\mathrm{rw}}$ \\
    \midrule

    \qwen & 25.7 & 22.7 & $-$3.0 & $+$0.9 & 51.0 & 53.1 & $+$2.0 & $-$1.3 & 50.4 & 42.8 & $-$7.6 & $+$2.9 & 48.5 & 46.5 & $-$2.1 & $+$1.3 \\
    \qwenmoe & 21.5 & 17.3 & $-$4.2 & $-$0.3 & 35.0 & 30.1 & $-$4.9 & $-$1.9 & 61.5 & 57.4 & $-$4.2 & $+$2.7 & 65.0 & 69.9 & $+$4.9 & $+$1.9 \\
    \qwennew & 22.6 & 18.5 & $-$4.1 & $-$0.2 & 50.0 & 41.9 & $-$8.1 & $-$3.4 & 45.2 & 44.3 & $-$1.0 & $+$2.1 & 49.7 & 58.0 & $+$8.3 & $+$3.5 \\
    \gemma & 18.3 & 16.7 & $-$1.6 & $+$1.3 & 96.5 & 96.5 & $+$0.1 & $+$0.3 & 19.0 & 17.3 & $-$1.7 & $+$1.3 & 3.4 & 3.3 & $-$0.1 & $-$0.3 \\
    \gemmanew & 18.0 & 14.7 & $-$3.3 & $+$0.3 & 53.4 & 51.9 & $-$1.5 & $-$4.1 & 33.7 & 28.3 & $-$5.4 & $+$2.5 & 46.3 & 47.8 & $+$1.5 & $+$4.0 \\
    \gptossl & 13.8 & 12.4 & $-$1.4 & $+$0.6 & 21.9 & 24.1 & $+$2.2 & $+$0.3 & 63.1 & 51.6 & $-$11.4 & $+$1.4 & 78.1 & 75.9 & $-$2.2 & $-$0.3 \\
    \gptosss & 19.0 & 16.5 & $-$2.4 & $+$0.7 & 36.0 & 36.9 & $+$0.9 & $-$1.9 & 52.7 & 44.8 & $-$7.9 & $+$4.3 & 64.0 & 63.1 & $-$0.9 & $+$1.9 \\
    \glm & 17.6 & 15.1 & $-$2.5 & $+$0.1 & 52.7 & 50.5 & $-$2.2 & $-$1.4 & 33.4 & 29.9 & $-$3.5 & $+$0.2 & 47.1 & 49.3 & $+$2.2 & $+$1.5 \\
    {\gemini}$^{\dagger}$ & 23.2 & 22.6 & $-$0.6 & $+$2.2 & 88.9 & 93.7 & $+$4.8 & $+$5.8 & 26.1 & 24.1 & $-$1.9 & $+$0.9 & 5.9 & 6.3 & $+$0.4 & $-$0.6 \\
    \bottomrule
    \end{tabular}

    \caption{\textbf{Representativeness of the \emph{small} \multiglobe split} (English, text modality, T1; $\mathrm{EM}\uparrow$, $\mathrm{FRR}\downarrow$). Metrics are computed over \emph{true-premise} questions only, 41{,}540 of 45{,}114 for the \emph{large} split and 4{,}979 of 5{,}636 for the \emph{small} one. $\Delta$ is \emph{small}~$-$~\emph{large}, and $\Delta^{\mathrm{rw}}$ is the same difference after re-weighting each small-split per-family rate by the large split's distribution over the 14 spatial-function families. $^{\dagger}$ marks runs with one seed only on both splits.}
    \label{tab:appendix_representativeness}
\end{table*}

Table \ref{tab:appendix_representativeness} compares the \emph{small} and the \emph{large} split at T1. $\mathrm{EM}$ is lower on the \emph{small} split for all models, by 0.6 (\gemini) to 4.2 (\qwenmoe) points, and the two splits order them almost identically (Spearman $\rho=0.97$). The \emph{large}-split runs restricted to the questions of the \emph{small} split reproduce the \emph{small}-split $\mathrm{EM}$ of the eight open-weight models within 0.2 points, so the difference reflects the composition of the \emph{small} split. After re-weighting ($\Delta^{\mathrm{rw}}$), the $\mathrm{EM}$ differences lie between $-$0.3 and +2.2 points and are at most 1.3 points for the open-weight models. The majority baseline is 27.8\% (\emph{large}) against 28.8\% (\emph{small}).

\subsection{Further Metrics and Boolean Questions}
\label{sec:appendix_further_metrics}

\begin{table*}[t]
\centering
\small
\setlength{\tabcolsep}{3pt}
\newcommand{\sdv}[1]{\textsubscript{\tiny #1}}

    \begin{tabular}{@{}llcgcgcgcgc@{\hspace{10pt}}g@{}}
    \toprule
     &  & \textbf{Qwen3.5} & \textbf{Qwen3.5} & \textbf{Qwen3.8} & \textbf{Gemma-3} & \textbf{Gemma-4} & \textbf{gpt-oss} & \textbf{gpt-oss} & \textbf{GLM-4.7} & \textbf{Gemini-3} & \emph{Majority} \\
    \textbf{Tier} &  & \textbf{27B} & \textbf{35B} & \textbf{27B} & \textbf{27B} & \textbf{31B} & \textbf{120B} & \textbf{20B} & \textbf{Flash} & \textbf{Flash} & \emph{baseline} \\
    \midrule
    
    T1 & Boolean & 68.7\sdv{0.2} & 56.7\sdv{0.0} & 58.6\sdv{0.2} & 44.9\sdv{0.0} & 40.6\sdv{0.0} & 34.6\sdv{0.4} & 49.3\sdv{0.2} & 48.0\sdv{0.6} & 51.8\sdv{0.1} & 75.2\phantom{\sdv{0.0}} \\
     & Other & \phantom{0}5.1\sdv{0.1} & \phantom{0}2.2\sdv{0.0} & \phantom{0}3.2\sdv{0.1} & \phantom{0}5.9\sdv{0.1} & \phantom{0}4.8\sdv{0.0} & \phantom{0}4.0\sdv{0.1} & \phantom{0}4.0\sdv{0.0} & \phantom{0}2.5\sdv{0.1} & 11.6\sdv{0.1} & 11.0\phantom{\sdv{0.0}} \\
     & Share & 83.7\sdv{0.2} & 90.8\sdv{0.1} & 87.6\sdv{0.3} & 74.4\sdv{0.3} & 76.6\sdv{0.0} & 77.0\sdv{0.5} & 82.6\sdv{0.1} & 87.9\sdv{0.5} & 63.1\sdv{0.2} & 72.3\phantom{\sdv{0.0}} \\[0.6ex]
     
    T2 & Boolean & 65.1\sdv{0.1} & 36.6\sdv{0.6} & 40.3\sdv{0.6} & 44.3\sdv{0.1} & 35.6\sdv{0.3} & 49.8\sdv{0.2} & 13.9\sdv{0.5} & 40.3\sdv{1.2} & 55.8\sdv{0.4} & 75.2\phantom{\sdv{0.0}} \\
     & Other & \phantom{0}5.2\sdv{0.1} & \phantom{0}0.3\sdv{0.0} & \phantom{0}3.5\sdv{0.3} & \phantom{0}5.3\sdv{0.0} & \phantom{0}3.8\sdv{0.2} & \phantom{0}5.3\sdv{0.1} & \phantom{0}0.8\sdv{0.1} & \phantom{0}2.9\sdv{0.1} & 12.7\sdv{0.1} & 11.0\phantom{\sdv{0.0}} \\
     & Share & 82.8\sdv{0.4} & 98.1\sdv{0.2} & 81.4\sdv{1.1} & 76.3\sdv{0.0} & 78.1\sdv{0.6} & 78.3\sdv{0.4} & 86.6\sdv{1.6} & 84.2\sdv{0.3} & 62.7\sdv{0.2} & 72.3\phantom{\sdv{0.0}} \\
    \bottomrule
    \end{tabular}

    \caption{\textbf{Accuracy ($\mathrm{EM}$) on Boolean and other questions without external context} (\emph{small} split, English, text modality; tiers T1 and T2). \emph{Boolean} is $\mathrm{EM}$ over the 1{,}378 \emph{true-premise} questions with a Boolean answer, \emph{Other} is $\mathrm{EM}$ over the remaining 3{,}601 \emph{true-premise} questions, and \emph{Share} is the percentage of a model's correct answers that are answers to Boolean questions. Answering every Boolean question with ``false'', the gold answer of 841 of them, scores 61.0. Subscripts give standard deviation over three seeds.}
    \label{tab:appendix_boolean_split}
\end{table*}

Table \ref{tab:appendix_boolean_split} separates Boolean from other questions at T1 and T2, and Table \ref{tab:appendix_further_metrics} complements Table \ref{tab:main_results} and Fig. \ref{fig:outcome_shares} with $\mathrm{NE}$, $\mathrm{Cov}$ and $\mathrm{FRR}$.

\subsection{Tier and Model Differences}
\label{sec:appendix_significance}

Tables \ref{tab:appendix_tier_gaps} and \ref{tab:appendix_model_differences} report differences in $\mathrm{EM}$ between tiers and models, with 95\% bootstrap intervals (Appendix \ref{sec:appendix_eval_metrics}). Between tiers, the interval excludes zero for all models where \kg retrieval is compared with parametric answering or with gold triples (T3a $-$ T1, T3-o $-$ T3a), and for a single model where web search is added to the \kg (T3c $-$ T3a).

\subsection{Abstention}
\label{sec:appendix_abstention}

\begin{figure*}[t]
    \centering
    \includegraphics[width=\textwidth]{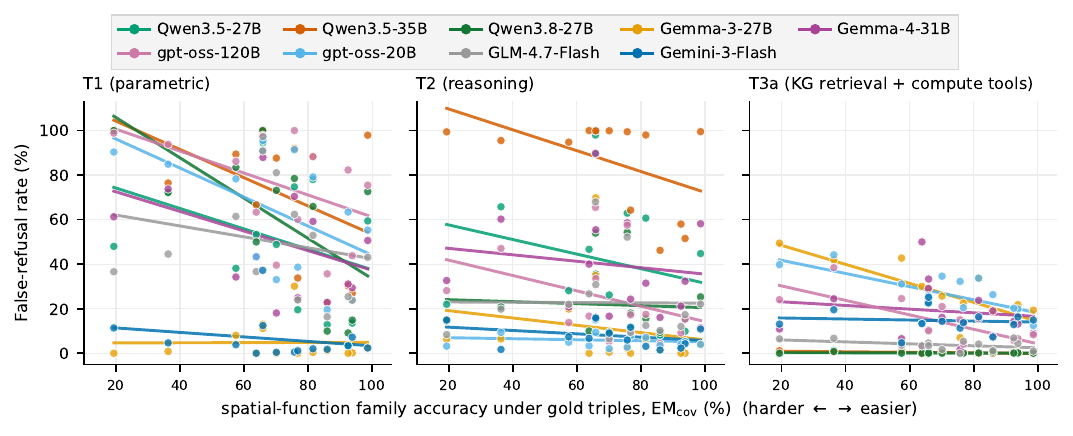}
    \caption{\textbf{False-refusal rate against spatial-function family difficulty} (\emph{small} split, English, text modality, 4,979 \emph{true-premise} questions). Each point is one model on one of the 14 spatial-function families. Difficulty is measured as the accuracy that family permits under gold triples ($\mathrm{EM_{cov}}$ at T3-o). Lines are least-squares fits shown for orientation. The reported coefficients are Spearman correlations, computed per seed and averaged over seeds.}
    \label{fig:appendix_abstention_difficulty}
\end{figure*}

Fig. \ref{fig:appendix_abstention_difficulty} plots each model's $\mathrm{FRR}$ per spatial-function family against the family's $\mathrm{EM_{cov}}$ at T3-o averaged over the models, which ranges from 19.4\% for grid indexing to 98.6\% for coordinate questions. Because this ordering is computed once, from a different condition, and excludes unanswered items, it shares
no denominator with the refusal rates plotted against it.
Without retrieval, the Spearman correlation $\rho$ between the $\mathrm{FRR}$ and the family accuracy under gold triples is negative for all models, from $-0.65$ to $-0.02$ at T1 and from $-0.49$ to $-0.12$ at T2. \gemini rarely refuses at either tier, and \qwenmoe refuses 83.6\% of true-premise questions at T2, between 46.2\% and 100\% per family.
Under \kg retrieval, $\rho$ is negative for the eight open-weight models ($-0.76$ to $-0.03$) and $+0.01$ for \gemini. \qwen, \qwenmoe and \qwennew abstain explicitly on at most 19 of the 4,979 questions per seed at T3a, so their coefficients rest on few abstentions. \gemma and \gptossl refuse most on grid indexing, shape and spatial transformation, the three families that remain hardest with gold facts, and \gptosss on shape and grid indexing. Ordering families by oracle $\mathrm{EM}$ instead, which counts unanswered questions as incorrect, ranks them almost identically ($\rho=0.93$).

\subsection{Oracle Context Representation}
\label{sec:oracle_context_ablation}

\begin{figure*}[!t]
    \centering
    \includegraphics[width=\textwidth]{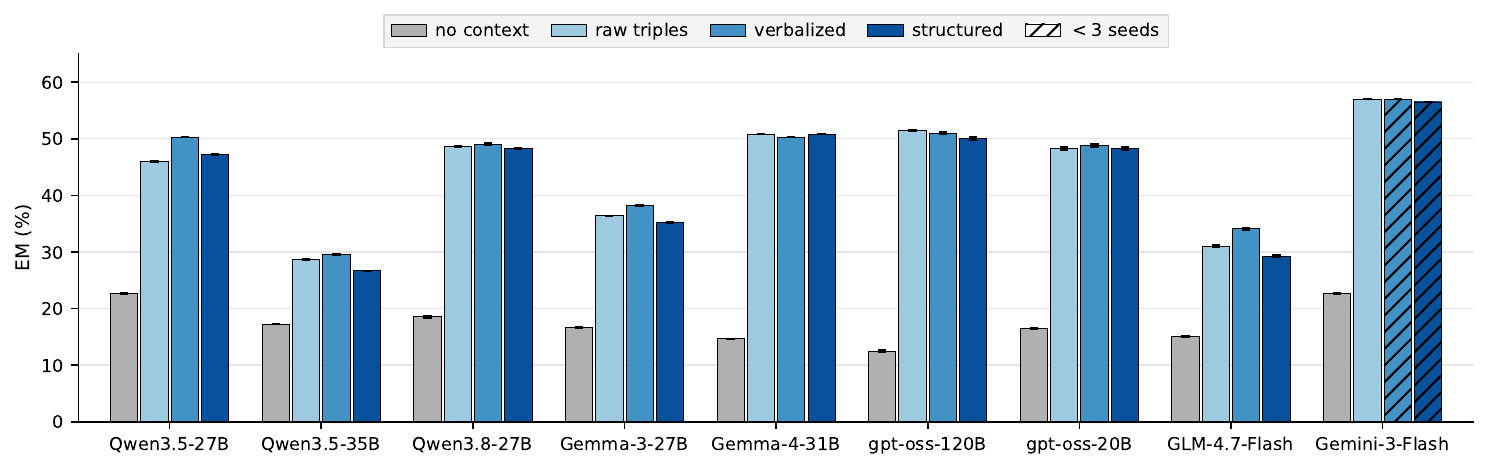}
    \caption{\textbf{Oracle ablation (T1)} (\emph{small} split, English, text modality; $\mathrm{EM}$ over 4,979 \emph{true-premise} questions).}
    \label{fig:appendix_oracle_ablation}
\end{figure*}

\begin{table*}[t]
\centering
\small
\setlength{\tabcolsep}{4.5pt}
\newcommand{\sdv}[1]{\textsubscript{\tiny #1}}

    \begin{tabular}{llcgcgcgcgc}
    \toprule
     &  & \textbf{Qwen3.5} & \textbf{Qwen3.5} & \textbf{Qwen3.8} & \textbf{Gemma-3} & \textbf{Gemma-4} & \textbf{gpt-oss} & \textbf{gpt-oss} & \textbf{GLM-4.7} & \textbf{Gemini-3} \\
    \textbf{Tier} &  & \textbf{27B} & \textbf{35B} & \textbf{27B} & \textbf{27B} & \textbf{31B} & \textbf{120B} & \textbf{20B} & \textbf{Flash} & \textbf{Flash} \\
    \midrule
    \multicolumn{11}{@{}l}{\emph{no external context}} \\
    T1 & NE & 77.6\sdv{0.1} & 82.8\sdv{0.0} & 81.6\sdv{0.1} & 83.9\sdv{0.0} & 85.7\sdv{0.0} & 88.0\sdv{0.1} & 83.7\sdv{0.1} & 85.0\sdv{0.1} & 78.5\sdv{0.1}\phantom{\sdv{0}}\phantom{$^{\dagger}$} \\
     & Cov & 53.1\sdv{0.1} & 30.1\sdv{0.1} & 41.9\sdv{0.1} & 96.5\sdv{0.0} & 51.9\sdv{0.1} & 24.1\sdv{0.1} & 36.9\sdv{0.1} & 50.5\sdv{0.2} & 93.5\sdv{0.2}\phantom{\sdv{0}}\phantom{$^{\dagger}$} \\
     & FRR & 46.5\sdv{0.2} & 69.9\sdv{0.1} & 58.0\sdv{0.1} & \phantom{0}3.3\sdv{0.0} & 47.8\sdv{0.1} & 75.9\sdv{0.1} & 63.1\sdv{0.1} & 49.3\sdv{0.2} & \phantom{0}6.5\sdv{0.2}\phantom{\sdv{0}}\phantom{$^{\dagger}$} \\[0.6ex]
    T2 & NE & 78.7\sdv{0.1} & 89.7\sdv{0.1} & 86.6\sdv{0.3} & 84.4\sdv{0.0} & 87.6\sdv{0.1} & 83.1\sdv{0.1} & 95.6\sdv{0.2} & 87.1\sdv{0.3} & 76.8\sdv{0.1}\phantom{\sdv{0}}\phantom{$^{\dagger}$} \\
     & Cov & 58.8\sdv{0.1} & 15.4\sdv{0.2} & 54.3\sdv{0.4} & 83.7\sdv{0.1} & 34.4\sdv{0.1} & 59.1\sdv{0.4} & \phantom{0}7.0\sdv{0.2} & 40.3\sdv{0.3} & 91.5\sdv{0.5}\phantom{\sdv{0}}\phantom{$^{\dagger}$} \\
     & FRR & 36.8\sdv{0.2} & 83.6\sdv{0.3} & 20.0\sdv{0.6} & 12.5\sdv{0.1} & 37.7\sdv{0.0} & 21.8\sdv{0.3} & \phantom{0}6.2\sdv{0.3} & 19.6\sdv{0.2} & \phantom{0}8.5\sdv{0.6}\phantom{\sdv{0}}\phantom{$^{\dagger}$} \\
    \midrule
    
    \multicolumn{11}{@{}l}{\emph{agentic retrieval from \kg (T3a), web (T3b), or both (T3c)}} \\
    T3a & NE & 57.0\sdv{0.3} & 63.7\sdv{0.1} & 52.6\sdv{0.0} & 70.2\sdv{0.0} & 54.6\sdv{0.0} & 60.5\sdv{0.2} & 71.3\sdv{0.4} & 69.1\sdv{0.4} & 71.7\sdv{3.3}\phantom{\sdv{0}}\phantom{$^{\dagger}$} \\
     & Cov & 61.2\sdv{0.4} & 49.4\sdv{0.1} & 68.0\sdv{0.1} & 59.2\sdv{0.0} & 81.2\sdv{0.0} & 77.4\sdv{0.1} & 56.9\sdv{0.5} & 51.0\sdv{0.5} & 73.3\sdv{6.2}\phantom{\sdv{0}}\phantom{$^{\dagger}$} \\
     & FRR & \phantom{0}0.2\sdv{0.0} & \phantom{0}0.3\sdv{0.0} & \phantom{0}0.1\sdv{0.0} & 27.5\sdv{0.1} & 16.8\sdv{0.0} & 14.0\sdv{0.1} & 26.3\sdv{0.2} & \phantom{0}4.3\sdv{0.1} & 15.2\sdv{10.9}\phantom{\sdv{}}\phantom{$^{\dagger}$} \\[0.6ex]
    T3b & NE & 77.9\sdv{0.7} & 78.0\sdv{0.4} & 77.4\sdv{0.8} & 79.5\sdv{0.4} & 69.5\sdv{0.6} & 74.0\sdv{0.4} & 86.2\sdv{0.1} & 84.0\sdv{0.5} & 73.8$^{\dagger}$\phantom{\sdv{0.0}}\phantom{\sdv{0}} \\
     & Cov & 35.0\sdv{0.7} & 36.6\sdv{0.8} & 35.0\sdv{1.3} & 57.4\sdv{0.6} & 65.5\sdv{1.0} & 59.2\sdv{0.1} & 34.1\sdv{0.2} & 29.4\sdv{0.3} & 74.1$^{\dagger}$\phantom{\sdv{0.0}}\phantom{\sdv{0}} \\
     & FRR & \phantom{0}0.0\sdv{0.0} & \phantom{0}0.2\sdv{0.1} & \phantom{0}0.2\sdv{0.1} & 24.8\sdv{0.1} & 32.5\sdv{1.1} & 22.8\sdv{0.5} & 43.0\sdv{1.0} & \phantom{0}1.6\sdv{0.1} & \phantom{0}1.1$^{\dagger}$\phantom{\sdv{0.0}}\phantom{\sdv{0}} \\[0.6ex]
    T3c & NE & 57.0\sdv{0.2} & 63.5\sdv{0.2} & 53.0\sdv{0.1} & 69.6\sdv{0.5} & 51.9\sdv{0.2} & 60.5\sdv{0.2} & 70.7\sdv{0.1} & 69.1\sdv{0.1} & 73.2$^{\dagger}$\sdv{0.0}\phantom{\sdv{0}} \\
     & Cov & 59.8\sdv{0.1} & 48.7\sdv{0.1} & 64.5\sdv{0.3} & 64.0\sdv{1.2} & 87.0\sdv{0.5} & 77.6\sdv{0.1} & 59.1\sdv{0.1} & 48.4\sdv{0.8} & 75.4$^{\dagger}$\sdv{0.0}\phantom{\sdv{0}} \\
     & FRR & \phantom{0}0.0\sdv{0.0} & \phantom{0}0.2\sdv{0.0} & \phantom{0}0.0\sdv{0.0} & 18.0\sdv{0.5} & 10.7\sdv{0.4} & 14.6\sdv{0.3} & 28.4\sdv{0.3} & \phantom{0}2.5\sdv{0.1} & \phantom{0}2.5$^{\dagger}$\sdv{0.0}\phantom{\sdv{0}} \\
    \midrule
    
    \multicolumn{11}{@{}l}{\emph{oracle conditions (gold triples injected)}} \\
    T1-o & NE & 50.3\sdv{0.0} & 70.1\sdv{0.1} & 51.2\sdv{0.0} & 62.1\sdv{0.0} & 49.8\sdv{0.0} & 49.4\sdv{0.1} & 51.6\sdv{0.2} & 65.6\sdv{0.1} & 43.0$^{\dagger}$\sdv{0.0}\phantom{\sdv{0}} \\
     & Cov & 77.1\sdv{0.1} & 42.5\sdv{0.1} & 76.9\sdv{0.1} & 85.8\sdv{0.0} & 76.1\sdv{0.0} & 63.9\sdv{0.1} & 63.9\sdv{0.1} & 69.5\sdv{0.0} & 89.2$^{\dagger}$\sdv{0.0}\phantom{\sdv{0}} \\
     & FRR & 18.1\sdv{0.2} & 53.6\sdv{0.1} & 21.4\sdv{0.1} & 10.3\sdv{0.0} & 21.6\sdv{0.0} & 34.0\sdv{0.1} & 33.1\sdv{0.1} & 28.1\sdv{0.1} & \phantom{0}9.9$^{\dagger}$\sdv{0.0}\phantom{\sdv{0}} \\[0.6ex]
    T3-o & NE & 39.2\sdv{0.2} & 45.7\sdv{0.1} & 35.1\sdv{0.0} & 55.6\sdv{0.1} & 37.1\sdv{0.2} & 39.4\sdv{0.1} & 43.3\sdv{0.1} & 86.7\sdv{0.3} & 40.0$^{\dagger}$\sdv{3.4}\phantom{\sdv{0}} \\
     & Cov & 80.0\sdv{0.3} & 69.6\sdv{0.1} & 84.7\sdv{0.1} & 61.9\sdv{0.1} & 80.7\sdv{0.3} & 80.4\sdv{0.2} & 73.1\sdv{0.2} & 24.5\sdv{0.3} & 73.9$^{\dagger}$\sdv{5.6}\phantom{\sdv{0}} \\
     & FRR & \phantom{0}5.5\sdv{0.1} & \phantom{0}6.0\sdv{0.1} & \phantom{0}6.6\sdv{0.1} & \phantom{0}9.6\sdv{0.4} & \phantom{0}9.0\sdv{0.1} & 11.2\sdv{0.2} & 14.6\sdv{0.0} & \phantom{0}4.6\sdv{0.1} & \phantom{0}3.9$^{\dagger}$\sdv{0.3}\phantom{\sdv{0}} \\
    \bottomrule
    \end{tabular}

    \caption{\textbf{Further metrics across evaluation tiers} (\emph{small} split, English, text modality; $\mathrm{NE}\downarrow$, $\mathrm{Cov}\uparrow$, $\mathrm{FRR}\downarrow$). Evaluation spans the 4{,}979 \emph{true-premise} questions. Subscripts give standard deviation over three seeds; $^{\dagger}$ marks cells with fewer seeds, and single-run cells show no standard deviation.}
    \label{tab:appendix_further_metrics}
\end{table*}

\begin{table*}[t]
\centering
\small
\setlength{\tabcolsep}{5pt}
    \begin{tabular}{@{}l cgcgcg@{}}
    \toprule
    \textbf{Model} & \textbf{T2\,$-$\,T1} & \textbf{T3a\,$-$\,T1} & \textbf{T3a\,$-$\,T3b} & \textbf{T3c\,$-$\,T3a} & \textbf{T3-o\,$-$\,T1-o} & \textbf{T3-o\,$-$\,T3a} \\
    \midrule
    \qwen & $-$1.0 & $+$21.6\rlap{\smash{$^{*}$}} & $+$21.7\rlap{\smash{$^{*}$}} & $-$0.2 & $+$11.3\rlap{\smash{$^{*}$}} & $+$17.3\rlap{\smash{$^{*}$}} \\[-0.4ex]
     & {\scriptsize [$-$3.4, $+$0.7]} & {\scriptsize [$+$11.0, $+$33.1]} & {\scriptsize [$+$12.7, $+$30.9]} & {\scriptsize [$-$1.4, $+$1.0]} & {\scriptsize [$+$6.6, $+$16.8]} & {\scriptsize [$+$10.5, $+$25.4]} \\[0.5ex]
    \qwenmoe & $-$6.9\rlap{\smash{$^{*}$}} & $+$19.8\rlap{\smash{$^{*}$}} & $+$14.3\rlap{\smash{$^{*}$}} & $+$0.1 & $+$25.5\rlap{\smash{$^{*}$}} & $+$18.0\rlap{\smash{$^{*}$}} \\[-0.4ex]
     & {\scriptsize [$-$11.4, $-$3.5]} & {\scriptsize [$+$8.6, $+$31.1]} & {\scriptsize [$+$4.0, $+$24.0]} & {\scriptsize [$-$1.5, $+$1.9]} & {\scriptsize [$+$13.2, $+$40.3]} & {\scriptsize [$+$13.3, $+$23.0]} \\[0.5ex]
    \qwennew & $-$4.8\rlap{\smash{$^{*}$}} & $+$30.4\rlap{\smash{$^{*}$}} & $+$25.8\rlap{\smash{$^{*}$}} & $-$0.5 & $+$16.2\rlap{\smash{$^{*}$}} & $+$16.3\rlap{\smash{$^{*}$}} \\[-0.4ex]
     & {\scriptsize [$-$10.4, $-$1.1]} & {\scriptsize [$+$18.0, $+$42.5]} & {\scriptsize [$+$16.9, $+$34.3]} & {\scriptsize [$-$1.4, $+$0.3]} & {\scriptsize [$+$8.9, $+$23.8]} & {\scriptsize [$+$8.8, $+$25.7]} \\[0.5ex]
    \gemma & $-$0.6\rlap{\smash{$^{*}$}} & $+$14.0\rlap{\smash{$^{*}$}} & $+$9.0\rlap{\smash{$^{*}$}} & $+$0.8 & $+$7.1 & $+$14.6\rlap{\smash{$^{*}$}} \\[-0.4ex]
     & {\scriptsize [$-$1.3, $-$0.1]} & {\scriptsize [$+$4.0, $+$24.1]} & {\scriptsize [$+$2.4, $+$15.6]} & {\scriptsize [$-$0.5, $+$2.6]} & {\scriptsize [$-$2.8, $+$17.2]} & {\scriptsize [$+$9.3, $+$20.7]} \\[0.5ex]
    \gemmanew & $-$2.0\rlap{\smash{$^{*}$}} & $+$32.3\rlap{\smash{$^{*}$}} & $+$15.2\rlap{\smash{$^{*}$}} & $+$2.8\rlap{\smash{$^{*}$}} & $+$13.0\rlap{\smash{$^{*}$}} & $+$16.3\rlap{\smash{$^{*}$}} \\[-0.4ex]
     & {\scriptsize [$-$4.5, $-$0.2]} & {\scriptsize [$+$21.1, $+$42.5]} & {\scriptsize [$+$8.5, $+$21.3]} & {\scriptsize [$+$1.1, $+$4.4]} & {\scriptsize [$+$5.3, $+$21.8]} & {\scriptsize [$+$8.3, $+$24.0]} \\[0.5ex]
    \gptossl & $+$5.1\rlap{\smash{$^{*}$}} & $+$28.3\rlap{\smash{$^{*}$}} & $+$13.8\rlap{\smash{$^{*}$}} & $+$0.1 & $+$10.4\rlap{\smash{$^{*}$}} & $+$20.7\rlap{\smash{$^{*}$}} \\[-0.4ex]
     & {\scriptsize [$+$0.5, $+$10.6]} & {\scriptsize [$+$16.0, $+$40.4]} & {\scriptsize [$+$7.1, $+$20.3]} & {\scriptsize [$-$0.7, $+$1.1]} & {\scriptsize [$+$6.2, $+$15.4]} & {\scriptsize [$+$14.5, $+$27.1]} \\[0.5ex]
    \gptosss & $-$12.1\rlap{\smash{$^{*}$}} & $+$12.9\rlap{\smash{$^{*}$}} & $+$15.3\rlap{\smash{$^{*}$}} & $+$0.7 & $+$8.5\rlap{\smash{$^{*}$}} & $+$27.9\rlap{\smash{$^{*}$}} \\[-0.4ex]
     & {\scriptsize [$-$19.4, $-$6.5]} & {\scriptsize [$+$4.3, $+$21.8]} & {\scriptsize [$+$8.3, $+$22.6]} & {\scriptsize [$-$0.2, $+$1.7]} & {\scriptsize [$+$4.1, $+$13.2]} & {\scriptsize [$+$20.2, $+$34.8]} \\[0.5ex]
    \glm & $-$1.9 & $+$16.4\rlap{\smash{$^{*}$}} & $+$15.0\rlap{\smash{$^{*}$}} & $-$0.1 & $-$20.6\rlap{\smash{$^{*}$}} & $-$18.0\rlap{\smash{$^{*}$}} \\[-0.4ex]
     & {\scriptsize [$-$6.0, $+$1.3]} & {\scriptsize [$+$7.9, $+$24.8]} & {\scriptsize [$+$8.6, $+$21.2]} & {\scriptsize [$-$0.8, $+$0.7]} & {\scriptsize [$-$30.5, $-$9.7]} & {\scriptsize [$-$27.4, $-$8.5]} \\[0.5ex]
    \gemini & $+$1.9 & $+$6.8\rlap{\smash{$^{*}$}} & $+$2.2\rlap{\smash{$^{\dagger}$}} & $-$1.5\rlap{\smash{$^{\dagger}$}} & $+$3.2\rlap{\smash{$^{\dagger}$}} & $+$30.8\rlap{\smash{$^{*\dagger}$}} \\[-0.4ex]
     & {\scriptsize [$-$0.2, $+$4.2]} & {\scriptsize [$+$0.6, $+$11.9]} & {\scriptsize [$-$1.0, $+$5.4]} & {\scriptsize [$-$3.4, $+$0.4]} & {\scriptsize [$-$5.1, $+$12.2]} & {\scriptsize [$+$21.5, $+$38.2]} \\
    \bottomrule
    \end{tabular}

    \caption{\textbf{Differences in EM between tiers} (\emph{small} split, English, text modality; 4{,}979 \emph{true-premise} questions). Each cell gives the $\mathrm{EM}$ of the first tier minus the $\mathrm{EM}$ of the second, each averaged over its seeds, and below it the 95\% bootstrap interval (Appendix \ref{sec:appendix_eval_metrics}). $^{*}$ marks intervals that exclude zero. $^{\dagger}$ marks cells with fewer seeds. \gemini has one seed at T3b and two seeds at T3c, T1-o and T3-o.}

    \label{tab:appendix_tier_gaps}
\end{table*}

\begin{table*}[t]
\centering
\small
\setlength{\tabcolsep}{5pt}
    \begin{tabular}{@{}l c r@{\hspace{1.7em}}l c c r@{\hspace{1.7em}}l@{}}
    \toprule
     & \multicolumn{3}{c}{\textbf{T3-o}} & & \multicolumn{3}{c}{\textbf{T3a}} \\
    \cmidrule(lr){2-4} \cmidrule(l){6-8}
    \textbf{Model} & EM & \multicolumn{1}{c@{\hspace{1.7em}}}{$\Delta$} & 95\% interval & & EM & \multicolumn{1}{c@{\hspace{1.7em}}}{$\Delta$} & 95\% interval \\
    \midrule
    \qwen & 61.6 & $-$3.7\rlap{\smash{$^{*}$}} & [$-$6.8, $-$0.9] &  & 44.3 & $-$4.6\rlap{\smash{$^{*}$}} & [$-$7.7, $-$2.0] \\
    \qwenmoe & 55.0 & $-$10.2\rlap{\smash{$^{*}$}} & [$-$16.7, $-$4.5] &  & 37.0 & $-$11.9\rlap{\smash{$^{*}$}} & [$-$17.4, $-$7.3] \\
    \qwennew & 65.3 & \multicolumn{2}{c}{--} &  & 49.0 & \multicolumn{2}{c@{}}{--} \\
    \gemma & 45.3 & $-$20.0\rlap{\smash{$^{*}$}} & [$-$28.9, $-$11.7] &  & 30.7 & $-$18.3\rlap{\smash{$^{*}$}} & [$-$23.8, $-$12.7] \\
    \gemmanew & 63.3 & $-$2.0 & [$-$4.9, $+$0.5] &  & 47.0 & $-$2.0 & [$-$4.2, $+$0.6] \\
    \gptossl & 61.4 & $-$3.9 & [$-$8.3, $+$0.3] &  & 40.7 & $-$8.2\rlap{\smash{$^{*}$}} & [$-$12.3, $-$4.2] \\
    \gptosss & 57.4 & $-$7.9\rlap{\smash{$^{*}$}} & [$-$13.1, $-$3.1] &  & 29.5 & $-$19.5\rlap{\smash{$^{*}$}} & [$-$24.9, $-$13.5] \\
    \glm & 13.5 & $-$51.7\rlap{\smash{$^{*}$}} & [$-$62.6, $-$39.9] &  & 31.5 & $-$17.4\rlap{\smash{$^{*}$}} & [$-$22.1, $-$12.6] \\
    \gemini & 60.2\rlap{\smash{$^{\dagger}$}} & $-$5.0\rlap{\smash{$^{*\dagger}$}} & [$-$9.5, $-$0.3] &  & 29.5 & $-$19.5\rlap{\smash{$^{*}$}} & [$-$26.0, $-$12.3] \\
    \bottomrule
    \end{tabular}

    \caption{\textbf{Differences in EM to the best model} (\emph{small} split, English, text modality; 4{,}979 \emph{true-premise} questions). \qwennew has the highest $\mathrm{EM}$ at T3-o and at T3a. $\Delta$ is the model's $\mathrm{EM}$ minus that of \qwennew, with the 95\% bootstrap interval (Appendix \ref{sec:appendix_eval_metrics}). $^{*}$ marks intervals that exclude zero. $\mathrm{EM}$ is the mean over three seeds, and over two seeds for \gemini at T3-o ($^{\dagger}$).}

    \label{tab:appendix_model_differences}
\end{table*}

\begin{figure}[!t]
    \centering
    \includegraphics[width=\columnwidth]{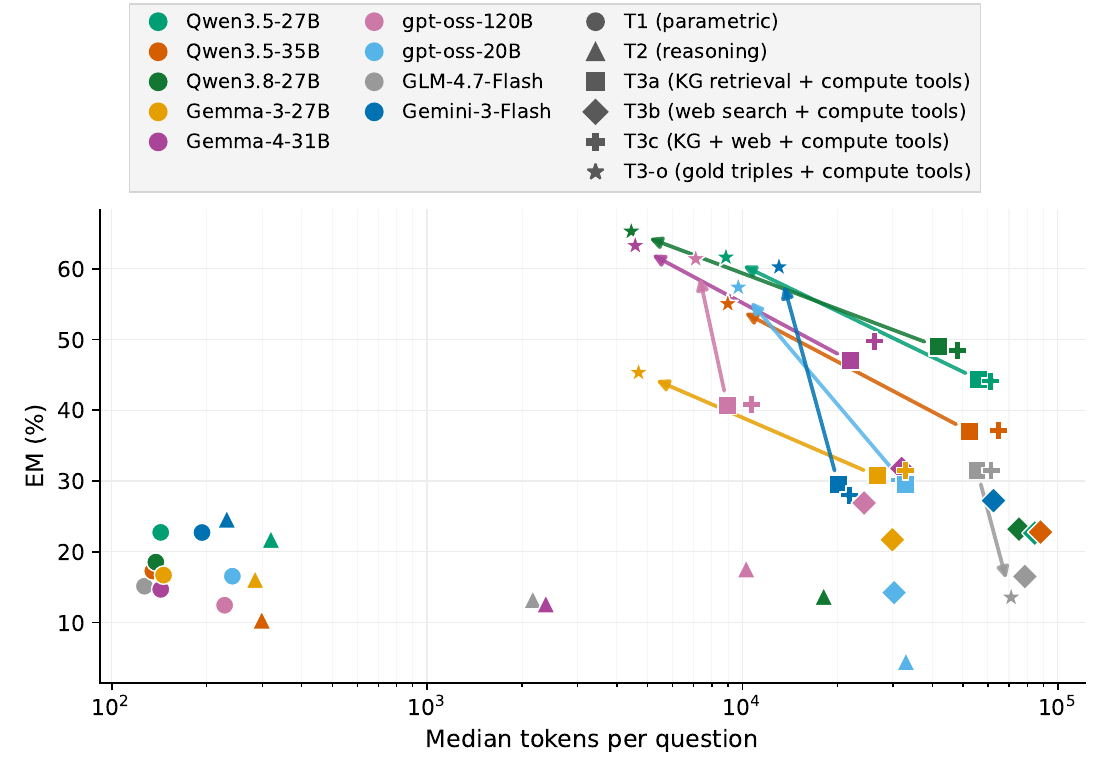}
    \caption{\textbf{Accuracy vs. compute cost} (\emph{small} split, English, text modality; $\mathrm{EM}$ over 4,979 \emph{true-premise} questions against median tokens per question, log scale). Color denotes model, shape denotes tier. Arrows run from T3a to T3-o.}
    \label{fig:appendix_compute_cost}
\end{figure}

In T1-o, we inject the gold triples in three surface forms: raw triples, structured JSON, and verbalized prose (\S\ref{sec:experimental_setup}). Fig. \ref{fig:appendix_oracle_ablation} shows the results. We find that verbalized prose is best for six of the nine models, by up to 4.8 $\mathrm{EM}$ points over structured JSON (\glm; 3.1 for \qwen and 3.0 for \gemma). For \gptossl, \gemmanew and \gemini raw triples score highest and verbalized prose is at most 0.5 points below them. Averaged over the nine models, verbalized prose reaches 45.4 $\mathrm{EM}$ against 44.3 for raw triples and 43.6 for structured JSON, so we use verbalized prose for T1-o throughout.

\subsection{Native Thinking}
\label{sec:appendix_thinking}
\begin{table*}[t]
\centering
\small
\setlength{\tabcolsep}{4pt}
    \begin{tabular}{@{}ll r cccccc @{\hspace{1.6em}}r@{\hspace{1.6em}}l@{}}
    \toprule
    \textbf{Model} & \textbf{Setting} & \textbf{Tokens/step} & \textbf{EM} & \textbf{Cov.} & \textbf{EM\textsubscript{cov}} & \textbf{FRR} & \textbf{Exh.} & \textbf{Err.} & \textbf{$\Delta$EM} & \textbf{95\% interval} \\
    \midrule
    \qwennew & T3-o & 4{,}096 & 65.3 & 84.7 & 77.0 & 6.6 & 6.2 & 2.4 &  &  \\[-0.55ex]
     & & & {\tiny (0.1)} & {\tiny (0.1)} & {\tiny (0.1)} & {\tiny (0.1)} & {\tiny (0.0)} & {\tiny (0.0)} & & \\[0.35ex]
     & T3-o+thinking & 32{,}768 & 67.6\rlap{\smash{$^{\dagger}$}} & 88.6\rlap{\smash{$^{\dagger}$}} & 76.4\rlap{\smash{$^{\dagger}$}} & 5.1\rlap{\smash{$^{\dagger}$}} & 1.4\rlap{\smash{$^{\dagger}$}} & 5.0\rlap{\smash{$^{\dagger}$}} & $+$2.4\rlap{\smash{$^{\dagger}$}} & [$-$1.2, $+$6.6] \\[-0.55ex]
     & & & {\tiny (0.3)} & {\tiny (0.1)} & {\tiny (0.2)} & {\tiny (0.1)} & {\tiny (0.1)} & {\tiny (0.1)} & & \\[0.35ex]
    \midrule
    \gemmanew & T3-o & 4{,}096 & 63.3 & 80.7 & 78.4 & 9.0 & 7.0 & 3.2 &  &  \\[-0.55ex]
     & & & {\tiny (0.2)} & {\tiny (0.3)} & {\tiny (0.1)} & {\tiny (0.1)} & {\tiny (0.1)} & {\tiny (0.3)} & & \\[0.35ex]
     & T3-o+thinking & 16{,}384 & 67.8\rlap{\smash{$^{\dagger}$}} & 87.1\rlap{\smash{$^{\dagger}$}} & 77.8\rlap{\smash{$^{\dagger}$}} & 7.7\rlap{\smash{$^{\dagger}$}} & 0.3\rlap{\smash{$^{\dagger}$}} & 4.8\rlap{\smash{$^{\dagger}$}} & $+$4.6\rlap{\smash{$^{*\dagger}$}} & [$+$2.6, $+$7.4] \\[-0.55ex]
     & & & {\tiny (0.0)} & {\tiny (0.0)} & {\tiny (0.0)} & {\tiny (0.1)} & {\tiny (0.1)} & {\tiny (0.1)} & & \\[0.35ex]
    \midrule
    \glm & T3-o & 4{,}096 & 13.5 & 24.5 & 55.3 & 4.6 & 68.0 & 2.9 &  &  \\[-0.55ex]
     & & & {\tiny (0.3)} & {\tiny (0.3)} & {\tiny (1.0)} & {\tiny (0.1)} & {\tiny (0.3)} & {\tiny (0.0)} & & \\[0.35ex]
     & T3-o+thinking & 32{,}768 & 53.1\rlap{\smash{$^{\dagger}$}} & 73.7\rlap{\smash{$^{\dagger}$}} & 72.0\rlap{\smash{$^{\dagger}$}} & 7.5\rlap{\smash{$^{\dagger}$}} & 4.6\rlap{\smash{$^{\dagger}$}} & 14.2\rlap{\smash{$^{\dagger}$}} & $+$39.5\rlap{\smash{$^{*\dagger}$}} & [$+$27.2, $+$51.8] \\[-0.55ex]
     & & & {\tiny (0.2)} & {\tiny (0.3)} & {\tiny (0.0)} & {\tiny (0.0)} & {\tiny (0.1)} & {\tiny (0.1)} & & \\[0.35ex]
    \bottomrule
    \end{tabular}

    \caption{\textbf{Native thinking with gold triples} (\emph{small} split, English, text modality; 4,979 \emph{true-premise} questions; $\mathrm{EM}\uparrow$, $\mathrm{FRR}\downarrow$). In T3-o+thinking native thinking is switched on and the output budget per agent step is raised (\emph{Tokens/step}). Parentheses give standard deviation over three seeds for T3-o and two seeds for T3-o+thinking ($^{\dagger}$). $\Delta\mathrm{EM}$ is \textit{T3-o+thinking - T3-o}, with a 95\% bootstrap interval (Appendix \ref{sec:appendix_eval_metrics}). $^{*}$ marks intervals that exclude zero.}
    \label{tab:appendix_thinking}
\end{table*}

We repeat T3-o experiments for \qwennew, \gemmanew and \glm with native thinking and a larger output budget per agent step (T3-o+thinking). $\mathrm{EM}$ rises for all three, by 2.4 to 39.5 points, and the gain is statistically significant for \gemmanew and \glm (Table \ref{tab:appendix_thinking}). It comes from runs that now finish, as budget exhaustion falls to at most 4.6\% of the questions, while the accuracy of \qwennew and \gemmanew on answered questions does not rise. Thinking and the output budget change together, so we report T3-o+thinking separately from T3-o.

\subsection{Frontier-Model Probe}
\label{sec:appendix_frontier_probe}
\begin{table*}[t]
\centering
\small
\newcommand{\sdv}[1]{\textsubscript{\tiny #1}}

    \textbf{(a) All models on the probe questions}\\[0.8ex]
    \setlength{\tabcolsep}{2.5pt}
    \begin{tabular}{@{}lccccccccc@{\hspace{8pt}}c@{}}
    \toprule
     & \textbf{Qwen3.5} & \textbf{Qwen3.5} & \textbf{Qwen3.8} & \textbf{Gemma-3} & \textbf{Gemma-4} & \textbf{gpt-oss} & \textbf{gpt-oss} & \textbf{GLM-4.7} & \textbf{Gemini-3} & \textbf{Claude} \\
     & \textbf{27B} & \textbf{35B} & \textbf{27B} & \textbf{27B} & \textbf{31B} & \textbf{120B} & \textbf{20B} & \textbf{Flash} & \textbf{Flash} & \textbf{Fable-5.1} \\
    \midrule
    T3-o & 47.1\sdv{1.9} & 37.6\sdv{1.3} & 52.9\sdv{0.9} & 26.3\sdv{1.1} & 47.4\sdv{0.9} & 49.8\sdv{1.9} & 45.6\sdv{1.9} & 13.5\sdv{1.1} & 41.7$^{\dagger}$\sdv{1.4} & \multirow{2}{*}{73.4$^{\ddagger}$} \\
    T3-o+thinking$^{\dagger}$ & -- & -- & 64.2\sdv{0.9} & -- & 54.1\sdv{1.8} & -- & -- & 40.4\sdv{1.8} & -- & \\
    \bottomrule
    \end{tabular}

    \vspace{2.2ex}

    \textbf{(b) By difficulty}\\[0.8ex]
    \setlength{\tabcolsep}{6pt}
    \begin{tabular}{@{}lccccc@{}}
    \toprule
     & \multicolumn{4}{c}{\textbf{Solved by (of nine models)}} & \\
    \cmidrule(lr){2-5}
     & \textbf{0} & \textbf{1--3} & \textbf{4--6} & \textbf{7--9} & \textbf{All} \\
    \midrule
    Questions & 28 & 28 & 28 & 25 & 109 \\
    \midrule
    \fable & \phantom{0}25.0\phantom{\sdv{0.0}} & \phantom{0}71.4\phantom{\sdv{0.0}} & 100.0\phantom{\sdv{0.0}} & 100.0\phantom{\sdv{0.0}} & 73.4\phantom{\sdv{0.0}} \\
    Best evaluated & \phantom{00}4.8\sdv{3.4} & \phantom{0}39.3\sdv{2.9} & \phantom{0}78.6\sdv{0.0} & 100.0\sdv{0.0} & 52.9\sdv{0.9} \\
    \bottomrule
    \end{tabular}
    \caption{\textbf{Frontier-model probe with gold triples} (109 \emph{true-premise} questions of the \emph{small} split, English, text modality; $\mathrm{EM}\uparrow$). The subset holds at most two questions per spatial-function family and difficulty stratum, and the stratum of a question is the number of the nine evaluated models that answer it correctly at T3-o. (a)~$\mathrm{EM}$ on these questions for the evaluated models, averaged over three seeds, over two where marked ($^{\dagger}$), and for \fable, which runs once per question in its own agent harness ($^{\ddagger}$). (b)~$\mathrm{EM}$ by stratum for \fable and for the evaluated model with the highest $\mathrm{EM}$ in the stratum at T3-o: \qwennew for 0, \gptossl for 1--3, \gemmanew for 4--6, and these three models for 7--9. Subscripts give standard deviation over seeds.}
    \label{tab:appendix_frontier_probe}
\end{table*}
In the probe (Table \ref{tab:appendix_frontier_probe}), \fable receives the gold triples and the answer-format instruction of T3-o and runs with its own agent harness (Table \ref{tab:appendix_models_used}), without a step cap, library restriction or web access.
\fable answers 89.9\% of the questions and is correct on 81.6\% of those. It leads \qwennew by 20.5 points at T3-o (95\% interval [12.5, 28.1]) and by 9.2 points at T3-o+thinking ([2.0, 16.3]), which we include because \fable uses its thinking mode in most sessions. The lead comes from the three harder strata.
The harness of \fable differs from ours, so the difference combines model and harness. The subset samples the difficulty strata evenly and is harder than the benchmark, on which \qwennew scores 12.4 points higher, so the probe does not estimate accuracy on all questions. 

\subsection{Compute Cost}
\label{sec:appendix_cost}

\begin{table*}[t]
\centering
\small
\setlength{\tabcolsep}{3pt}
    \begin{tabular*}{\textwidth}{@{\extracolsep{\fill}}l rgrgrg @{\hskip 14pt} rgrgrg@{\hskip 4pt}}
    \toprule
     & & \textbf{\#tokens/} & \textbf{\#tokens/} & \textbf{\#tool} & \textbf{Exh.} & \textbf{Ovf.} & & \textbf{\#tokens/} & \textbf{\#tokens/} & \textbf{\#tool} & \textbf{Exh.} & \textbf{Ovf.} \\
    \textbf{Model} & \textbf{EM} & \textbf{question} & \textbf{correct} & \textbf{calls} & \textbf{(\%)} & \textbf{(\%)} & \textbf{EM} & \textbf{question} & \textbf{correct} & \textbf{calls} & \textbf{(\%)} & \textbf{(\%)} \\
    \midrule

     & \multicolumn{6}{c}{\emph{T1 (parametric)}} & \multicolumn{6}{c}{\emph{T3b (web search)}} \\
    \cmidrule(lr){2-7} \cmidrule(lr){8-13}
    \qwen & 22.7 & 143 & 629 & -- & 0.2 & 0.0 & 22.6 & 85k & 374k & 8.6 & 60.3 & 4.7 \\
    \qwenmoe & 17.3 & 135 & 782 & -- & 0.0 & 0.0 & 22.8 & 88k & 387k & 8.4 & 54.6 & 8.5 \\
    \qwennew & 18.5 & 138 & 745 & -- & 0.1 & 0.0 & 23.2 & 75k & 326k & 8.6 & 64.8 & 0.0 \\
    \gemma & 16.7 & 146 & 875 & -- & 0.1 & 0.0 & 21.7 & 30k & 138k & 5.6 & 13.4 & 4.3 \\
    \gemmanew & 14.7 & 143 & 974 & -- & 0.0 & 0.0 & 31.8 & 32k & 101k & 5.4 & 1.9 & 0.0 \\
    \gptossl & 12.4 & 228 & 1.8k & -- & 0.0 & 0.0 & 26.9 & 24k & 90k & 5.1 & 18.0 & 0.0 \\
    \gptosss & 16.5 & 241 & 1.5k & -- & 0.0 & 0.0 & 14.2 & 30k & 213k & 2.5 & 22.7 & 0.0 \\
    \glm & 15.1 & 127 & 840 & -- & 0.1 & 0.0 & 16.5 & 79k & 477k & 8.8 & 68.9 & 0.1 \\
    \gemini & 22.7 & 193 & 851 & -- & n/r & 0.0 & 27.2\rlap{\smash{$^{\dagger}$}} & 63k\rlap{\smash{$^{\dagger}$}} & 230k\rlap{\smash{$^{\dagger}$}} & 5.8\rlap{\smash{$^{\dagger}$}} & 24.7\rlap{\smash{$^{\dagger}$}} & 0.0\rlap{\smash{$^{\dagger}$}} \\[0.35ex]
    \hdashline

     & \multicolumn{6}{c}{\emph{T2 (reasoning)}} & \multicolumn{6}{c}{\emph{T3c (\kg + web)}} \\
    \cmidrule(lr){2-7} \cmidrule(lr){8-13}
    \qwen & 21.7 & 320 & 1.5k & -- & 4.2 & 0.0 & 44.1 & 61k & 139k & 6.8 & 38.7 & 1.5 \\
    \qwenmoe & 10.3 & 299 & 2.9k & -- & 1.1 & 0.0 & 37.2 & 65k & 174k & 7.1 & 40.6 & 10.4 \\
    \qwennew & 13.7 & 18k & 132k & -- & 25.7 & 0.0 & 48.5 & 48k & 99k & 6.5 & 35.4 & 0.0 \\
    \gemma & 16.1 & 285 & 1.8k & -- & 3.8 & 0.0 & 31.5 & 33k & 105k & 5.8 & 12.9 & 5.1 \\
    \gemmanew & 12.6 & 2.4k & 19k & -- & 27.9 & 0.0 & 49.8 & 26k & 53k & 4.6 & 2.3 & 0.0 \\
    \gptossl & 17.6 & 10k & 58k & -- & 19.1 & 0.0 & 40.8 & 11k & 26k & 3.0 & 7.8 & 0.0 \\
    \gptosss & 4.5 & 33k & 741k & -- & 86.9 & 0.0 & 30.1 & 31k & 104k & 2.4 & 12.2 & 0.0 \\
    \glm & 13.2 & 2.2k & 16k & -- & 40.0 & 0.0 & 31.5 & 61k & 195k & 7.7 & 48.5 & 0.5 \\
    \gemini & 24.6 & 232 & 942 & -- & n/r & 0.0 & 28.0\rlap{\smash{$^{\dagger}$}} & 22k\rlap{\smash{$^{\dagger}$}} & 78k\rlap{\smash{$^{\dagger}$}} & 2.2\rlap{\smash{$^{\dagger}$}} & 22.1\rlap{\smash{$^{\dagger}$}} & 0.0\rlap{\smash{$^{\dagger}$}} \\[0.35ex]
    \hdashline

     & \multicolumn{6}{c}{\emph{T3a (\kg retrieval)}} & \multicolumn{6}{c}{\emph{T3-o (gold triples injected)}} \\
    \cmidrule(lr){2-7} \cmidrule(lr){8-13}
    \qwen & 44.3 & 56k & 127k & 6.7 & 37.0 & 1.7 & 61.6 & 8.9k & 14k & 2.6 & 5.6 & 8.9 \\
    \qwenmoe & 37.0 & 52k & 141k & 6.8 & 38.4 & 11.7 & 55.0 & 9.0k & 16k & 2.3 & 4.7 & 19.6 \\
    \qwennew & 49.0 & 42k & 85k & 6.3 & 31.9 & 0.0 & 65.3 & 4.4k & 6.8k & 2.1 & 6.2 & 2.4 \\
    \gemma & 30.7 & 27k & 87k & 5.3 & 8.5 & 4.7 & 45.3 & 4.7k & 10k & 2.6 & 12.3 & 16.2 \\
    \gemmanew & 47.0 & 22k & 47k & 4.1 & 1.9 & 0.0 & 63.3 & 4.6k & 7.2k & 2.0 & 7.0 & 2.5 \\
    \gptossl & 40.7 & 9.0k & 22k & 2.9 & 8.4 & 0.0 & 61.4 & 7.1k & 12k & 1.8 & 5.5 & 2.8 \\
    \gptosss & 29.5 & 33k & 111k & 2.6 & 16.6 & 0.0 & 57.4 & 9.7k & 17k & 1.7 & 9.2 & 3.0 \\
    \glm & 31.5 & 55k & 176k & 7.4 & 44.3 & 0.5 & 13.5 & 71k & 526k & 7.4 & 68.0 & 2.9 \\
    \gemini & 29.5 & 20k & 72k & 2.9 & 11.5 & 0.0 & 60.2\rlap{\smash{$^{\dagger}$}} & 13k\rlap{\smash{$^{\dagger}$}} & 22k\rlap{\smash{$^{\dagger}$}} & 2.1\rlap{\smash{$^{\dagger}$}} & 19.7\rlap{\smash{$^{\dagger}$}} & 1.2\rlap{\smash{$^{\dagger}$}} \\
    \bottomrule
    \end{tabular*}

    \caption{\textbf{Compute costs} (\emph{small} split, English, text modality; 4,979 \emph{true-premise} questions). Values are averaged over three seeds; $^{\dagger}$ marks cells with fewer seeds. $\mathrm{EM}$ is computed over all questions. \emph{\#tokens/question} is the median of input plus output tokens, summed over all model calls of a question, over the questions that do not end in a harness error. \emph{\#tokens/correct} is this median divided by $\mathrm{EM}$. \emph{\#tool calls} is the mean number of executed code steps over the same questions and does not apply without an agent (--). \emph{Exh.} is the share of all questions that end in budget exhaustion (\S\ref{sec:experimental_setup}), and \emph{Ovf.} the share that end in a context-overflow error, i.e., the request exceeds the model's context window. Output tokens include reasoning tokens for the open-weight models and exclude them for \gemini. \gemini's responses record no stop reason, so its cut-off completions at T1 and T2 cannot be counted (n/r).}

    \label{tab:appendix_compute_cost}
\end{table*}

Web search is the most expensive source for eight of nine models, at 24k-88k median tokens (Table \ref{tab:appendix_compute_cost}, Fig. \ref{fig:appendix_compute_cost}), and adding it to the \kg (T3c) raises cost over \kg retrieval alone (T3a), while changing accuracy by at most 2.8 points (\gemmanew). \gptosss is the exception.
The oracle runs the same agent with the same step budget and is cheaper for eight of nine models, by 1.3$\times$ (\gptossl) to 9.4$\times$ (\qwennew), while scoring 15 to 31 points higher. Their tool calls fall from 2.6-6.8 per question to 1.7-2.6. \glm uses 1.3 times as many tokens at T3-o as at T3a and scores 18.0 points lower.

\subsection{Performance by \sfid and Answer Format}
\label{sec:appendix_performance_sfid_answer}

\begin{figure*}[t]
    \centering
    \includegraphics[width=\textwidth]{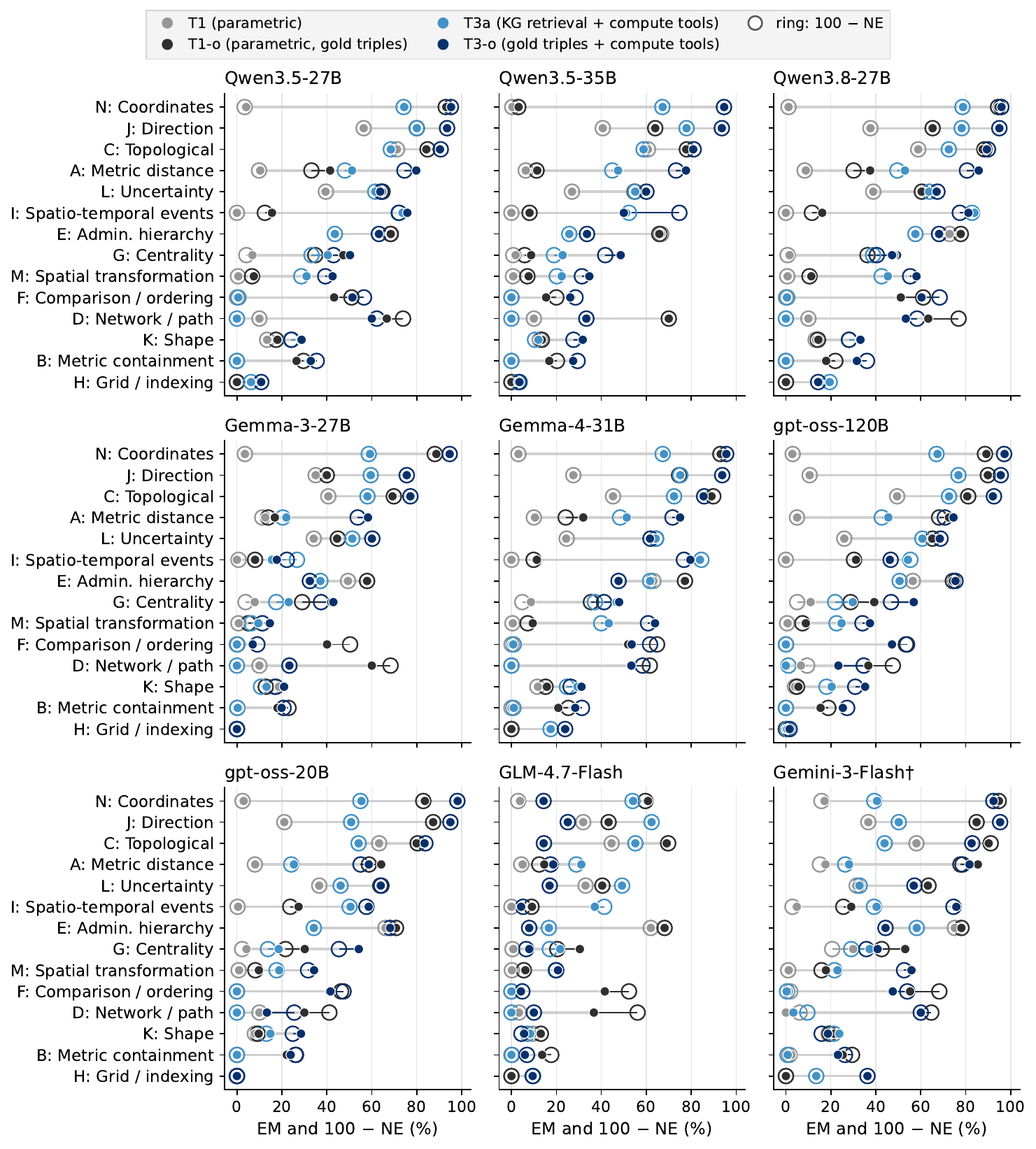}
    \caption{\textbf{Accuracy ($\mathrm{EM}$) and normalized error ($\mathrm{NE}$) by spatial function and model across evaluation conditions} (\emph{small} split, English, text modality; 4,979 \emph{true-premise} questions). Dots mark $\mathrm{EM}$ and rings $100{-}\mathrm{NE}$. Each value is the mean over three seeds. For \gemini ($\dagger$), the T1-o and T3-o values are means over two seeds. Rows are ordered as in Fig.~\ref{fig:sfid_performance_em_overall}.}   
    \label{fig:appendix_sfid_performance_em_model}
\end{figure*}

\begin{figure*}[t]
    \centering
    \includegraphics[width=\textwidth]{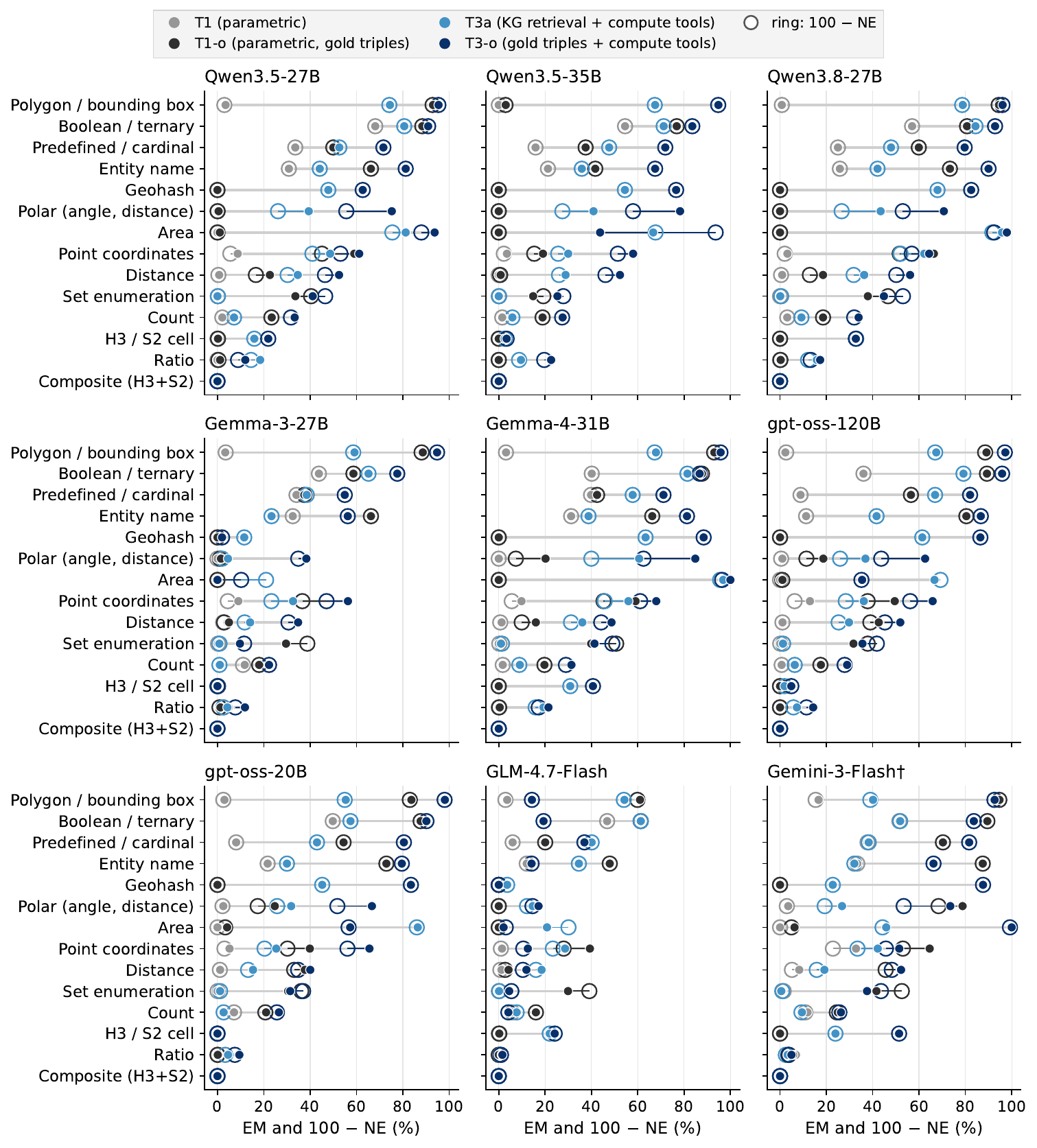}
    \caption{\textbf{Accuracy ($\mathrm{EM}$) and normalized error ($\mathrm{NE}$) by answer format and model across evaluation conditions} (\emph{small} split, English, text modality; 4,979 \emph{true-premise} questions). Dots mark $\mathrm{EM}$ and rings $100{-}\mathrm{NE}$. Each value is the mean over three seeds. For \gemini ($\dagger$), the T1-o and T3-o values are means over two seeds. Rows are ordered as in Fig.~\ref{fig:answer_format_em_overall}. We omit \emph{Date} as it covers only 8 questions.}
    \label{fig:appendix_answer_format_em_model}
\end{figure*}

Figs. \ref{fig:appendix_sfid_performance_em_model} and \ref{fig:appendix_answer_format_em_model} break Fig.~\ref{fig:rem_sfid_answer} down by model.

\subsection{Failure Analysis}
\label{sec:appendix_failure_analysis}
\begin{table*}[t]
\centering
\small
\setlength{\tabcolsep}{3.2pt}
    \begin{tabular}{@{}l cgcgc cgcgc cgc@{}}
    \toprule
     & \multicolumn{5}{c}{\textbf{(a) Outcome at T3a}} & \multicolumn{5}{c}{\textbf{(a) Outcome at T3-o}} & \multicolumn{3}{c}{\textbf{(b) Failure causes}} \\
    \cmidrule(lr){2-6} \cmidrule(lr){7-11} \cmidrule(l){12-14}
     & & & & & & & & & & & Empty & Failed & Missing \\
     & & & & & & & & & & & lookup & step & \texttt{h3} \\
    \textbf{Model} & Correct & Wrong & Abst. & Exh. & Err. & Correct & Wrong & Abst. & Exh. & Err. & T3a & T3a & T3-o \\
    \midrule
    \qwen & 44.3 & 16.9 & 0.2 & 37.0 & 1.7 & 61.6 & 18.4 & 5.5 & 5.6 & 8.9 & 43.5 & 29.8 & 92.8 \\[-0.55ex]
     & {\tiny (0.4)} & {\tiny (0.3)} & {\tiny (0.0)} & {\tiny (0.4)} & {\tiny (0.0)} & {\tiny (0.2)} & {\tiny (0.3)} & {\tiny (0.1)} & {\tiny (0.2)} & {\tiny (0.1)} & {\tiny (0.1)} & {\tiny (0.7)} & {\tiny (1.2)} \\[0.35ex]
    \qwenmoe & 37.0 & 12.5 & 0.3 & 38.4 & 11.7 & 55.0 & 14.7 & 6.0 & 4.7 & 19.6 & 43.1 & 31.9 & 51.6 \\[-0.55ex]
     & {\tiny (0.1)} & {\tiny (0.0)} & {\tiny (0.0)} & {\tiny (0.1)} & {\tiny (0.1)} & {\tiny (0.2)} & {\tiny (0.2)} & {\tiny (0.1)} & {\tiny (0.3)} & {\tiny (0.2)} & {\tiny (0.1)} & {\tiny (0.5)} & {\tiny (4.1)} \\[0.35ex]
    \qwennew & 49.0 & 19.0 & 0.1 & 31.9 & 0.0 & 65.3 & 19.5 & 6.6 & 6.2 & 2.4 & 37.7 & 22.3 & 5.2 \\[-0.55ex]
     & {\tiny (0.0)} & {\tiny (0.1)} & {\tiny (0.0)} & {\tiny (0.1)} & {\tiny (0.0)} & {\tiny (0.1)} & {\tiny (0.1)} & {\tiny (0.1)} & {\tiny (0.0)} & {\tiny (0.0)} & {\tiny (0.2)} & {\tiny (0.1)} & {\tiny (0.7)} \\[0.35ex]
    \gemma & 30.7 & 28.6 & 27.5 & 8.5 & 4.7 & 45.3 & 16.6 & 9.6 & 12.3 & 16.2 & 34.5 & 49.3 & 47.0 \\[-0.55ex]
     & {\tiny (0.0)} & {\tiny (0.0)} & {\tiny (0.1)} & {\tiny (0.1)} & {\tiny (0.1)} & {\tiny (0.2)} & {\tiny (0.2)} & {\tiny (0.4)} & {\tiny (0.2)} & {\tiny (0.1)} & {\tiny (0.0)} & {\tiny (0.0)} & {\tiny (0.5)} \\[0.35ex]
    \gemmanew & 47.0 & 34.3 & 16.8 & 1.9 & 0.0 & 63.3 & 17.4 & 9.0 & 7.0 & 3.2 & 33.8 & 28.1 & 96.0 \\[-0.55ex]
     & {\tiny (0.0)} & {\tiny (0.0)} & {\tiny (0.0)} & {\tiny (0.0)} & {\tiny (0.0)} & {\tiny (0.2)} & {\tiny (0.1)} & {\tiny (0.1)} & {\tiny (0.1)} & {\tiny (0.3)} & {\tiny (0.0)} & {\tiny (0.1)} & {\tiny (0.5)} \\[0.35ex]
    \gptossl & 40.7 & 36.7 & 14.0 & 8.4 & 0.2 & 61.4 & 19.0 & 11.2 & 5.5 & 2.8 & 17.1 & 36.8 & 96.1 \\[-0.55ex]
     & {\tiny (0.2)} & {\tiny (0.1)} & {\tiny (0.1)} & {\tiny (0.0)} & {\tiny (0.0)} & {\tiny (0.1)} & {\tiny (0.3)} & {\tiny (0.2)} & {\tiny (0.2)} & {\tiny (0.0)} & {\tiny (0.4)} & {\tiny (0.8)} & {\tiny (0.8)} \\[0.35ex]
    \gptosss & 29.5 & 27.5 & 26.3 & 16.6 & 0.1 & 57.4 & 15.8 & 14.6 & 9.2 & 3.0 & 4.4\rlap{\smash{$^{\ddagger}$}} & 57.7 & 77.4 \\[-0.55ex]
     & {\tiny (0.5)} & {\tiny (0.3)} & {\tiny (0.2)} & {\tiny (0.7)} & {\tiny (0.0)} & {\tiny (0.2)} & {\tiny (0.0)} & {\tiny (0.0)} & {\tiny (0.2)} & {\tiny (0.0)} & {\tiny (0.1)} & {\tiny (0.8)} & {\tiny (1.6)} \\[0.35ex]
    \glm & 31.5 & 19.5 & 4.3 & 44.3 & 0.5 & 13.5 & 10.9 & 4.6 & 68.0 & 2.9 & 41.3 & 59.1 & 41.7 \\[-0.55ex]
     & {\tiny (0.4)} & {\tiny (0.2)} & {\tiny (0.1)} & {\tiny (0.6)} & {\tiny (0.1)} & {\tiny (0.3)} & {\tiny (0.2)} & {\tiny (0.1)} & {\tiny (0.3)} & {\tiny (0.0)} & {\tiny (0.6)} & {\tiny (0.4)} & {\tiny (1.2)} \\[0.35ex]
    \gemini & 29.5 & 43.9 & 15.2 & 11.5 & 0.0 & 60.2\rlap{\smash{$^{\dagger}$}} & 13.6\rlap{\smash{$^{\dagger}$}} & 3.9\rlap{\smash{$^{\dagger}$}} & 19.7\rlap{\smash{$^{\dagger}$}} & 2.5\rlap{\smash{$^{\dagger}$}} & 33.7 & 53.6 & 42.2\rlap{\smash{$^{\dagger}$}} \\[-0.55ex]
     & {\tiny (3.5)} & {\tiny (3.3)} & {\tiny (10.9)} & {\tiny (5.1)} & {\tiny (0.0)} & {\tiny (3.1)} & {\tiny (2.5)} & {\tiny (0.3)} & {\tiny (4.5)} & {\tiny (1.3)} & {\tiny (6.2)} & {\tiny (19.2)} & {\tiny (12.4)} \\[0.35ex]
    \midrule
    \emph{Average} & 37.7 & 26.5 & 11.6 & 22.1 & 2.1 & 53.7 & 16.2 & 7.9 & 15.4 & 6.8 & 32.1 & 41.0 & 61.1 \\
    \bottomrule
    \end{tabular}

    \caption{\textbf{Question outcomes and failure causes under agentic retrieval} (\emph{small} split, English, text modality; \% of the 4{,}979 \emph{true-premise} questions unless stated). (a)~Outcome of each question at T3a (\kg retrieval) and T3-o (gold triples injected), with the outcome classes of Appendix \ref{sec:appendix_eval_metrics}. \emph{Wrong} includes the answers that the scorer could not parse (at most 0.2\% of the questions of a model). (b)~Share of questions with at least one agent step of the given kind. \emph{Empty lookup} is the share at T3a in which a \kg lookup returns nothing, because the entity identifier is unknown or a name or type search has no match. The detector reads the output of each step as stored in the trace, where a tool reply appears only if the code prints it, so the share is a lower bound. \emph{Failed step} is the share at T3a in which a code step ends in an error. \emph{Missing} \texttt{h3} is the share of the 346 grid-indexing questions at T3-o in which the code calls an \texttt{h3} function that does not exist. Parentheses give standard deviation over three seeds; $^{\dagger}$ marks cells with fewer seeds. $^{\ddagger}$~\gptosss prints a tool reply for 1.9\% of the questions.}

    \label{tab:appendix_failure_analysis}
\end{table*}

Table \ref{tab:appendix_failure_analysis} gives the outcome of every question at T3a and T3-o per model, with three failure causes detected in the stored agent traces. Under \kg retrieval, the two failure patterns of \S\ref{sec:error_analysis} are complementary, as the models that exhaust their budget abstain on at most 4.3\% of the questions and those that abstain exhaust it on at most 16.6\%. The nonexistent \texttt{h3} function is most often \texttt{geo\_to\_h3}, a name from an earlier version of the library, and the share of grid-indexing questions with a call to a nonexistent \texttt{h3} function ranges from 5.2\% (\qwennew) to 96.1\% (\gptossl).

\subsection{Question Complexity}
\label{sec:appendix_complexity}
\begin{table*}[t]
\centering
\small
\setlength{\tabcolsep}{4.5pt}
    \begin{tabular}{@{}l cgcg cgcg cc@{}}
    \toprule
     & \multicolumn{4}{c}{\textbf{T3a, by gold facts}} & \multicolumn{4}{c}{\textbf{T3-o, by gold facts}} & \multicolumn{2}{c}{\textbf{Polygon $-$ point}} \\
    \cmidrule(lr){2-5} \cmidrule(lr){6-9} \cmidrule(l){10-11}
    \textbf{Model} & 1--4 & 5--8 & 9--16 & 17+ & 1--4 & 5--8 & 9--16 & 17+ & T3a & T3-o \\
    \midrule
    \emph{Questions} & 1{,}397 & 1{,}602 & 1{,}427 & 553 & 1{,}397 & 1{,}602 & 1{,}427 & 553 & \multicolumn{2}{c}{1{,}567 and 3{,}240} \\
    \midrule
    \qwen & 35.8 & 52.8 & 59.5 & 2.4 & 43.3 & 65.8 & 79.7 & 49.2 & $-$11.0 & $-$19.9 \\[-0.55ex]
     & {\tiny (0.6)} & {\tiny (0.2)} & {\tiny (0.5)} & {\tiny (0.1)} & {\tiny (0.6)} & {\tiny (0.1)} & {\tiny (0.2)} & {\tiny (0.1)} & {\tiny (0.5)} & {\tiny (0.2)} \\[0.35ex]
    \qwenmoe & 26.6 & 44.7 & 52.3 & 1.9 & 41.5 & 61.1 & 72.9 & 25.6 & $-$25.7 & $-$22.6 \\[-0.55ex]
     & {\tiny (0.1)} & {\tiny (0.2)} & {\tiny (0.1)} & {\tiny (0.1)} & {\tiny (0.4)} & {\tiny (0.2)} & {\tiny (0.1)} & {\tiny (0.6)} & {\tiny (0.3)} & {\tiny (0.5)} \\[0.35ex]
    \qwennew & 42.1 & 56.9 & 64.8 & 2.5 & 48.4 & 68.8 & 81.1 & 56.8 & $-$8.1 & $-$23.0 \\[-0.55ex]
     & {\tiny (0.1)} & {\tiny (0.2)} & {\tiny (0.2)} & {\tiny (0.2)} & {\tiny (0.4)} & {\tiny (0.1)} & {\tiny (0.1)} & {\tiny (0.6)} & {\tiny (0.2)} & {\tiny (0.3)} \\[0.35ex]
    \gemma & 20.2 & 39.1 & 43.0 & 1.5 & 34.6 & 53.1 & 61.5 & 8.4 & $-$13.1 & $-$18.7 \\[-0.55ex]
     & {\tiny (0.0)} & {\tiny (0.2)} & {\tiny (0.1)} & {\tiny (0.1)} & {\tiny (0.2)} & {\tiny (0.3)} & {\tiny (0.3)} & {\tiny (0.6)} & {\tiny (0.1)} & {\tiny (0.4)} \\[0.35ex]
    \gemmanew & 37.5 & 54.8 & 64.4 & 3.4 & 49.7 & 67.4 & 76.8 & 50.6 & $-$6.7 & $-$28.0 \\[-0.55ex]
     & {\tiny (0.0)} & {\tiny (0.0)} & {\tiny (0.1)} & {\tiny (0.0)} & {\tiny (0.0)} & {\tiny (0.3)} & {\tiny (0.5)} & {\tiny (0.4)} & {\tiny (0.0)} & {\tiny (0.8)} \\[0.35ex]
    \gptossl & 28.1 & 50.5 & 57.0 & 2.6 & 42.9 & 67.6 & 79.7 & 42.9 & $-$13.5 & $-$13.4 \\[-0.55ex]
     & {\tiny (0.2)} & {\tiny (0.3)} & {\tiny (0.3)} & {\tiny (0.2)} & {\tiny (0.2)} & {\tiny (0.3)} & {\tiny (0.3)} & {\tiny (0.2)} & {\tiny (0.9)} & {\tiny (0.3)} \\[0.35ex]
    \gptosss & 21.5 & 37.2 & 39.1 & 2.3 & 41.5 & 63.5 & 73.1 & 39.3 & $-$10.0 & $-$12.8 \\[-0.55ex]
     & {\tiny (0.7)} & {\tiny (0.4)} & {\tiny (0.7)} & {\tiny (0.2)} & {\tiny (0.1)} & {\tiny (0.1)} & {\tiny (1.0)} & {\tiny (1.4)} & {\tiny (0.4)} & {\tiny (0.2)} \\[0.35ex]
    \glm & 22.1 & 38.9 & 44.0 & 1.7 & 9.8 & 13.9 & 19.7 & 5.8 & $-$17.9 & $-$6.9 \\[-0.55ex]
     & {\tiny (0.2)} & {\tiny (0.6)} & {\tiny (0.7)} & {\tiny (0.1)} & {\tiny (0.3)} & {\tiny (0.9)} & {\tiny (0.2)} & {\tiny (1.0)} & {\tiny (0.3)} & {\tiny (0.6)} \\[0.35ex]
    \gemini & 21.4 & 34.5 & 42.5 & 1.6 & 45.1\rlap{\smash{$^{\dagger}$}} & 64.1\rlap{\smash{$^{\dagger}$}} & 76.9\rlap{\smash{$^{\dagger}$}} & 44.4\rlap{\smash{$^{\dagger}$}} & $+$2.0 & $-$32.5\rlap{\smash{$^{\dagger}$}} \\[-0.55ex]
     & {\tiny (3.1)} & {\tiny (4.0)} & {\tiny (4.5)} & {\tiny (0.3)} & {\tiny (0.7)} & {\tiny (1.2)} & {\tiny (3.6)} & {\tiny (13.5)} & {\tiny (2.3)} & {\tiny (0.4)} \\[0.35ex]
    \midrule
    \emph{Average} & 28.4 & 45.5 & 51.8 & 2.2 & 39.6 & 58.4 & 69.0 & 35.9 & $-$11.6 & $-$19.8 \\
    \bottomrule
    \end{tabular}

    \caption{\textbf{Accuracy ($\mathrm{EM}$) by question complexity} (\emph{small} split, English, text modality; 4{,}979 \emph{true-premise} questions). Columns 2 to 9 give $\mathrm{EM}$ by the number of distinct gold triples (facts) of a question at T3a and T3-o. The last two columns give $\mathrm{EM}$ on polygon questions minus $\mathrm{EM}$ on point questions, in percentage points. A question is a polygon question if at least one entity in its gold triples has a polygon geometry (1{,}567 questions) and a point question if all of them have point geometries (3{,}240). The remaining 172 questions have a line geometry and no polygon geometry, and they are not included in either group. Parentheses give standard deviation over three seeds; $^{\dagger}$ marks cells with two seeds.}

    \label{tab:appendix_complexity}
\end{table*}

Table \ref{tab:appendix_complexity} relates $\mathrm{EM}$ to a question's number of gold facts and the geometry type of its entities. With gold triples, questions with 17 or more facts remain limited by context overflow. Polygon questions score below point questions for all models at T3-o and for the eight open-weight models at T3a.

\subsection{Income \& Density Equity}
\label{sec:appendix_equity}
\begin{figure*}[!t]
    \centering
    \includegraphics[width=\textwidth]{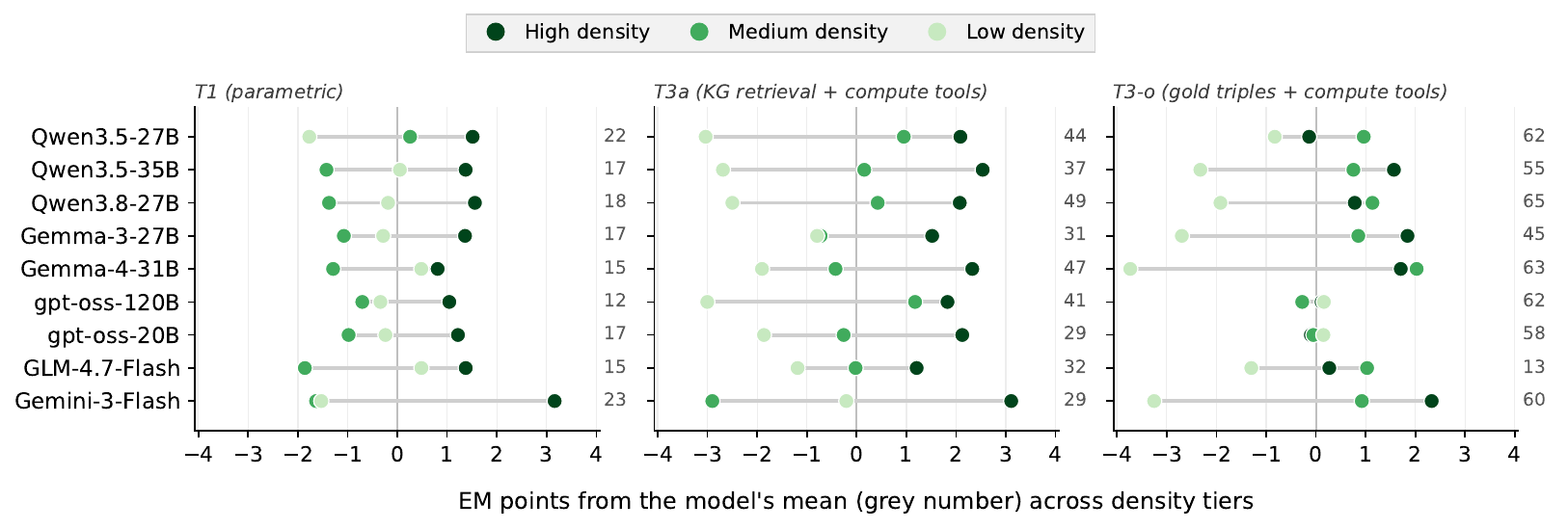}
    \caption{\textbf{Accuracy ($\mathrm{EM}$) by population-density tier} at T1, T3a and T3-o (\emph{small} split, English, text modality;  4{,}979 \emph{true-premise} questions; standardized $\mathrm{EM}$). Each row is one model under one condition, with every tier placed at its distance from that model's mean across tiers (grey number). Standardization follows \S\ref{sec:stratified_analysis}, over the families that the tier contains.}

    \label{fig:appendix_density_equity_small}
\end{figure*}

\begin{figure*}[t]
    \centering
    \includegraphics[width=\textwidth]{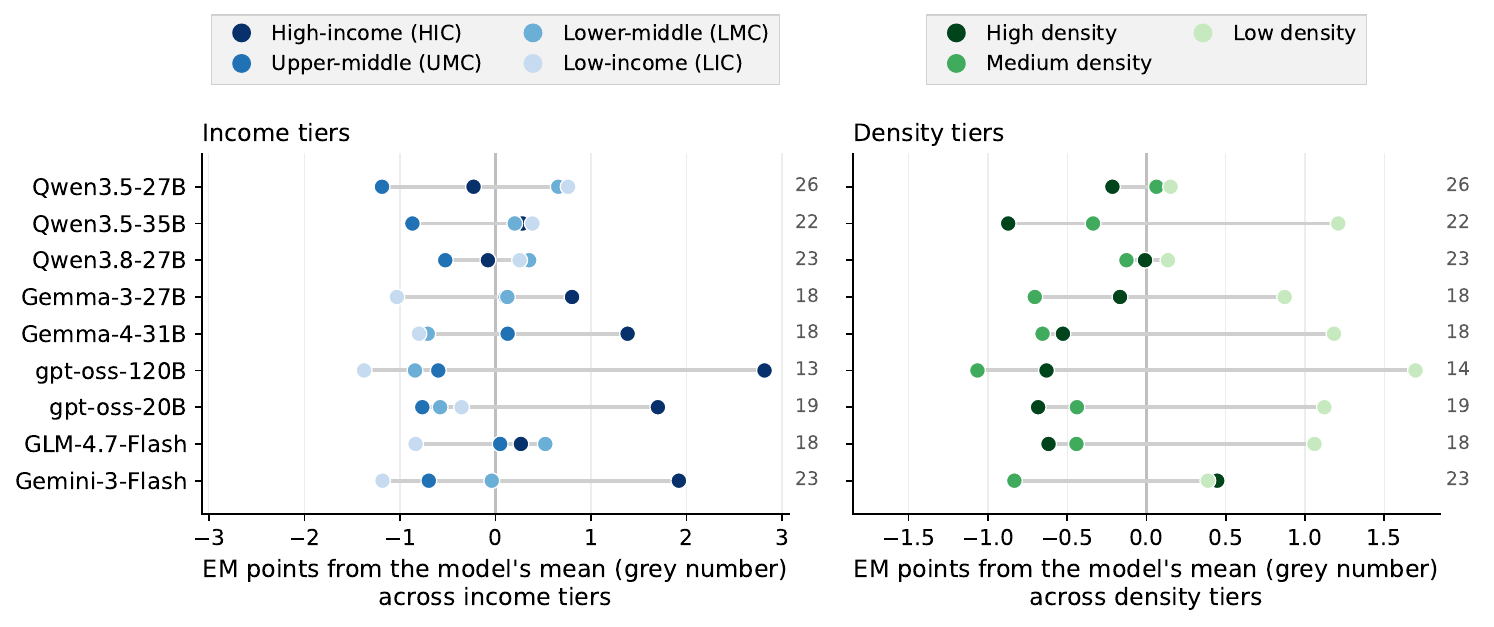}
    \caption{\textbf{Accuracy ($\mathrm{EM}$) by income and density tier on the \emph{large} split} (English, text modality, T1, 41,540 \emph{true-premise} questions; standardized $\mathrm{EM}$ with the spatial-function distribution of the \emph{large} split as weights). Each row places every tier at its distance from that model's mean across tiers (grey number).}
    \label{fig:appendix_equity_large}
\end{figure*}

Fig.~\ref{fig:appendix_density_equity_small} repeats the analysis of \S\ref{sec:stratified_analysis} for the three population-density tiers. On the \emph{small} split, questions about high-density regions score higher than those about low-density regions in 25 of the 27 model-condition cases, by up to 5.6 points. This does not hold at T1 on the \emph{large} split (Fig.~\ref{fig:appendix_equity_large}), where the gap is negative for all eight open-weight models ($-$2.3 to $-$0.1) and +0.1 for \gemini. The \emph{small} split samples up to 200 questions per (\kg, \sfid) cell, so its density composition differs from the benchmark's. The income gap at T1 is also less consistent on the \emph{large} split, positive for six of the nine models (+1.1 to +4.2) and negative for the three Qwen models ($-$1.0 to $-$0.1).

\subsection{False-premise Questions}
\label{sec:appendix_false_premise}

\begin{figure*}[t]
    \centering
    \includegraphics[width=\textwidth]{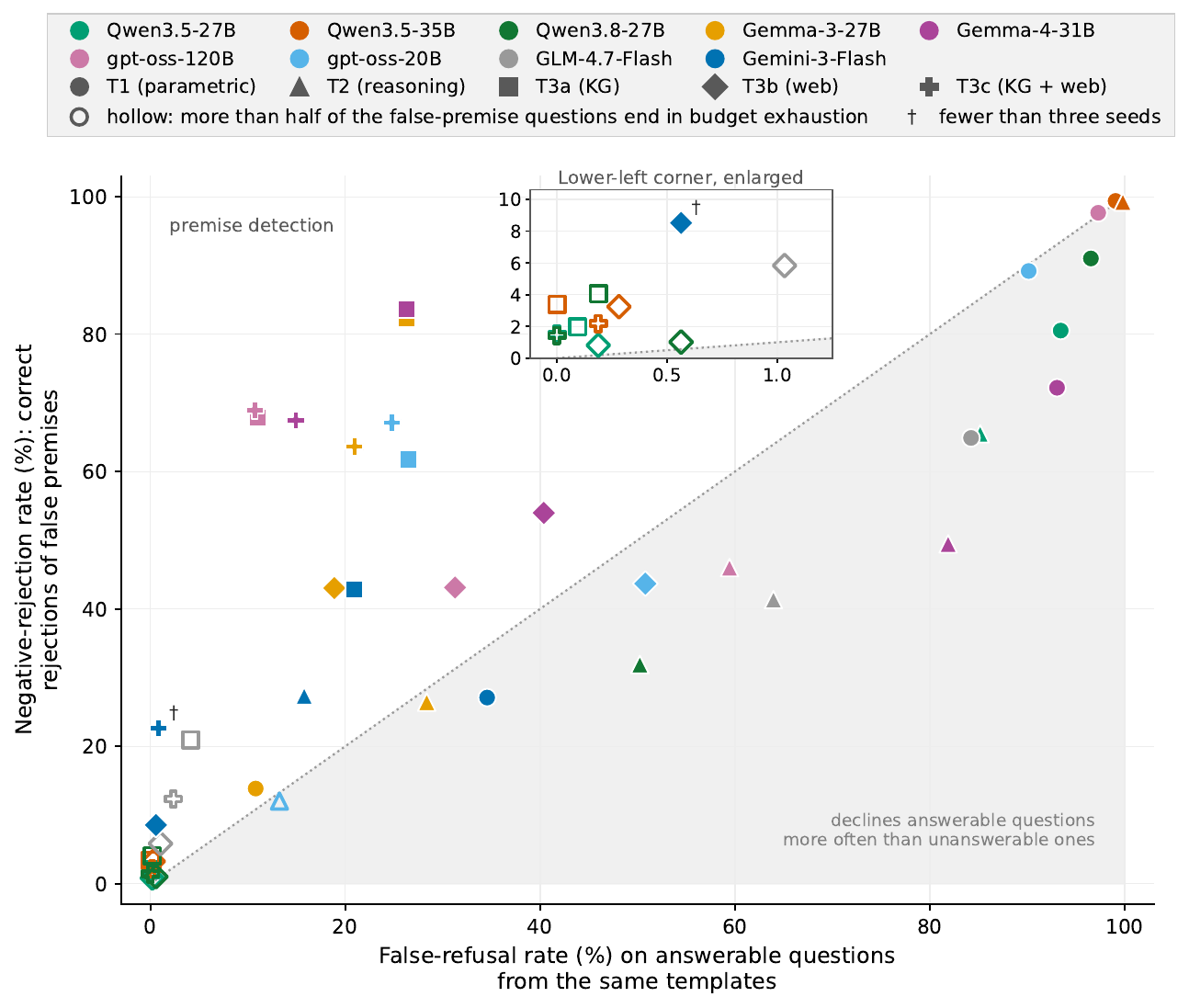}
    \caption{\textbf{Premise detection against indiscriminate abstention} (\emph{small} split, English, text modality; 657 \emph{false-premise} questions and the 355 \emph{true-premise} questions of the same 13 sub-templates). Color denotes model, shape denotes tier. $\mathrm{NRR}$ is the share of the false-premise questions scored correct without budget exhaustion (explicit declines and correct typed answers), and $\mathrm{FRR}$ is computed over the matched true-premise questions. The dotted line is chance. Hollow markers denote cells in which more than half of the false-premise questions end in budget exhaustion, and the inset enlarges the lower-left corner. $^{\dagger}$ marks cells with fewer than three seeds (\gemini: one at T3b, two at T3c).}

    \label{fig:appendix_false_premise}
\end{figure*}

False premises occur in 13 of the 124 sub-templates of the \emph{small} split. Fig.~\ref{fig:appendix_false_premise} compares the rejection of their false-premise questions with the refusal of their answerable ones. We use this matched $\mathrm{FRR}$ since, without retrieval, all models refuse the answerable questions of these sub-templates more often than the benchmark as a whole, by 7.1 to 48.4 points, so the overall $\mathrm{FRR}$ would overstate selectivity. Under retrieval the matched and the overall $\mathrm{FRR}$ lie within 9.5 points of each other. Moreover, the Qwen models and \glm exhaust their budget under retrieval on 74.4\% to 96.3\% of these answerable questions (hollow markers), which keeps their $\mathrm{FRR}$ at or below 4.1\%.
\begin{figure*}[t]
    \centering
    \begin{subfigure}[t]{0.32\textwidth}
        \centering
        \includegraphics[width=\linewidth]{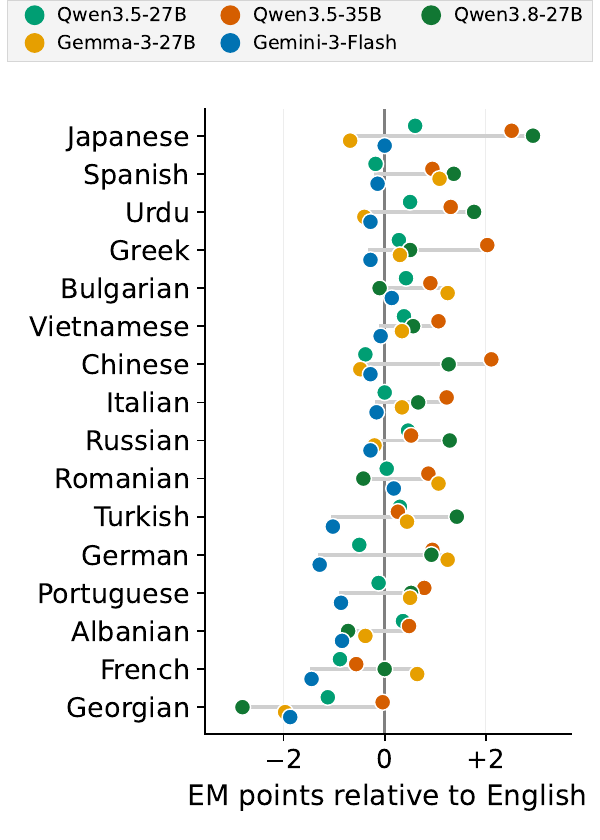}
        \caption{T1, five models.}
        \label{fig:appendix_multilingual_t1}
    \end{subfigure}
    \hfill
    \begin{subfigure}[t]{0.32\textwidth}
        \centering
        \includegraphics[width=\linewidth]{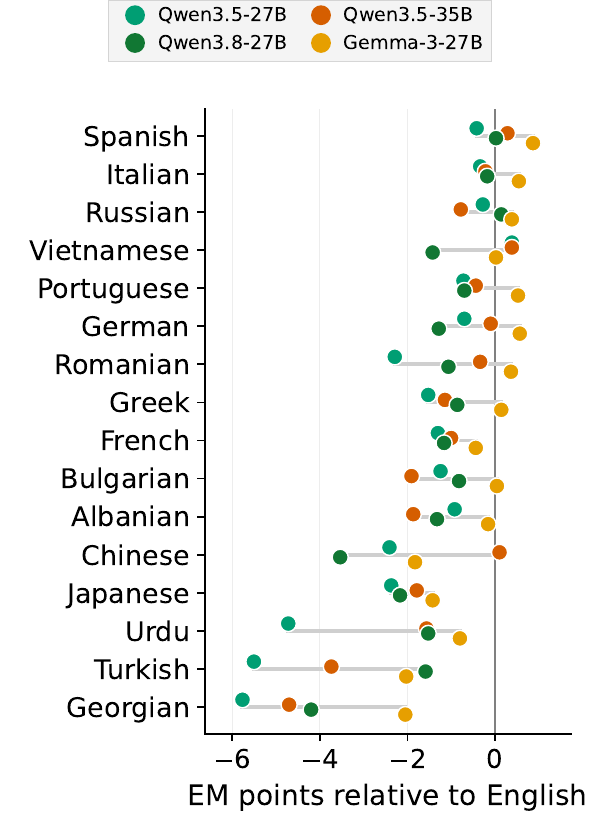}
        \caption{T1-o, four models.}
        \label{fig:appendix_multilingual_t1o}
    \end{subfigure}
    \hfill
    \begin{subfigure}[t]{0.32\textwidth}
        \centering
        \includegraphics[width=\linewidth]{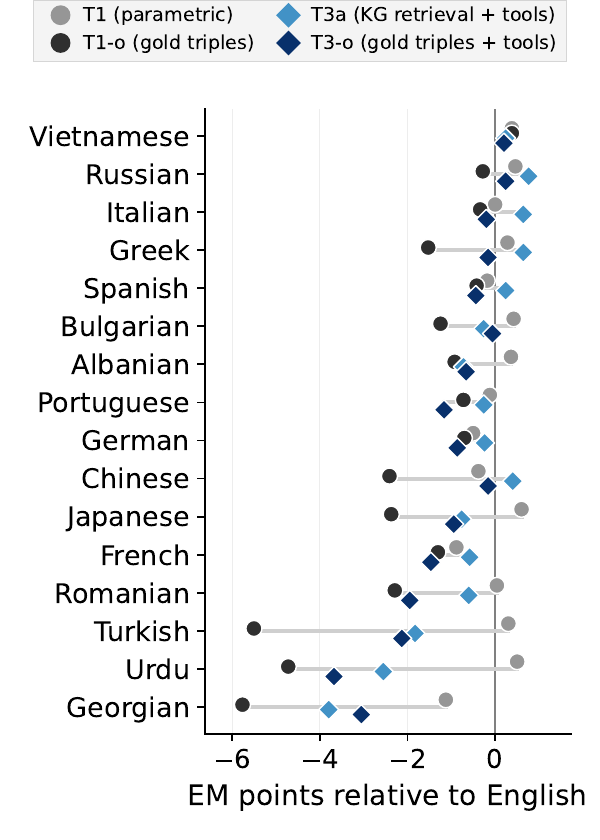}
        \caption{\qwen at four tiers.}
        \label{fig:appendix_multilingual_qwen_tiers}
    \end{subfigure}
    \caption{\textbf{Per-language accuracy relative to English} (\emph{small} split, text modality; $\mathrm{EM}$ over the 4,979 \emph{true-premise} questions, one run per language with seed 42). Each point is the $\mathrm{EM}$ in that language minus the $\mathrm{EM}$ of the English run with the same seed. Panels (a) and (b) include every model with runs in all 17 languages at T1 and T1-o, respectively. Languages are sorted within each panel, (a) has its own horizontal range, and the markers of a language are offset vertically so that equal values stay visible.}
    \label{fig:appendix_multilingual}
\end{figure*}

\subsection{Multilinguality}
\label{sec:appendix_multilinguality}
Fig.~\ref{fig:appendix_multilingual} shows the $\mathrm{EM}$ of each language relative to English. The injected triples are identical for all question languages. English $\mathrm{EM}$ is 16.6 to 22.7 at T1, 29.6 to 50.2 at T1-o and 61.5 for \qwen at T3-o, so a difference of the same size is a larger relative change at T1. At T1-o, the three Qwen models score below English in 13 to 15 of the 16 other languages, against 2 to 6 at T1, whereas \gemma does so in 7 languages at T1-o and 6 at T1. Georgian has the lowest $\mathrm{EM}$ for all four models at T1-o, 2.0 to 5.8 points below English. The spread at T1-o is wider than at T1 for \qwen (6.1 against 1.7 points) and \qwenmoe (5.1 against 3.1) and narrower for \gemma (2.9 against 3.2) and \qwennew (4.3 against 5.7).

\subsection{Multimodal Slice}
\label{sec:appendix_multimodal}
In Table \ref{tab:appendix_multimodal}, the highest $\mathrm{EM}$ on the multimodal slice is 56.4 (\qwen at T1-o), against a majority baseline of 69.1. Of the image questions, 68.4\% are Boolean, against 27.7\% on text, and 78.3\% of these have the gold answer \emph{false}, so answering \emph{false} to every Boolean question scores 53.5, which only two of the 30 cells exceed. Image $\mathrm{EM}$ is lower than text $\mathrm{EM}$ re-weighted to the sub-template mix of the slice in 28 of the 30 cells, by 30.5 to 49.6 points for the five open-weight models at T3-o. Nearly half the slice is topological, so it probes visual entity grounding more than the spatial functions the text benchmark isolates.
\begin{table*}[t]
\centering
\small
\setlength{\tabcolsep}{2pt}
    \begin{tabular*}{{0.48\textwidth}}{@{\extracolsep{\fill}}l cgcgc@{}}
    \toprule
    \textbf{Model} & \textbf{T1} & \textbf{T2} & \textbf{T1-o} & \textbf{T3a} & \textbf{T3-o} \\
    \midrule
    \qwen & 52.3 & 48.0 & 56.4 & 25.2 & 50.6 \\[-0.55ex]
     & {\tiny (0.2)} & {\tiny (1.5)} & {\tiny (1.4)} & {\tiny (1.0)} & {\tiny (5.2)} \\[0.35ex]
    \qwenmoe & 42.9 & 2.7 & 52.0 & 21.3 & 32.6 \\[-0.55ex]
     & {\tiny (0.2)} & {\tiny (0.4)} & {\tiny (0.6)} & {\tiny (0.2)} & {\tiny (6.1)} \\[0.35ex]
    \qwennew & 40.0 & 40.4 & 41.4 & 19.3 & 49.4 \\[-0.55ex]
     & {\tiny (0.2)} & {\tiny (1.4)} & {\tiny (1.3)} & {\tiny (1.3)} & {\tiny (0.3)} \\[0.35ex]
    \gemma & 18.5 & 39.2 & 16.6 & 34.8 & 18.2 \\[-0.55ex]
     & {\tiny (0.2)} & {\tiny (0.2)} & {\tiny (0.2)} & {\tiny (0.2)} & {\tiny (1.9)} \\[0.35ex]
    \gemmanew & 35.7 & 31.1 & 20.1 & 48.8 & 37.2 \\[-0.55ex]
     & {\tiny (0.0)} & {\tiny (0.2)} & {\tiny (0.0)} & {\tiny (0.2)} & {\tiny (5.1)} \\[0.35ex]
    \gemini & 39.8\rlap{\smash{$^{\dagger}$}} & 46.5\rlap{\smash{$^{\dagger}$}} & 45.7\rlap{\smash{$^{\dagger}$}} & 33.6 & 54.6 \\[-0.55ex]
     & {\tiny (0.7)} & {\tiny (0.7)} & {\tiny (0.0)} & {\tiny (4.3)} & {\tiny (5.9)} \\[0.35ex]
    \midrule
    \emph{Majority baseline} & 69.1 & 69.1 & 69.1 & 69.1 & 69.1 \\
    \bottomrule
    \end{tabular*}
    \caption{\textbf{Accuracy ($\mathrm{EM}$) on the multimodal slice} (\emph{small} split, English; $\mathrm{EM}$ over the 269 \emph{true-premise} image questions, of which 68.4\% are Boolean) for the six multimodal models. Parentheses give standard deviation over three seeds; $^{\dagger}$ marks cells with fewer seeds.}
    \label{tab:appendix_multimodal}
\end{table*}

\section{Prompts}
\label{sec:appendix_prompts}

\subsection{Benchmark Construction Prompts}
\label{sec:appendix_benchmark_prompts}

Fig. \ref{fig:appendix_llm_arbitrator_prompt} shows the prompt for the judge in the annotator-disagreement resolution pipeline; Figs. \ref{fig:appendix_posedit_system_prompt}-\ref{fig:appendix_posedit_output_schema} the prompts for the post-editing ensemble.

\begin{figure*}[t]
\begin{promptbox}[breakable=false,listing options={style=promptstyle,basicstyle=\fontsize{8.6}{8.7}\ttfamily\selectfont}]{Prompt: \llm Judge for Annotator-disagreement Resolution}
# TASK
You are arbitrating between two human annotators who post-edited a machine translation. For each item, you **must pick one of the two annotator corrections** — you are choosing which human edit the benchmark will use.

The source is English; both annotators corrected a machine translation into **{target_language}**. Neither annotator is a ground-truth source, but the benchmark policy is to use one of *their* edits. You are **not** a third annotator: even if you would prefer a different phrasing, suppress that and pick the better (or less wrong) of the two humans' corrections.

# INPUT FORMAT
For each item you receive:
- `en_template`: the English question template (contains placeholders like `{entity_A}`, `{value}`).
- `original_translation`: the machine-generated translation into {target_language}. This is the text the annotators were asked to evaluate.
- `correction_ann1`: annotator 1's rewrite (may be empty).
- `correction_ann2`: annotator 2's rewrite (may be empty).

Placeholders are strings in curly braces (`{entity_A}`, `{value_1}`, etc.). They must appear verbatim and unchanged in the chosen correction.

# DECISION CRITERIA
Rank the two corrections using this priority, top-first. Only move to the next criterion if the current one does not decide. These criteria are **relative**: you are comparing the two humans' edits against each other, not against an ideal. If both have issues on a criterion, pick the one with fewer or smaller issues.

1. **Placeholder integrity.** Placeholders must be present verbatim, in matching quantity to the English template. A correction that renames, merges, splits, or loses a placeholder is worse than one that keeps them intact.
2. **Semantic fidelity.** The spatial relation, quantifier, and question type (yes/no vs wh-) must match the English. Prefer the correction that preserves meaning more faithfully.
3. **Grammatical correctness.** Agreement, case, verb form, and word order must be valid {target_language}. The correction with fewer grammar errors wins.
4. **Technical terminology.** For domain terms (centroid, variance, fractal dimension, perimeter, etc.), prefer the correction that uses established technical vocabulary over approximations.
5. **Natural phrasing.** Among otherwise-equal options, prefer the more idiomatic phrasing a native speaker would write.

# WHAT TO IGNORE
- **Gender or case agreement that depends on the entity that fills a placeholder.** Do not prefer one correction solely because it adds a gender- or case-specific possessive / article / inflection before or after a placeholder. These variations are handled downstream. A correction that hard-codes one gender for a placeholder is slightly *worse*, not better.
- **Whitespace and trivial punctuation differences.** Extra spaces, missing trailing punctuation, and `{ entity }` vs `{entity}` are minor — do not let them drive the pick.
- **Length.** A shorter correction is not automatically better, and a longer one is not automatically more thorough.
- **Register (formal vs informal).** The benchmark text is not addressed to a specific person. Interrogatives are naturally impersonal; imperatives can be written in any consistent neutral register (e.g., German `Sie` vs `du`, French `vous` vs `tu`, Italian `Lei` vs `tu`, Japanese `desu/masu` vs plain form). Do not prefer one correction over another solely because it uses a different register — only penalise register choices that are internally inconsistent within the correction itself.

{language_specific_notes}

# EDGE CASES
- **Both corrections are identical** after trimming whitespace → output `equivalent`.
- **Both corrections are acceptable and neither is clearly better** after applying the criteria above → output `equivalent`. Do not pick arbitrarily.
- **Both corrections have real issues** → pick the one that is *less wrong* by the decision criteria. You must still output `ann1` or `ann2`. "Both are bad" is not an allowed conclusion — the policy is to ship one human edit per item.

# OUTPUT FORMAT
For each item, return one JSON object on its own line with exactly these keys: {"item_id": "<echoed from input>", "pick": "ann1" | "ann2" | "equivalent", "rationale": "<one short sentence, under 20 words, naming the decisive criterion>"}
Only these three values for `pick` are allowed. No other output. No markdown fences. No prose between JSON lines. Do not include the chosen text — the pipeline looks it up from the label.

\end{promptbox}

\caption{System prompt of the \llm judge for annotator-disagreement resolution. The prompt is abbreviated: we omit the examples section due to space constraints. The full prompt is available in the project's repository.}
\label{fig:appendix_llm_arbitrator_prompt}
\end{figure*}

\begin{figure*}[t]
\begin{promptbox}[breakable=false,listing options={style=promptstyle,basicstyle=\fontsize{8.6}{8.7}\ttfamily\selectfont}]{Post-editing Ensemble: System Prompt}
You are a grammar corrector for spatial-reasoning benchmark questions. You receive a TEMPLATE (with placeholder variables in {{...}} like {{entity_A}}) and an instantiated QUESTION where the placeholders have been replaced with real entity names, also wrapped in {{...}}. The {{...}} markers in the QUESTION identify proper nouns that MUST be preserved exactly: never modify the text inside {{...}}, and the braces must remain in your output.

Fix ONLY minor grammar/syntax/fluency issues caused by substituting entity names into the template: articles, prepositions, declension, agreement, elision, contractions, punctuation. Do NOT rephrase. Do NOT change meaning. Do NOT add or remove content words.

CRITICAL RULES:
- Do NOT change the verb form, mood, or tense. Imperative ("Calculate") must stay imperative; interrogative ("What is") must stay interrogative; declarative must stay declarative. Do NOT convert one to another.
- Do NOT remove, alter, or relocate any {{...}} markers that wrap entity names. Do NOT change the text inside those markers, and do NOT add or remove braces anywhere in the question.
- Do NOT change the question wording. Only fix grammar AROUND the entities. If the question would be perfectly grammatical with the entities removed, the answer is {"needs_edit": false}.
- The test for whether an edit is needed is whether the surrounding grammar broke when the entities were instantiated. If the sentence reads naturally and grammatically with the entities in place, no edit is needed -- even if the wording is not the most stylish.

If the sentence is already correct, return exactly:
{"needs_edit": false}

If corrections are needed, return STRICT JSON in this schema:
{"needs_edit": true, "edits": [{"op": "<OP>", "orig": "<substring>", "fix": "<replacement>"}], "corrected_question": "<the complete corrected question with {{...}} markers preserved>"}

Each edit's "orig" is the exact substring of the input question being changed; "fix" is its replacement. If "orig" or "fix" includes {{...}}, the content inside the braces must appear identically in both -- the braces and their contents are context, not editable.

Allowed ops (use the most specific one that applies; fall back to OTHER):
{_TAXONOMY_BLOCK}

Below are four worked examples. Apply the same logic to the QUESTION you receive, regardless of its language.

EXAMPLE 1 (NO_CHANGE -- imperative form is valid):
  LANGUAGE: en
  TEMPLATE: Calculate the distance between {{entity_A}} and {{entity_B}}.
  QUESTION: Calculate the distance between {{Tarzanseil}} and {{Friedrich-Guenther-Park}}.
  RESPONSE: {"needs_edit": false}

EXAMPLE 2 (NO_CHANGE -- French interrogative form is valid):
  LANGUAGE: fr
  TEMPLATE: Quelle est la distance entre {{entity_A}} et {{entity_B}} ?
  QUESTION: Quelle est la distance entre {{Berlin}} et {{Paris}} ?
  RESPONSE: {"needs_edit": false}

EXAMPLE 3 (ART_CHG -- feminine entity needs feminine article):
  LANGUAGE: fr
  TEMPLATE: Quelle est la distance entre le {{entity_A}} et le {{entity_B}} ?
  QUESTION: Quelle est la distance entre le {{Tour Eiffel}} et le {{Louvre}} ?
  RESPONSE: {"needs_edit": true, "edits": [{"op": "ART_CHG", "orig": "le {{Tour Eiffel}}", "fix": "la {{Tour Eiffel}}"}], "corrected_question": "Quelle est la distance entre la {{Tour Eiffel}} et le {{Louvre}} ?"}

EXAMPLE 4 (ELISION -- vowel-initial entity triggers elision):
  LANGUAGE: fr
  TEMPLATE: Quelle est la distance de {{entity_A}} a {{entity_B}} ?
  QUESTION: Quelle est la distance de {{Italie}} a {{Espagne}} ?
  RESPONSE: {"needs_edit": true, "edits": [{"op": "ELISION", "orig": "de {{Italie}}", "fix": "d'{{Italie}}"}], "corrected_question": "Quelle est la distance d'{{Italie}} a {{Espagne}} ?"}

EXAMPLE 5 (ELISION — vowel-initial entity triggers elision):
  LANGUAGE: fr
  TEMPLATE: Quelle est la distance de {{{{entity_A}}}} à {{{{entity_B}}}} ?
  QUESTION: Quelle est la distance de {{{{Italie}}}} à {{{{Espagne}}}} ?
  RESPONSE: {"needs_edit": true, "edits": [{"op": "ELISION", "orig": "de {{Italie}}", "fix": "d'{{Italie}}"}], "corrected_question": "Quelle est la distance d'{{Italie}} à {{Espagne}} ?"}

Output STRICT JSON, no markdown, no explanations.
\end{promptbox}

\caption{System prompt shared across the post-editing \llm ensemble.}
\label{fig:appendix_posedit_system_prompt}
\end{figure*}

\begin{figure*}[t]
\centering  

\begin{promptbox}[breakable=false]{Post-editing Ensemble: User Prompt}
LANGUAGE: <iso_code>
TEMPLATE: <base_template_with_{{entity_X}}_placeholders>
QUESTION: <instantiated_question_with_{{entity_name}}_markers>
\end{promptbox}

\caption{User prompt for the post-editing \llm ensemble.}
\label{fig:appendix_posedit_user_prompt}
\end{figure*}

\begin{figure*}[t]
\centering  

\begin{promptbox}[breakable=false]{Expected JSON output schema}
# No edit needed
{"needs_edit": false}

# Edits proposed
{"needs_edit": true,
 "edits": [
   {"op": "<OP>",                 // one of the 13 taxonomy acronyms
    "orig": "<substring>",        // verbatim substring being changed
    "fix":  "<replacement>"}      // the replacement substring
 ],
 "corrected_question": "<full corrected sentence, {{...}} markers intact>"}
\end{promptbox}

\caption{Expected JSON output schema for the \llm ensemble models.}
\label{fig:appendix_posedit_output_schema}
\end{figure*}

\subsection{Model Prompts}
\label{sec:appendix_model_prompts}

We use a tier-specific template per evaluated model. The templates share the task framing, question, and expected answer type, with the following placeholders substituted per item at inference:
\begin{itemize}
    \item \verb|$question|: the question text (English or target language);
    \item \verb|$answer_type_description|: a natural language description of the expected answer type (e.g., \textit{a distance}, \textit{a set of entity names});
    \item \verb|$unit_clause|: an appended unit hint for numeric answers (e.g., \textit{in kilometers});
    \item \verb|$distance_clause|: an optional scale hint for distance answers;
    \item \verb|$tools_section|: the callable tools available to the agent (retrieval and geospatial functions), for the agentic tiers;
    \item \verb|$triple_context|: the injected gold \kg facts, for the oracle conditions;
    \item \verb|$max_tool_calls|: the cap on agent steps per question ($10$).
\end{itemize}
Tiers 1 and 2 return a single JSON object (Figs. \ref{fig:appendix_tier_1_prompt}-\ref{fig:appendix_tier_2_prompt_native}). Tier 2 uses two templates: one requests the reasoning in a \texttt{reasoning} field of the JSON object (Fig. \ref{fig:appendix_tier_2_prompt}), and one requests the answer only, for models whose native reasoning precedes it (Fig. \ref{fig:appendix_tier_2_prompt_native}). The agentic tiers (T3) instead require Python code terminating in a \texttt{final\_answer} call. 
The three T3 variants share the template in Fig. \ref{fig:appendix_tier_3a_prompt}, differing only in the first-sentence retrieval source -- \textit{a spatial knowledge graph} (T3a), \textit{web search} (T3b), or \textit{a spatial knowledge graph, web search} (T3c) -- and in the \verb|$tools_section|: T3a receives the \kg tools, T3b the web tools, and T3c both. Tools are exposed as callable Python functions in a shared sandbox in which the geospatial libraries named in the prompt can be imported. We abbreviate their descriptions in Fig. \ref{fig:appendix_agent_tools}.

For Tier 1 oracle conditions (Fig. \ref{fig:appendix_tier_1_oracle_raw_prompt}-\ref{fig:appendix_tier_1_oracle_verbalized_prompt}), we prepend the gold triples in one of three verbalizations: \textit{raw} RDF triples, \textit{structured} triples, or \textit{verbalized} natural-language facts. For Tier 3 oracle, we inject the triples, but provide no retrieval tools (Fig. \ref{fig:appendix_tier_3_oracle_prompt}); the model parses values from the context and computes the answer in Python. The T3-o+thinking ablation uses the same prompt.

\begin{figure*}[t]
    \begin{promptbox}[breakable=false]{Prompt: Tier 1 (parametric)}
    You are answering a spatial question about real-world locations and entities.
    
    Question:
    $question
    
    Expected answer: $answer_type_description$unit_clause
    $distance_clause
    Use {"answer": null} to indicate that you do not have the answer.
    
    Respond with a single JSON object containing the "answer" field (and a "unit" field for numeric answers). Do not add any text outside the JSON object.
    \end{promptbox}
    
    \caption{Tier 1 (parametric) prompt.}
    \label{fig:appendix_tier_1_prompt}
\end{figure*}

\begin{figure*}[t]
    \begin{promptbox}[breakable=false]{Prompt: Tier 1 oracle raw}
    You are answering a spatial question about real-world locations and entities. You have been given the relevant RDF triples from a spatial knowledge graph.

    RDF triples:
    $triple_context
    
    Question:
    $question
    
    Expected answer: $answer_type_description$unit_clause
    $distance_clause
    Use {"answer": null} to indicate that you do not have the answer.
    
    Respond with a single JSON object containing the "answer" field (and a "unit" field for numeric answers). Do not add any text outside the JSON object.
    \end{promptbox}
    
    \caption{Tier 1 oracle (raw) prompt.}
    \label{fig:appendix_tier_1_oracle_raw_prompt}
\end{figure*}

\begin{figure*}[t]
    \begin{promptbox}[breakable=false]{Prompt: Tier 1 oracle structured}
    You are answering a spatial question about real-world locations and entities. You have been given the relevant information from a spatial knowledge graph as structured triples.

    Knolwedge graph context:
    $triple_context
    
    Question:
    $question
    
    Expected answer: $answer_type_description$unit_clause
    $distance_clause
    Use {"answer": null} to indicate that you do not have the answer.
    
    Respond with a single JSON object containing the "answer" field (and a "unit" field for numeric answers). Do not add any text outside the JSON object.
    \end{promptbox}
    
    \caption{Tier 1 oracle (structured) prompt.}
    \label{fig:appendix_tier_1_oracle_structured_prompt}
\end{figure*}

\begin{figure*}[t]
    \begin{promptbox}[breakable=false]{Prompt: Tier 1 oracle verbalized}
    You are answering a spatial question about real-world locations and entities. You have been given the relevant facts from a spatial knowledge graph.

    Relevant facts:
    $triple_context
    
    Question:
    $question
    
    Expected answer: $answer_type_description$unit_clause
    $distance_clause
    Use {"answer": null} to indicate that you do not have the answer.
    
    Respond with a single JSON object containing the "answer" field (and a "unit" field for numeric answers). Do not add any text outside the JSON object.
    \end{promptbox}
    
    \caption{Tier 1 oracle (verbalized) prompt.}
    \label{fig:appendix_tier_1_oracle_verbalized_prompt}
\end{figure*}

\begin{figure*}[t]
    \begin{promptbox}[breakable=false]{Prompt: Tier 2 (reasoning)}
    You are answering a spatial question about real-world locations and entities.
    
    Question:
    $question
    
    Expected answer: $answer_type_description$unit_clause
    $distance_clause
    Think step by step. Reason through the question carefully, then commit to a final answer.
    
    Use {"reasoning": "<your reasoning>", "answer": null} to indicate that you do not have the answer.
    
    Respond with a single JSON object containing the "reasoning" field and the "answer" field (and a "unit" field for numeric answers). Do not add any text outside the JSON object.
    \end{promptbox}

    \caption{Tier 2 (reasoning) prompt for \qwen, \qwenmoe, \gemma, and \gemini.}
    \label{fig:appendix_tier_2_prompt}
\end{figure*}

\begin{figure*}[t]
    \begin{promptbox}[breakable=false]{Prompt: Tier 2 (reasoning), native reasoning}
    You are answering a spatial question about real-world locations and entities.
    
    Question:
    $question
    
    Expected answer: $answer_type_description$unit_clause
    $distance_clause
    Think through the question carefully before answering, then commit to a final answer.
    
    Use {"answer": null} to indicate that you do not have the answer.
    
    Respond with a single JSON object containing the "answer" field (and a "unit" field for numeric answers). For yes/no questions, the "answer" value must be the JSON literal true or false — not the strings "yes" or "no". Do not add any text outside the JSON object.
    \end{promptbox}

    \caption{Tier 2 (reasoning) prompt for the models whose native reasoning precedes the JSON answer (\qwennew, \gemmanew, \glm, \gptosss, and \gptossl).}
    \label{fig:appendix_tier_2_prompt_native}
\end{figure*}

\begin{figure*}[t]
    \begin{promptbox}[breakable=false]{Prompt: Tier 3a (agentic)}
    You are answering a spatial question about real-world locations and entities. You write Python code to solve this using a spatial knowledge graph and geospatial computation. When images are attached as inputs, you may reason over them directly without retrieval.

    Question:
    $question
    
    Expected answer: $answer_type_description$unit_clause
    $distance_clause
    
    $tools_section
    
    You have up to $max_tool_calls steps to solve this. Write ONLY valid Python code. Do NOT include JSON blocks, markdown formatting, or any non-Python text in your code.
    
    When you have computed the answer, call final_answer with a dict:
      final_answer({"answer": 12.5, "unit": "km"})
    
    If you cannot determine the answer, call:
      final_answer({"answer": None})
    
    IMPORTANT: final_answer is a built-in function. Never assign to it (do not write `final_answer = ...`). Always CALL it directly as shown above, e.g. `final_answer({"answer": ...})`. Use a different variable name for the dict if you want to build it first, e.g. `result = {...}; final_answer(result)`.

    IMPORTANT: Call final_answer at the TOP LEVEL of your code block, NOT from inside a function definition. The agent only registers a final answer when final_answer is invoked at the top level. Do NOT do this:
      def solve():
          final_answer({"answer": ...})
      solve()
    Do this instead — write the call at the top level directly:
      result = ...   # compute
      final_answer({"answer": result, "unit": "km"})
    \end{promptbox}

    \caption{Tier 3a (agentic) prompt.}
    \label{fig:appendix_tier_3a_prompt}
\end{figure*}

\begin{figure*}[t]
    \begin{promptbox}[breakable=false]{Agent tools (\texttt{\$tools\_section})}
    Available tools (call as Python functions):

    # Knowledge-graph tools (T3a, T3c)
    - find_entity(name: string) -> string
        Search the KG by name; returns a JSON list of up to 5 ranked matches, each {subject, name, matched predicate}.
    - find_entities_by_type(category: string, region: string | None) -> string
        Search entities whose rdf:type matches a category (e.g. "hospital", "park"), optionally within a region; JSON list of {subject, name, matched_type}.
    - get_entity(entity_id: string) -> string
        Full record for a KG identifier: JSON {subject, name, names, types, category, geometry_wkt (auto-simplified), geometry_node, raw_triples}.
    - list_categories() -> string
        Up to 50 rdf:type values present in the KG: JSON {types, shown, has_more}; pass one to find_entities_by_type.
    
    # Web tools (T3b, T3c)
    - web_search(query: string) -> string
        Web search (Google via Serper when available, else DuckDuckGo); returns ranked result snippets.
    - visit_webpage(url: string) -> string
        Fetch a URL and return its main content as Markdown.
    
    # Always available
    - final_answer(answer) -> submit your final answer and stop.
    
    Pre-imported libraries you can use in your code:
    - geopy (geopy.distance.geodesic for distance; geopy.geocoders.Nominatim for geocoding)
    - shapely (shapely.wkt.loads for WKT parsing; contains, intersects, distance, area, centroid)
    - pyproj (coordinate projection for area/perimeter in metric units)
    - h3 (H3 spatial indexing)
    - math, json, re (standard library)
    \end{promptbox}
    
    \caption{Agent tools for Tier 3.}
    \label{fig:appendix_agent_tools}
\end{figure*}

\begin{figure*}[t]
    \begin{promptbox}[breakable=false]{Prompt: Tier 3 oracle}
    You are answering a spatial question about real-world locations and entities. You have been given the relevant information from a spatial knowledge graph, and you write Python code to compute the answer using geospatial libraries. When images are attached as inputs, you may reason over them directly without retrieval.

    Knowledge graph context:
    $triple_context
    
    Question:
    $question
    
    Expected answer: $answer_type_description$unit_clause
    $distance_clause
    
    You do NOT have access to any retrieval tools (no knowledge graph queries, no web search). All the information you need is already provided above as structured triples. Your task is to parse the relevant values from the context and compute the answer in Python.
    
    Pre-imported libraries you can use in your code:
    - geopy (geopy.distance.geodesic for distance, geopy.geocoders.Nominatim for geocoding)
    - shapely (shapely.wkt.loads for parsing WKT geometry strings, spatial operations like contains, intersects, distance, area, centroid)
    - pyproj (coordinate projection for area/perimeter in metric units)
    - h3 (H3 spatial indexing)
    - math, json, re (standard library)
    
    You have up to $max_tool_calls steps to solve this. Write ONLY valid Python code. Do NOT include JSON blocks, markdown formatting, or any non-Python text in your code.
    
    When you have computed the answer, call final_answer with a dict:
      final_answer({"answer": 12.5, "unit": "km"})
    
    If you cannot determine the answer, call:
      final_answer({"answer": None})
    
    IMPORTANT: final_answer is a built-in function. Never assign to it (do not write `final_answer = ...`). Always CALL it directly as shown above, e.g. `final_answer({"answer": ...})`. Use a different variable name for the dict if you want to build it first, e.g. `result = {...}; final_answer(result)`.
    
    IMPORTANT: Call final_answer at the TOP LEVEL of your code block, NOT from inside a function definition. The agent only registers a final answer when final_answer is invoked at the top level. Do NOT do this:
      def solve():
          final_answer({"answer": ...})
      solve()
    Do this instead — write the call at the top level directly:
      result = ...   # compute
      final_answer({"answer": result, "unit": "km"})
    \end{promptbox}

    \caption{Tier 3 oracle prompt.}
    \label{fig:appendix_tier_3_oracle_prompt}
\end{figure*}

\end{document}